\documentclass[10pt,twocolumn,letterpaper]{article}

\usepackage[pagenumbers]{cvpr}
\usepackage[dvipsnames]{xcolor}
\usepackage{algorithm}
\usepackage{algpseudocode}
\makeatletter
\AtBeginDocument{\def\theHALG@line{\thealgorithm.\arabic{ALG@line}}}
\makeatother

\usepackage{graphicx}
\usepackage{array}
\usepackage{adjustbox}
\usepackage{tikz}

\newlength{\qualimgsize}

\newcommand{\qualimg}[1]{%
  \raisebox{-0.5\height}{%
    \includegraphics[
      width=\qualimgsize,
      height=\qualimgsize,
      keepaspectratio
    ]{#1}%
  }%
}

\newcommand{\qualheader}[1]{%
  \makebox[\qualimgsize]{\scriptsize\textbf{#1}}%
}

\newcommand{\qualinstruction}[1]{%
  \parbox[t]{\dimexpr 2\qualimgsize+4pt\relax}{%
    \centering\scriptsize #1%
  }%
}

\newcommand{\Loss}{\mathcal{L}}

\newcommand{\dirn}{\delta_\text{N}}
\newcommand{\dird}{\delta_\text{D}}
\newcommand{\dirc}{\delta_\text{C}}

\newcommand{\z}{\mathbf{z}}
\newcommand{\x}{\mathbf{x}}
\newcommand{\grad}[1]{\nabla_{\!#1}}
\newcommand{\unet}{\epsilon_{\phi}}
\newcommand{\epred}{\unet(\z_t;c,t)}

\newcommand{\epredcfgx}{\unet^{\gamma}(\z_t(\x);c,t)}

\definecolor{cvprblue}{rgb}{0.21,0.49,0.74}
\usepackage[pagebackref,breaklinks,colorlinks,allcolors=cvprblue]{hyperref}

\def\paperID{*****} %
\def\confName{CVPR}
\def\confYear{2027}

\title{ACEdit: Stable Multi-Turn Image Editing via Asymmetric Conditioning}

\author{
Sara Dorfman\footnotemark[1] \qquad
Nir Goren\footnotemark[1] \qquad
Gal Barak\footnotemark[1] \qquad
Daniel Cohen-Or \qquad
Or Patashnik\\[6pt]
Tel Aviv University
}

\begin{document}
\twocolumn[{%
    \renewcommand\twocolumn[1][]{#1}%
    \maketitle
    
\begin{center}
        \includegraphics[width=\textwidth]{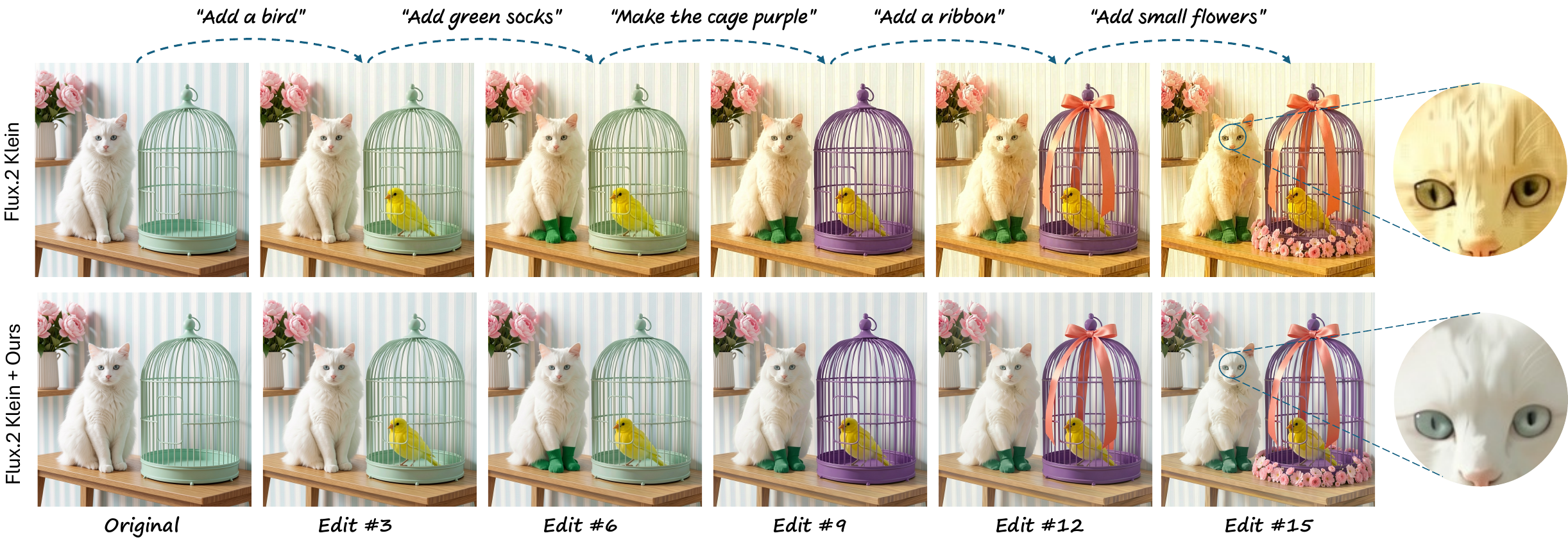}
        \captionof{figure}{\textbf{Stable Multi-Turn Image Editing.} We show selected turns from a 15-step sequence. Each of the five edits is followed by two no-op instructions (“keep the image as is”). While the original FLUX.2 Klein gradually loses detail and develops a strong color shift, fine-tuning it with our training scheme yields a more stable editing process, better preserving the scene and previously edited content while faithfully following the requested changes. The zoom-ins highlight the loss of fine fur texture and the emergence of harsh, unnatural lines in the baseline, while our model preserves the fur's soft texture and natural appearance.}
        \label{fig:teaser}
    \end{center}

}]
\renewcommand{\thefootnote}{\fnsymbol{footnote}}
\footnotetext[1]{Equal contribution.}
\renewcommand{\thefootnote}{\arabic{footnote}}

\begin{abstract}
Image editing models produce impressive results for individual edits, but their quality often deteriorates over multiple editing turns.
Each editing turn introduces small distribution shifts that accumulate over successive edits, creating a gap between the input distributions encountered during the editor's original training and multi-turn inference.
We introduce a training scheme that makes multi-turn image editing stable under this accumulating distribution shift. 
A key is to force the editor to operate on its own outputs from previous editing turns during training, bringing the principle of Self-Forcing to image editing.
The main challenge is that self-generated editing trajectories deviate from ground-truth trajectories, making direct supervision unreliable.
We address this challenge through a distillation objective that leverages a teacher image editor to provide a reliable training signal when the model operates on its own previous outputs.
We introduce an asymmetric conditioning scheme: while the student performs edits sequentially on its own outputs, the teacher remains anchored to the original clean image and receives a compound editing instruction.
Experiments show substantially reduced degradation over long multi-turn sequences while maintaining editing quality and instruction adherence.
\end{abstract}
     
\vspace{17pt}
\section{Introduction}

Image editing with modern generative models has reached an impressive level of quality, enabling complex modifications while faithfully preserving the appearance and content of the input image~\cite{labs2025flux, wu2025qwen, tan2025ominicontrol, brooks2023instructpix2pix, kumari2025npedit}.
Yet, practical editing is rarely a single operation. 
Users often refine an image through a sequence of edits, where the output of one edit becomes the input to the next. 
This setting is commonly referred to as \emph{multi-turn image editing}.
With current image editing models, each editing turn introduces small distribution shifts that accumulate as the model repeatedly operates on its own outputs.
These shifts create a growing gap between the input distributions encountered during the editor's original training and multi-turn inference.
This gap leads to accumulating artifacts, loss of fine details, and unintended changes in appearance (see Figure~\ref{fig:teaser}).
Effective image editing therefore requires not only high-quality individual edits, but also stability across successive editing turns.

Several approaches have been proposed to mitigate degradation in multi-turn image editing.
Inference-time methods target specific sources of degradation~\cite{zhou2025multiturn,liao2026freqEdit,excave2026,wang2026vaelfa}, but may incur computational overhead and are limited by the complexity of distribution shifts in multi-turn editing.
Other approaches train models on multi-turn editing sequences~\cite{qu2026vincie,xu2026anchoredit,huang2026mteditflow,zhao2026mtopsd}.
Training on ground-truth sequences leaves a gap between training and inference, while training on self-generated sequences poses supervision challenges, leading to inconsistent targets, restrictions on training sequences, or objectives that do not explicitly address distribution shifts.

We present a training scheme for multi-turn image editing that maintains stability under accumulating distribution shifts.
Our scheme is inspired by autoregressive video generation, where models similarly operate on their own outputs and face accumulating distribution shifts.
Following the principle of Self-Forcing~\cite{huang2025selfforcing}, we train the editor on its own outputs from previous editing turns, allowing it to learn from the distribution shifts encountered during multi-turn inference.
However, self-generated editing trajectories inevitably deviate from ground-truth trajectories, making direct supervision with ground-truth images unreliable.

To address this challenge, we employ score distillation in a teacher-student framework to train the editor on its self-generated trajectories without requiring ground-truth targets for its intermediate outputs~\cite{poole2022dreamfusion, wang2023prolificdreamer, yin2024dmd}.
We introduce an asymmetric conditioning scheme for the student and teacher.
The student's input may already deviate significantly from the teacher's training distribution, making it unsuitable for reliable teacher guidance.
We therefore condition the teacher on the original image and a compound instruction combining all edits up to the current turn, while the student performs edits sequentially on its own outputs.
This provides a training signal that reflects the intended edits without inheriting the errors accumulated along the student's trajectory.

We evaluate our method against existing multi-turn image editing approaches and demonstrate significant qualitative and quantitative improvements in preserving editing capabilities while mitigating accumulated distribution drift.
Our training is efficient and introduces no inference-time overhead, while making no assumptions about the nature of distribution shifts or restrictions on editing sequences.
Through ablation studies, we further demonstrate the importance of our asymmetric conditioning scheme for stable multi-turn editing.

\section{Related Work}
\label{sec:related_work}

\paragraph{Text-guided image editing.}
Early text-driven manipulation largely steered pre-trained GAN latent spaces via vision-language models~\cite{patashnik2021styleclip,gal2022stylegan}.
The advent of diffusion models broadened semantic coverage and synthesized complex scenes more flexibly.
Subsequent approaches improved source-image preservation and fine-grained control through noise perturbation and trajectory inversion~\cite{meng2021sdedit,mokady2023null,wallace2023edict,huberman2024edit,brack2024ledits++,deutch2024turboedit}, alongside spatial blending and cross-attention steering~\cite{avrahami2022blended,hertz2022prompt,cao2023masactrl,tumanyan2023plug,alaluf2024cross}.
Recent efforts extend inference-time control to continuous-time flow models through structural layer control~\cite{avrahami2025stable}, flow inversion~\cite{rout2024semantic,kadosh2025tight}, inversion-free sampling trajectories~\cite{kulikov2025flowedit}, and text-embedding control for continuous editing~\cite{kamenetsky2026saedit,huberman2026token}.
Complementing inference-time methods, instruction-tuned architectures learn a direct mapping using synthetic pairs~\cite{brooks2023instructpix2pix}, human annotations~\cite{zhang2023magicbrush}, and multi-task supervision~\cite{geng2024instructdiffusion,sheynin2024emu}.
Emerging state-of-the-art transformer-based diffusion and flow backbones further push visual quality and compositional reasoning, demonstrating unprecedented instruction fidelity and fine-grained spatial control~\cite{labs2025flux,wu2026omnigen2,wu2025qwen,flux-2-2025}.
Despite these single-step capabilities, isolated edit accuracy does not prevent error propagation, identity degradation, or semantic drift when applied iteratively across extended editing sequences.

\vspace{-8pt}
\paragraph{Multi-turn image editing.}
Maintaining image quality and content consistency across successive edits is a central challenge in multi-turn image editing, as errors can accumulate over the editing sequence.
Agentic frameworks tackle this challenge through planning, memory, and reflection~\cite{shen2026imagagent,ye2026agent}, while training-free methods improve consistency through inversion and attention control~\cite{zhou2025multiturn}, high-frequency feature preservation~\cite{liao2026freqEdit}, inversion sharing~\cite{excave2026}, and low-frequency latent alignment~\cite{wang2026vaelfa}.
Other approaches learn sequential editing directly: VINCIE learns in-context editing from video-derived sequences~\cite{qu2026vincie}, while AnchorEdit combines supervised causal multi-turn training with self-generated histories and identity-anchoring memory~\cite{xu2026anchoredit}.
Reinforcement learning methods optimize trajectory-level editing rewards~\cite{huang2026mteditflow} or jointly optimize history-dependent reasoning and image generation~\cite{ye2026edit}.
More recently, MT-OPSD distills clean-conditioned editing behavior onto states generated through identity-edit rollouts~\cite{zhao2026mtopsd}, which accumulate errors while keeping the intended content fixed.
Our training instead follows the editing process as it unfolds during inference: successive semantic edits with evolving content and preservation requirements.
Through distribution matching, we align the student's evolving output distribution, modeled by a learned auxiliary editor, with the target distribution defined by a teacher conditioned on the original image and cumulative instructions, without paired intermediate targets.

\begin{figure*}[t]
    \centering
    \includegraphics[width=\textwidth]{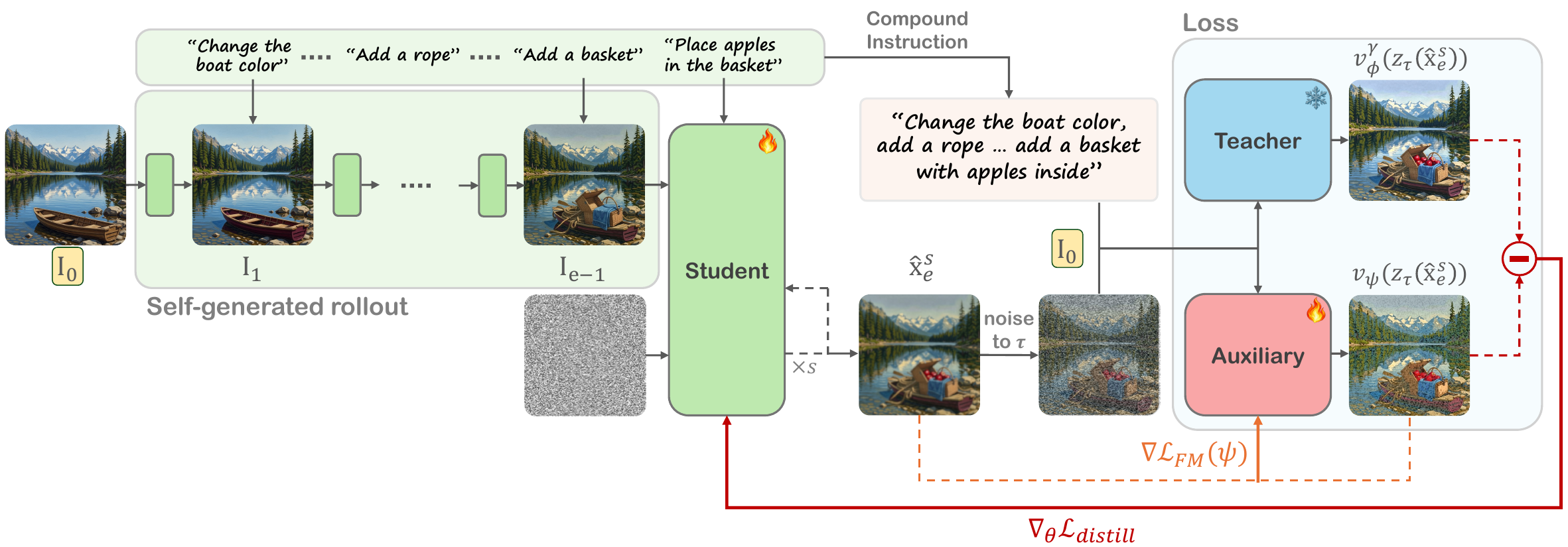}
    \caption{\textbf{Method overview.} \textit{Left:} The student performs edits along a self-generated rollout, where each output becomes the input to the next turn. Given the previous output $I_{e-1}$ and the current instruction, the few-step student produces a clean estimate $\hat{x}^s_e$. \textit{Right:} Score distillation supervision with \emph{asymmetric conditioning}. The student's estimate $\hat{x}^s_e$ is noised to level $\tau$ and fed to both a frozen teacher and a trainable auxiliary editor. Unlike the student, both are conditioned identically on the original clean image $I_0$ and a compound instruction aggregating all edits up to turn $e$; however, the teacher represents the desired clean-conditioned distribution, while the auxiliary editor is trained to model the distribution actually produced by the student at edit $e$. The difference between their predictions yields the distillation gradient, which pushes the student's multi-turn output distribution toward the teacher's clean, non-degraded distribution.}
    \label{fig:overview}
\end{figure*}

\section{Preliminaries: Score Distillation}

Score Distillation Sampling (SDS)~\cite{poole2022dreamfusion} uses a pretrained text-to-image diffusion model $\epsilon_\phi$ to optimize a differentiable renderer $G$ parameterized by $\theta$.
It relies on the observation that the model's noise prediction at each noise level is proportional to the score of the noisy image distribution~\cite{song2020score}.
SDS uses this score to guide images produced by $G$ without requiring paired target images.
For a sampled timestep $t$ and Gaussian noise $\epsilon$, the SDS gradient is:
\begin{equation}
\grad{\theta}\Loss_{\mathrm{SDS}}
=
w(t)\left(\epredcfgx-\epsilon\right)
\frac{\partial\x}{\partial\theta},
\label{eq}
\end{equation}
where $\x$ is an image produced by $G$, $\z_t(\x)$ is obtained by adding noise $\epsilon$ to $\x$ at timestep $t$, and $c$ is a condition such as a text prompt. 
Here, $\epredcfgx$ denotes the model's noise prediction with classifier-free guidance, and $w(t)$ weights the gradient at different timesteps.

SDS is known to produce blurry or oversaturated results, which has motivated numerous follow-up works~\cite{chung2023luciddreamer, zhu2023hifa, mcallister2024rethinking, yu2024text}. 
Some of these works have identified the high variance associated with subtracting the sampled noise $\epsilon$ from the model's prediction as a contributing factor.
In particular, Variational Score Distillation (VSD)~\cite{wang2023prolificdreamer} addresses this issue by replacing the sampled noise $\epsilon$ subtracted from the model's prediction with the prediction of an auxiliary model $\epsilon_{\mathrm{LoRA}}$ fine-tuned on images produced by $G$.

Noise-Free Score Distillation (NFSD)~\cite{katzir2024noise} provides a new interpretation of score distillation.
Specifically, it decomposes the conditional noise prediction into three components: condition alignment, correction toward the model's image distribution, and denoising: $\epred=\dirc+\dird+\dirn$.
Under this interpretation, subtracting the auxiliary prediction in VSD approximately cancels the denoising component, leaving the directions associated with condition alignment and domain correction to guide optimization.

While SDS and VSD optimize individual instances for a given prompt, Distribution Matching Distillation (DMD)~\cite{yin2024dmd,yin2024dmd2} extends this approach to training a generative model. 
DMD uses the VSD loss to train a \textit{student} model, guided by a pretrained diffusion model referred to as the \textit{teacher}.
The student learns to generate images for arbitrary prompts. 
Our method adopts this teacher–student formulation for training a stable editor.

Although these methods were originally formulated for noise-predicting diffusion models, score distillation can also be applied to flow-matching models~\cite{huang2025selfforcing}. This extension relies on converting predicted flow velocities into score estimates. We adopt this formulation in our method.

\section{Method}
\label{sec:method}

Image editing models are typically trained on pairs of input and target images~\cite{labs2025flux}. 
However, in multi-turn editing, the model repeatedly receives inputs generated by its own previous predictions, which it was not trained to handle. 
This creates a train–test gap that can cause errors and artifacts to accumulate over successive edits.

To address this gap, we train our multi-turn image editing model $G_\theta$ to edit images produced by repeatedly applying the model itself, as illustrated in Figure~\ref{fig:overview}.
We refer to the resulting sequences of edited images as rollouts. 
Supervising the model on these rollouts is challenging even when a ground-truth editing sequence is available.
Since an edit can have many valid outcomes, the model's output at a given turn may differ from the corresponding ground-truth image, even if the edit was applied correctly.
As a result, subsequent edits in the rollout are applied to images that differ from those in the ground-truth sequence.
Therefore, the ground-truth images cannot serve as paired targets for supervising these edits.

Inspired by Self-Forcing~\cite{huang2025selfforcing}, which addresses a similar challenge in autoregressive video generation, we use score distillation to supervise our student model $G_\theta$, which becomes our stable editor after training.
We compute the score-distillation signal using a pretrained flow-matching editor $v_\phi$ that serves as the teacher.

A key challenge is that the student's intermediate outputs may contain accumulated errors, making them unreliable inputs for the teacher. 
To address this, we introduce asymmetric conditioning of the student and teacher through compound-edit supervision.
While the student applies edits sequentially to its own outputs, the teacher is always conditioned on the original image and a compound instruction combining all edits up to the current turn. 
This allows the teacher to provide a training signal that does not inherit the errors accumulated along the student's trajectory.

\subsection{Self-Generated Editing Rollouts}
\label{sec:rollouts}

We generate editing rollouts by repeatedly applying the student editor $G_\theta$ to its own outputs (left part of Figure~\ref{fig:overview}).
Each rollout starts from a source image $I_0$ and a sequence of $E$ instructions $c_1,\ldots,c_E$, with $E=10$ during training.
We describe the dataset of instruction sequences in Section~\ref{sec:experiments}.
At turn $e$, the editor applies the next instruction $c_e$ to the completed output from the previous turn:
\[
I_e = G_\theta(I_{e-1},c_e), \qquad e=1,\ldots,E.
\]
Thus, after the first turn, the editor operates on its own generated images, as it does at inference.

We implement the student with FLUX.2-klein-9B, an image editor that operates in four denoising steps. 
We denote the complete four-step editing process by $G_\theta$ and a single model prediction within this process by $v_\theta$.
We use the convention where $\sigma=1$ corresponds to pure noise and $\sigma=0$ to a clean image, with the predicted velocity pointing toward noise.
At a denoising step with noise level $\sigma$, let $x_\sigma$ be the current state. 
The model's prediction provides an estimate of the clean output: $\hat{x}=x_\sigma-\sigma v_\theta(x_\sigma)$, where, 
for brevity, we omit the dependence of $v_\theta$ on the noise level, the previous image, and the current instruction.

For a given training rollout, we consider an edit turn $e\in\{1,\ldots,E\}$ and a denoising exit step $s\in\{1,\ldots,4\}$.
We first obtain $I_{e-1}$ by applying the preceding $e-1$ edits, each using the full four-step denoising process without tracking gradients.
At turn $e$, we execute the denoising process up to step $s$, where we compute the clean output estimate $\hat{x}^s_e$. 
Gradients are propagated only through the model's prediction at this step, without backpropagating through earlier denoising steps or edit turns. 
Throughout the rollout, we pass the edited images directly between turns in the model's latent space, avoiding repeated decoding and encoding.

\subsection{Training}
\label{sec:training}

\paragraph{Asymmetric Conditioning.}

Having obtained the clean-output estimate $\hat{x}^s_e$ at turn $e$, we now define the training signal.
This signal should encourage adherence to the editing instructions, preservation of relevant image details, and generation of realistic, artifact-free images.
We obtain this signal from a pretrained flow-matching image editor $v_\phi$, which serves as our teacher. 
Specifically, we evaluate its score on a noisy version of $\hat{x}^s_e$ at noise level $\tau$, denoted by $\mathbf{z}_\tau(\hat{x}^s_e)$. 
Unlike conventional text-to-image score distillation, our editing teacher is conditioned on both an image and an instruction. 
It can therefore guide the output toward results that satisfy the requested edit while preserving relevant details of the conditioning image.

The remaining question is how to condition the pretrained editing teacher $v_\phi$.
Its conditioning image should serve as the source of the edit, while its text input should describe the intended change from that image to $\hat{x}^s_e$. 
A natural choice is the previous output $I_{e-1}$ and the current instruction $c_e$, which are also the inputs to the student $G_\theta$ at turn $e$.
However, beyond the first turn, $I_{e-1}$ is generated by the student and may already contain artifacts.
This creates a distribution shift for the teacher, which was trained on clean images.
Moreover, since the teacher is expected to preserve details of its conditioning image that are not affected by the instruction, it may also preserve these artifacts rather than correct them.
We examine this conditioning choice in the ablation study in Section~\ref{sec:exp:ablations}.

Instead, we condition the teacher on the original clean image $I_0$ and the accumulated instructions $h_e=(c_1,\ldots,c_e)$.
This allows the teacher to score the result of multiple edits as a single compound edit, without relying on potentially corrupted intermediate images.
For example, a sequence of four editing instructions can be treated as a single compound edit:
\[
\vspace{-3pt}
\small
\begin{aligned}
\textbf{Student:}\quad
&\xrightarrow{\substack{\text{``add}\\\text{handbag''}}}
\;\;
\xrightarrow{\substack{\text{``make it}\\\text{red''}}}
\;\;
\xrightarrow{\substack{\text{``make it}\\\text{leather''}}}
\;\;
\xrightarrow{\substack{\text{``add a gold}\\\text{buckle''}}}
\\
\textbf{Teacher:}\quad
&\xrightarrow{\text{``add a red leather handbag with a gold buckle''}}
\end{aligned}
\]
To implement this supervision, the teacher $v_\phi$ evaluates a noisy version of the student's clean-output estimate $\hat{x}^s_e$, denoted by $\mathbf{z}_\tau(\hat{x}^s_e)$, and produces the velocity prediction $v_\phi(\mathbf{z}_\tau(\hat{x}^s_e);I_0,h_e,\tau)$ used for score distillation. 

This creates an asymmetry between the student and teacher.
While the student applies each instruction to its own previous output, the teacher evaluates the result relative to the original image and the accumulated instructions.
Consequently, the supervision remains anchored to the clean source image, even when errors accumulate across the student's editing turns.
At the first turn, the teacher's conditioning $(I_0,h_1)$ coincides with the student's input $(I_0,c_1)$, so every rollout also supervises an ordinary single edit; we evaluate single edits separately in Section~\ref{sec:exp:comparison}.

\paragraph{Loss.}
Following VSD~\cite{wang2023prolificdreamer} and its adaptation to generative models in DMD~\cite{yin2024dmd,yin2024dmd2}, we update the student $G_\theta$ using the difference between predictions of the pretrained teacher $v_\phi$ and an auxiliary flow-matching editor $v_\psi$ trained alongside the student on its generated outputs.
For the student's clean-output estimate $\hat{x}^s_e$, both models evaluate the same noisy version $\mathbf{z}_\tau(\hat{x}^s_e)$ under the conditioning $(I_0,h_e)$.
We denote their velocity predictions by:
\begin{equation*}
\hat{v}_\phi^\gamma =
v_\phi^\gamma(\mathbf{z}_\tau(\hat{x}^s_e); I_0,h_e,\tau),
\quad
\hat{v}_\psi =
v_\psi(\mathbf{z}_\tau(\hat{x}^s_e); I_0,h_e,\tau),
\end{equation*}
where $\gamma$ is the teacher's classifier-free guidance scale.
The resulting distillation gradient is:
\begin{equation}
    \grad{\theta}\Loss_\textrm{distill} =
    w_e(\tau)(\hat{v}_\phi^\gamma-\hat{v}_\psi)
    \frac{\partial\hat{x}^s_e}{\partial\theta},
\end{equation}
where the weight $w_e(\tau)$ accounts for the conversion from velocities to clean-output estimates and normalizes the update.
For classifier-free guidance, the teacher's unconditional branch drops the instructions but retains $I_0$.
Following common practice~\cite{yin2024dmd}, the auxiliary editor does not use classifier-free guidance.

\vspace{-6pt}
\paragraph{Training Procedure.}
We alternate between updating the auxiliary editor $v_\psi$ and the student $G_\theta$.
For efficiency, at each training iteration, we uniformly sample a denoising exit step $s$ and average each training loss across all editing turns $e=1,\ldots,E$.
Specifically, we generate a new rollout of $E$ turns for each update of either the student or the auxiliary editor.
When updating the student, we retain gradients through the clean-output estimate $\hat{x}_e^s$ at step $s$ of each turn.
The completed images $I_1,\ldots,I_E$ are passed between turns in latent space, without propagating gradients. 

The clean output estimates $\hat{x}^s_1,\ldots,\hat{x}^s_E$ are used to train the auxiliary editor. 
Specifically, we add noise to these estimates and optimize $v_\psi$ using the flow-matching loss:
\begin{equation*}
    \Loss_{\textrm{FM}}(\psi) =
    \frac{1}{E}\sum_{e=1}^{E}
    \left\|
    v_\psi(\mathbf{z}_\tau(\hat{x}^s_e);I_0,h_e,\tau)
    - (\epsilon-\hat{x}^s_e)
    \right\|_2^2.
\end{equation*}

We first warm up the auxiliary editor for five iterations without updating the student.
Subsequently, we update the student once for every five auxiliary-editor updates, computing the distillation gradients from the retained estimates $\hat{x}_e^s$ and averaging them over all editing turns.
The teacher remains fixed throughout training, while the student and auxiliary editor are updated alternately, with one held fixed while the other is optimized.

\begin{figure*}[t]
  \centering
  \begingroup

  \setlength{\qualimgsize}{0.14\textwidth}
  \setlength{\tabcolsep}{2pt}
  \renewcommand{\arraystretch}{1.0}

  \edef\qualSepOne{\the\dimexpr-4\qualimgsize-16pt\relax}
  \edef\qualSepTwo{\the\dimexpr-2\qualimgsize-8pt\relax}

  \begin{adjustbox}{width=\textwidth}
    \begin{tikzpicture}
      \node[inner sep=0pt, outer sep=0pt] (qualTable) {%
        \begin{tabular}{@{}c cc cc cc@{}}

          {\scriptsize Original}
          & \multicolumn{2}{c}{%
              \qualimg{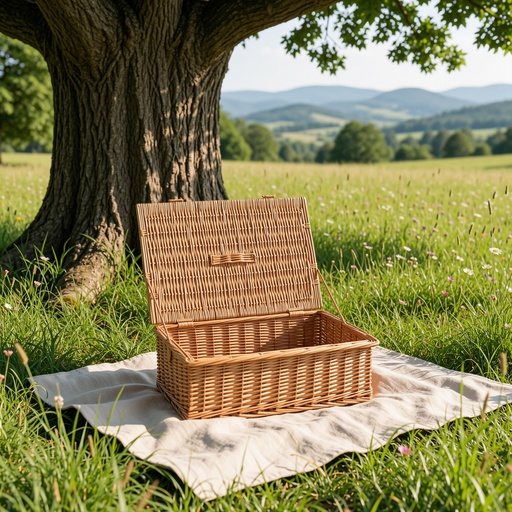}}
          & \multicolumn{2}{c}{%
              \qualimg{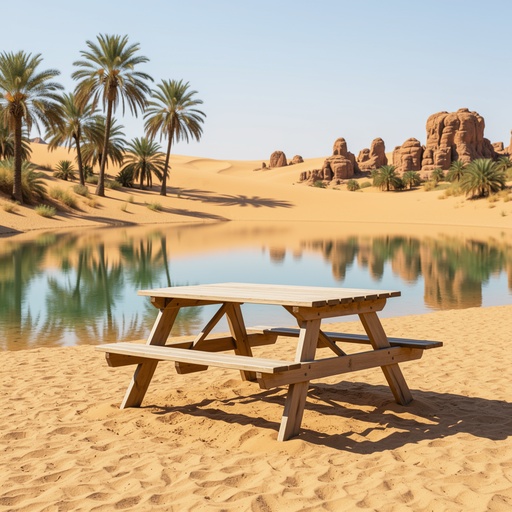}}
          & \multicolumn{2}{c}{%
              \qualimg{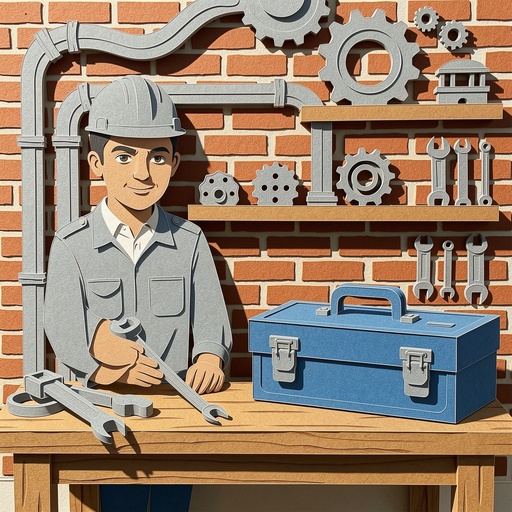}}

          \\[3pt]

          & \qualheader{Flux.2 Klein}
          & \qualheader{+Ours}
          & \qualheader{Flux.2 Klein}
          & \qualheader{+Ours}
          & \qualheader{Flux.2 Klein}
          & \qualheader{+Ours}
          \\[2pt]

          \rotatebox{90}{\scriptsize Edit 5}
          & \qualimg{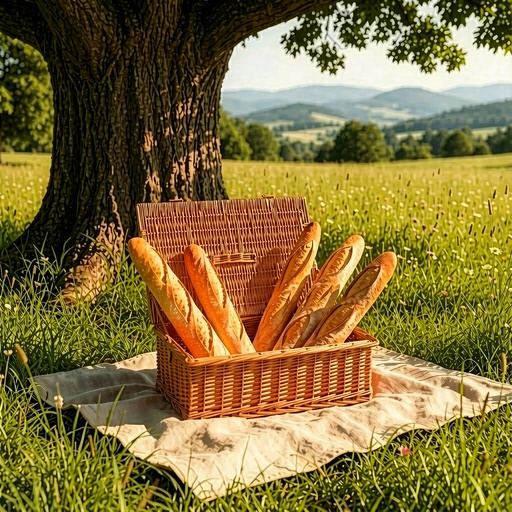}
          & \qualimg{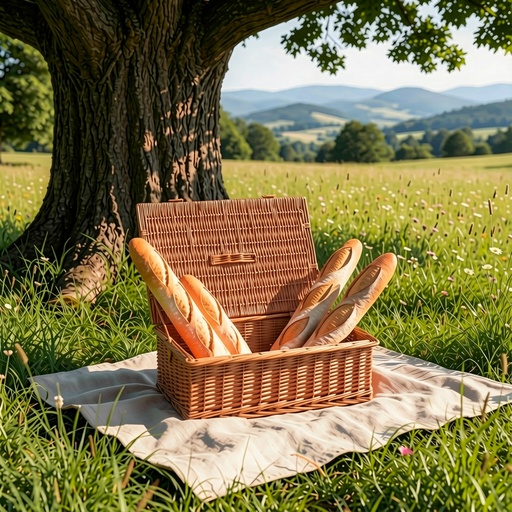}
          & \qualimg{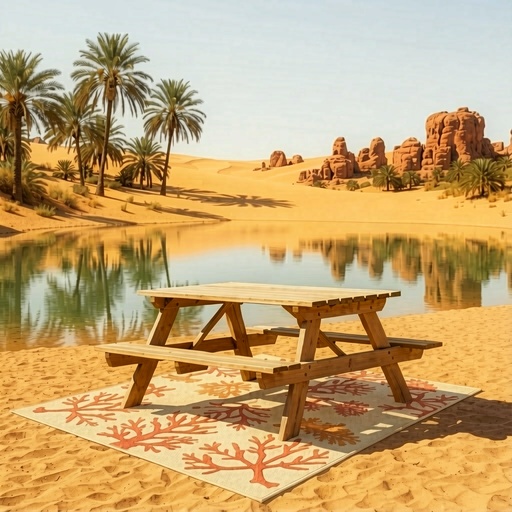}
          & \qualimg{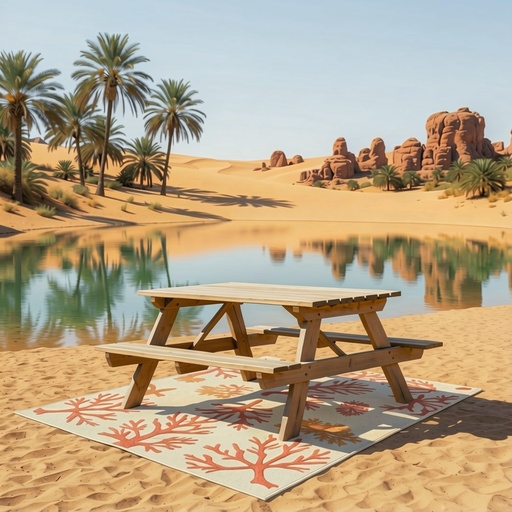}
          & \qualimg{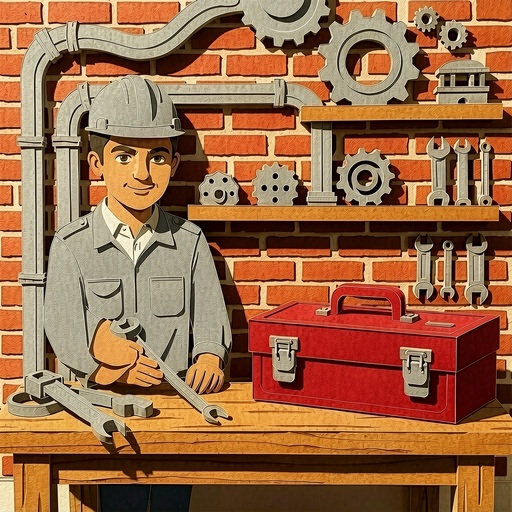}
          & \qualimg{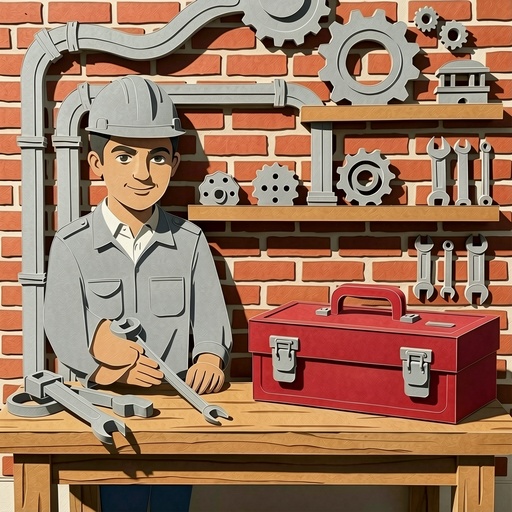}
          \\[3pt]

          & \multicolumn{2}{c}{%
              \qualinstruction{``Add baguettes to the basket''}}
          & \multicolumn{2}{c}{%
              \qualinstruction{``Place a coral patterned rug''}}
          & \multicolumn{2}{c}{%
              \qualinstruction{``Change the toolbox's color to red''}}
          \\[8pt]

          \rotatebox{90}{\scriptsize Edit 10}
          & \qualimg{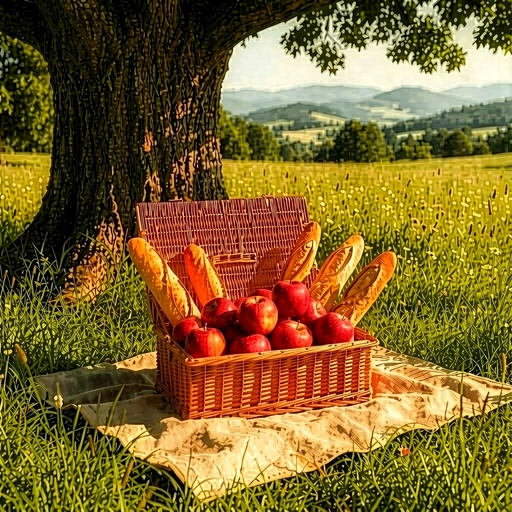}
          & \qualimg{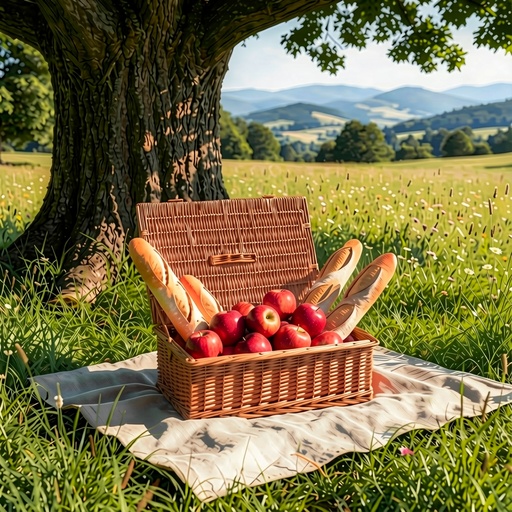}
          & \qualimg{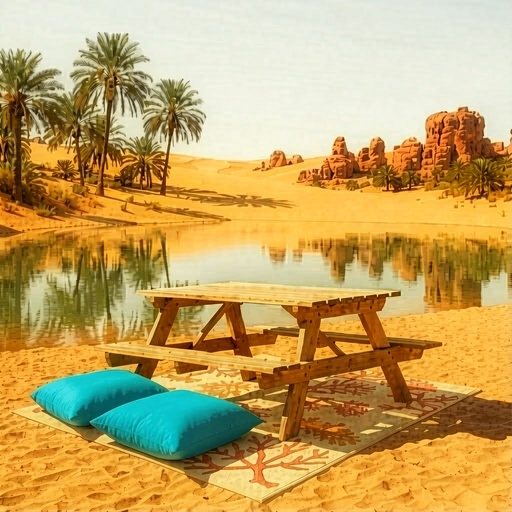}
          & \qualimg{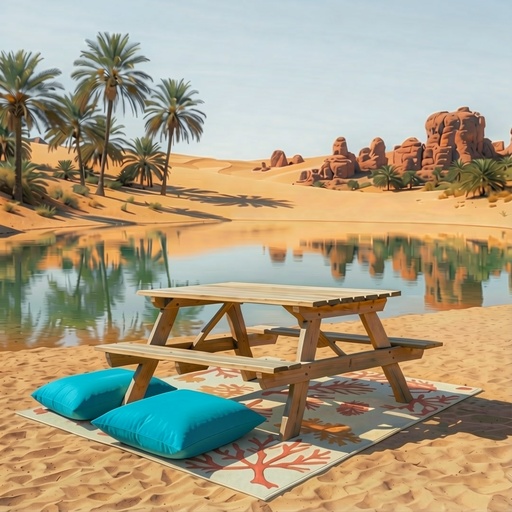}
          & \qualimg{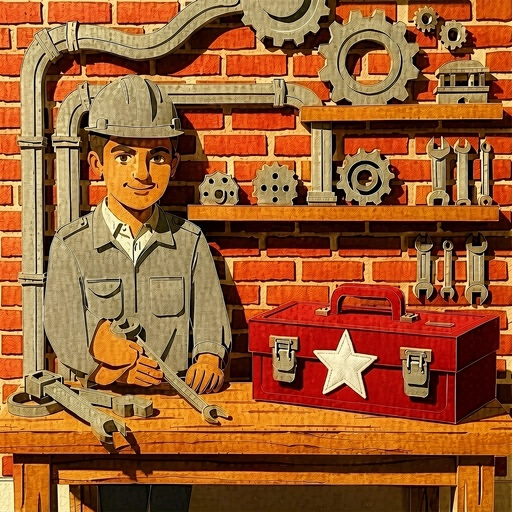}
          & \qualimg{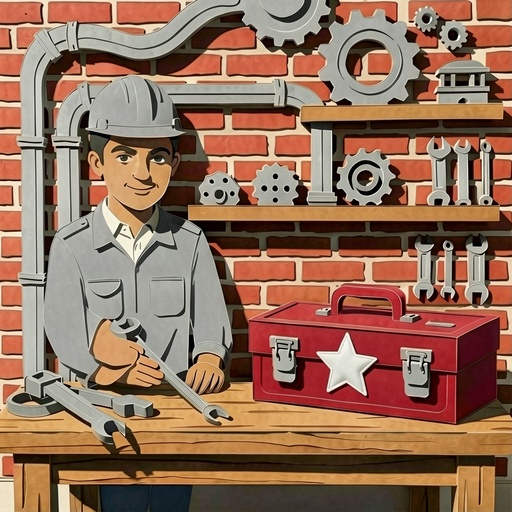}
          \\[3pt]

          & \multicolumn{2}{c}{%
              \qualinstruction{``Add apples''}}
          & \multicolumn{2}{c}{%
              \qualinstruction{``Add two turquoise cushions on the rug''}}
          & \multicolumn{2}{c}{%
              \qualinstruction{``Add a white star patch on the toolbox''}}
          \\[8pt]

          \rotatebox{90}{\scriptsize Edit 15}
          & \qualimg{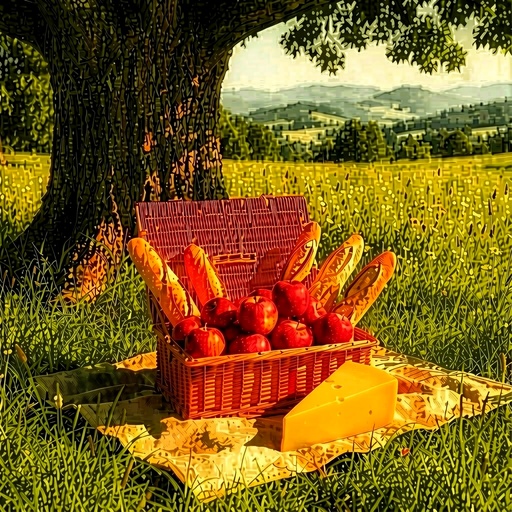}
          & \qualimg{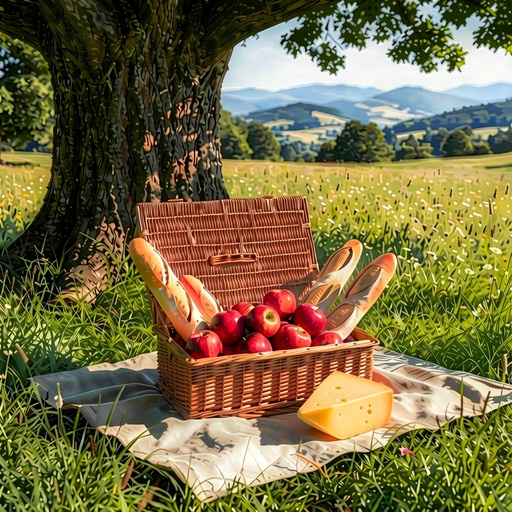}
          & \qualimg{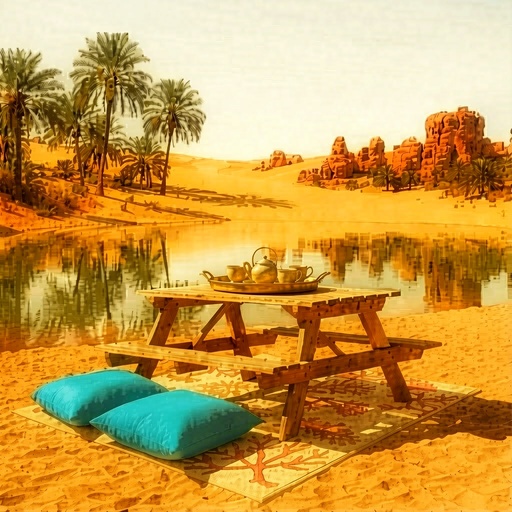}
          & \qualimg{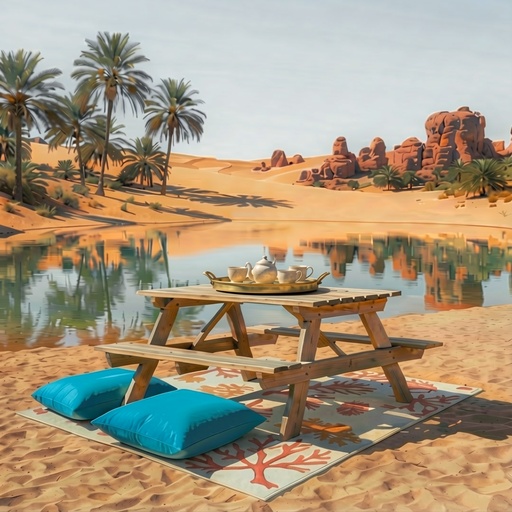}
          & \qualimg{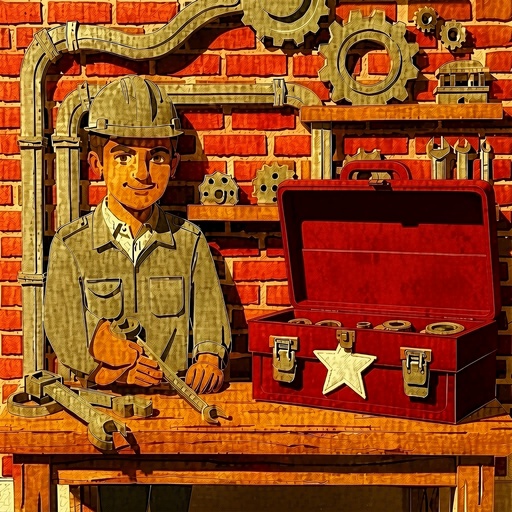}
          & \qualimg{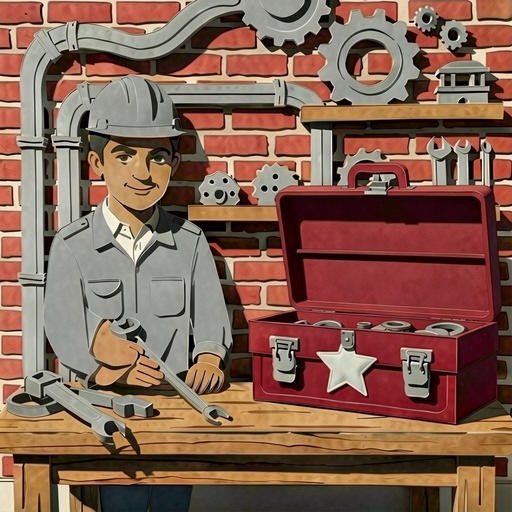}
          \\[3pt]

          & \multicolumn{2}{c}{%
              \qualinstruction{``Add a wedge of cheese''}}
          & \multicolumn{2}{c}{%
              \qualinstruction{``Place a tea tray on the table''}}
          & \multicolumn{2}{c}{%
              \qualinstruction{``Open the toolbox's lid''}}
          \\[8pt]

          \rotatebox{90}{\scriptsize Edit 20}
          & \qualimg{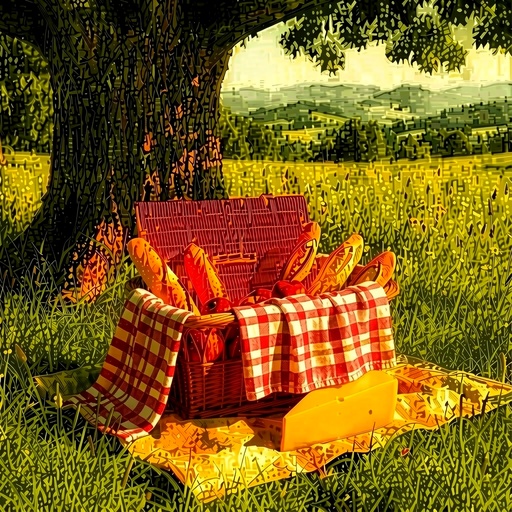}
          & \qualimg{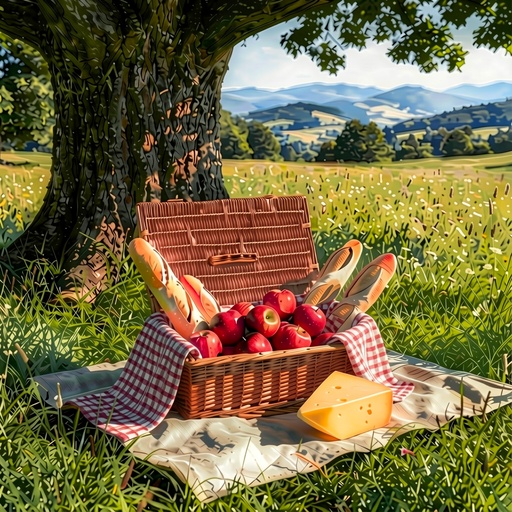}
          & \qualimg{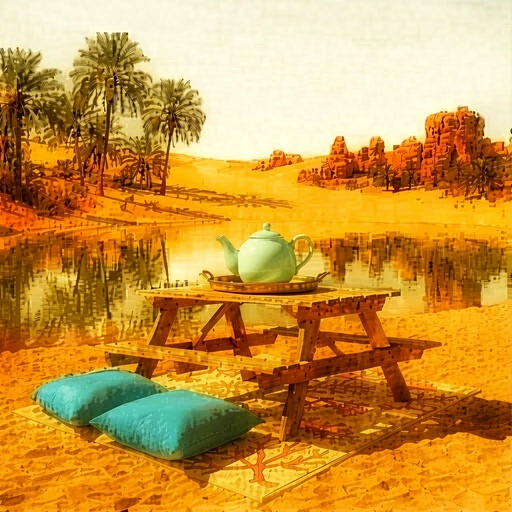}
          & \qualimg{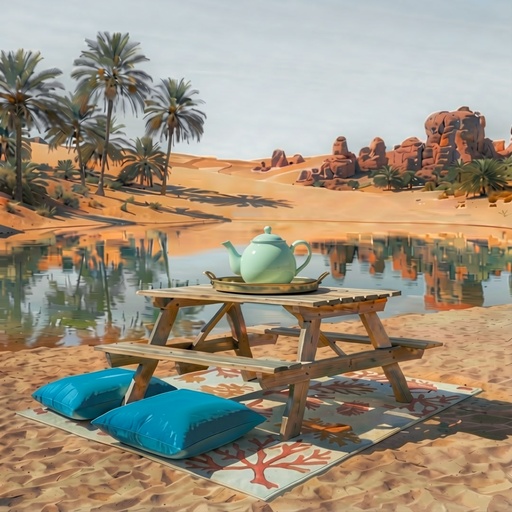}
          & \qualimg{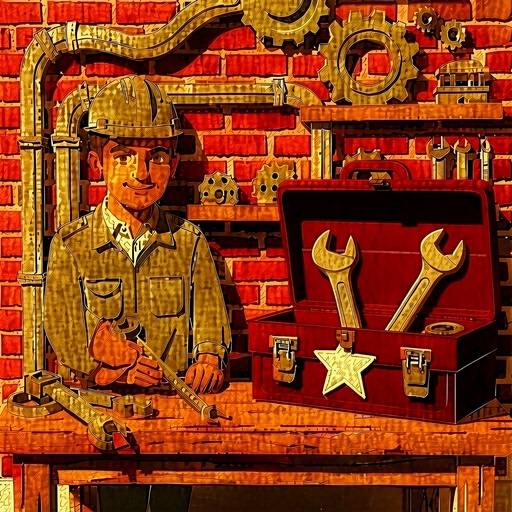}
          & \qualimg{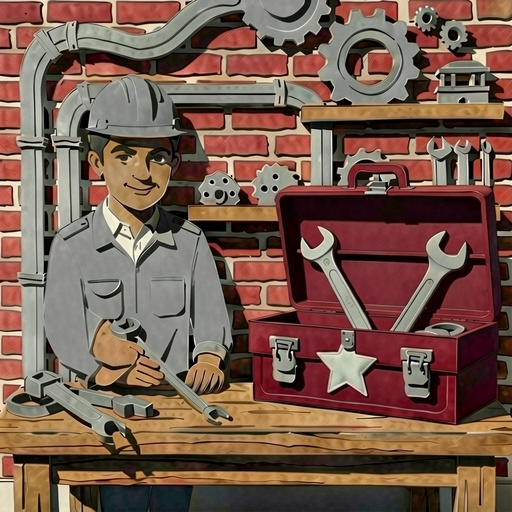}
          \\[3pt]

          & \multicolumn{2}{c}{%
              \qualinstruction{``Drape a checkered napkin over the basket.''}}
          & \multicolumn{2}{c}{%
              \qualinstruction{``Place a mint-green teapot on the tray''}}
          & \multicolumn{2}{c}{%
              \qualinstruction{``Place two wrenches inside''}}
          \\[3pt]

        \end{tabular}%
      };

      \draw[
        gray,
        line width=0.4pt,
        dash pattern=on 2pt off 2pt
      ]
        ([xshift=\qualSepOne]qualTable.north east)
        --
        ([xshift=\qualSepOne]qualTable.south east);

      \draw[
        gray,
        line width=0.4pt,
        dash pattern=on 2pt off 2pt
      ]
        ([xshift=\qualSepTwo]qualTable.north east)
        --
        ([xshift=\qualSepTwo]qualTable.south east);

    \end{tikzpicture}
  \end{adjustbox}

  \endgroup

\vspace{-8pt}
  \caption{%
    \textbf{Multi-turn image editing.}
      We show selected turns from 20-step editing sequences.
      Each displayed instruction is followed by four no-op instructions
      (``keep the image as is''), and the images show the results after
      those no-ops.
      While the original FLUX.2 Klein accumulates color shifts and loses
      detail, fine-tuning it with our training paradigm (+Ours) yields
      a more stable editing process, preserving the scene and prior edits
      while following the requested changes.
      }
\vspace{-8pt}
    \label{fig:qualitative_results}
\end{figure*}

\section{Experiments}
\label{sec:experiments}
\begin{figure*}[t]
  \centering
  \begingroup

  \def\gardir{images/garage}

  \setlength{\tabcolsep}{1pt}
  \renewcommand{\arraystretch}{1.0}

  \newcommand{\garimg}[1]{%
    \raisebox{-0.5\height}{%
      \includegraphics[width=0.185\textwidth]{#1}%
    }%
  }
  \newcommand{\garlabel}[1]{%
  \raisebox{-0.5\height}{%
    \rotatebox{90}{\scriptsize #1}%
  }%
}
  \newcommand{\garhead}[1]{\makebox[0.185\textwidth]{\scriptsize #1}}
  \newcommand{\garturnrow}[2]{%
    \garlabel{#1}
    & \garimg{\gardir/ours/#2}
    & \garimg{\gardir/emu/#2}
    & \garimg{\gardir/lfa/#2}
    & \garimg{\gardir/freqedit/#2}
    & \garimg{\gardir/mtopsd/#2}
  }

  \begin{adjustbox}{width=\textwidth}
    \begin{tabular}{@{}c *{5}{c}@{}}

      & \garhead{\textbf{Ours}}
      & \garhead{Emu Edit}
      & \garhead{VAE-LFA}
      & \garhead{FreqEdit}
      & \garhead{MT-OPSD}
      \\[2pt]

      \garlabel{Source}
      & \garimg{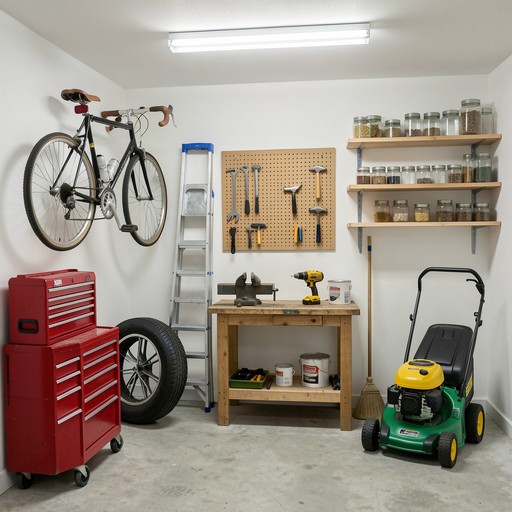}
      & \garimg{\gardir/source.jpg}
      & \garimg{\gardir/source.jpg}
      & \garimg{\gardir/source.jpg}
      & \garimg{\gardir/source.jpg}
      \\ \noalign{\vskip 2pt}
      \garturnrow{Edit 5}{edit_05.jpg} \\ \noalign{\vskip 2pt}
      \garturnrow{Edit 10}{edit_10.jpg} \\

    \end{tabular}
  \end{adjustbox}

  \endgroup

  \caption{%
    \textbf{Ten successive removals.}
    Each turn removes one object from a synthetic garage scene: the drill, broom, car tire, ladder and bicycle (turns 1--5), then the lawn mower, vise, toolbox, pegboard and shelves (turns 6--10).
    By the last turn, every other method has visibly damaged the scene, while our model leaves the walls, the floor and the workbench as they were.
    Emu Edit leaves pale, blocky patches on the wall, VAE-LFA darkens the image under a coarse, mottled texture, FreqEdit casts the walls and floor yellow, and MT-OPSD washes the room out into a blurred, almost white space.
  }
  \label{fig:garage_removal}
\end{figure*}

\subsection{Experimental Setup}
\label{sec:exp:setup}

\paragraph{Implementation details.}
Our student is FLUX.2-klein-9B~\cite{flux-2-2025}, a 4-step distilled model, which we fine-tune with a LoRA~\cite{hu2022lora} of rank 16 (27.8M trainable parameters, 0.3\% of the model).
The teacher and the auxiliary editor are both FLUX.2-klein-base-9B, the undistilled model of the same family.
The teacher is frozen and uses classifier-free guidance with scale 2.
The auxiliary editor trains its own LoRA of the same size as the student and is not guided.
The student, being distilled, runs without classifier-free guidance, with 4 sampling steps in training rollouts as at inference.
Each training rollout is a chain of 10 edits, and the student's output latent is passed directly to the next edit, without going through the VAE.
The noise level $\tau$ is drawn uniformly over $[0,1)$, independently for every turn, both for the student's loss and for training the auxiliary editor.

The student and the auxiliary editor are optimized with AdamW~\cite{loshchilov2019adamw} ($\beta=(0, 0.999)$, weight decay 0.01, gradient clipping at 10, bf16), with learning rates of $4\times10^{-5}$ for the student and $8\times10^{-6}$ for the auxiliary editor.
We train for 505 iterations: 5 warm-up iterations that update only the auxiliary editor, followed by 500 iterations in which the auxiliary editor is updated at every iteration and the student at every fifth, i.e., 100 student updates.
Training takes about 4.4 GPU-hours on a single NVIDIA B200.
At inference we use the same 4-step unguided sampler.

\vspace{-8pt}
\paragraph{Training data.}
The inputs for the training rollouts are 96 synthetic $1024\times1024$ images generated with FLUX.2-klein-9B from short scene descriptions. %
Each image comes with one cumulative chain of 10 edit instructions (960 in total).
All edits are local: each adds, removes, replaces or resizes an object, or changes one of its attributes, such as its color.
Our evaluation benchmark, EdiVal-10 (Sec.~\ref{sec:exp:benchmark}), consists of real photographs, so no image is shared between training and evaluation.

\vspace{-8pt}
\paragraph{Baselines.}
We compare our stable editor with its pretrained backbone model, training-free strategies for reducing multi-turn drift, and a trained multi-turn editor.
Our backbone model baseline is FLUX.2-klein-9B without our fine-tuning, using the same 4-step sampler as our student.
We evaluate Emu Edit's pixel-thresholding strategy~\cite{sheynin2024emu}, VAE-LFA~\cite{wang2026vaelfa}, and FreqEdit~\cite{liao2026freqEdit} using the same backbone model.
For Emu Edit, we follow MT-OPSD's implementation and settings (blur $\sigma=3$, threshold $\alpha=0.03$)~\cite{zhao2026mtopsd}.
For VAE-LFA, we follow its official FLUX.2-klein-9B code, which differs from the paper's description.
To match our student's previous-image-only input setting, we port FreqEdit's official code for the variant that conditions on the previous turn, without the original-source quality guidance of its Kontext version, and use its Kontext settings.
Our model, its backbone model and the training-free methods sample each chain with the same fixed seed, so that their differences come from the method rather than from sampling noise.
We also compare with MT-OPSD~\cite{zhao2026mtopsd}, using its released LoRA on FLUX.2-klein-base-9B, and include the same backbone without this LoRA as its corresponding base-model baseline.
Both use the authors' inference script and its fixed seed, at 512-pixel resolution, with 30 sampling steps and guidance scale 4; their outputs are upsampled to $1024\times1024$ for evaluation.
We fix the seed per chain across all model evaluations.

\subsection{Qualitative Evaluation}
\label{sec:exp:qualitative}

\begin{table*}[t]
\centering
\caption{\textbf{Quantitative comparison on EdiVal-10.}
All metrics are computed after 10 cumulative edits and averaged over the 252 chains, except the anchor distance, which is averaged over the 148 chains whose anchor object lies mostly outside the edited objects (Appendix~\ref{app:metrics}).
Quality is measured by the KRIS-Bench image-quality score of the final image, assigned by a VLM judge (1--5), and by the change in HPSv3 from turn 1 to turn 10.
Preservation is measured by the KRIS-Bench consistency score assigned by the same judge on the whole image (1--5; higher means fewer unrequested changes), the DINOv2 distance between the source and the final image on unedited anchor objects, and the LPIPS distance on the background outside all edited objects.
Editing is measured by the KRIS-Bench instruction-following score at turn 10 (1--5); it checks that stability does not come from skipping edits.
Methods are grouped by backbone; the best result in each column is shown in bold and the second best is underlined.
$^*$: significantly different from our model (more than two standard errors, paired over chains).}
\label{tab:comparison}
\newcommand{\tabhdr}[1]{\begin{tabular}[b]{@{}c@{}}#1\end{tabular}}%
\small
\setlength{\tabcolsep}{4pt}
\begin{tabular*}{\textwidth}{@{\extracolsep{\fill}}lcccccc}
\toprule
 & \multicolumn{2}{c}{Quality} & \multicolumn{3}{c}{Preservation} & Editing \\
\cmidrule(lr){2-3} \cmidrule(lr){4-6} \cmidrule(lr){7-7}
Method & \tabhdr{Quality\\(KRIS, 1--5) $\uparrow$} & \tabhdr{$\Delta$HPSv3\\(turn 1$\to$10) $\uparrow$} & \tabhdr{Consistency\\(KRIS, 1--5) $\uparrow$} & \tabhdr{Anchor dist.\\(DINOv2) $\downarrow$} & \tabhdr{Background\\dist. (LPIPS) $\downarrow$} & \tabhdr{Instruction\\following\\(KRIS, 1--5) $\uparrow$} \\
\midrule
\multicolumn{7}{l}{\textit{FLUX.2-klein-base-9B (undistilled)}} \\
Base model                         & 1.42$^*$ & $-3.17^*$ & 1.08$^*$ & 0.660$^*$ & 0.531$^*$ & 3.10$^*$ \\
MT-OPSD~\cite{zhao2026mtopsd}      & 1.95$^*$ & $-2.54^*$ & 1.29$^*$ & 0.505$^*$ & 0.455$^*$ & 3.16$^*$ \\
\midrule
\multicolumn{7}{l}{\textit{FLUX.2-klein-9B (4-step distilled)}} \\
Base model                         & 2.73$^*$ & $-0.78^*$ & 3.03$^*$ & 0.314$^*$ & 0.351$^*$ & 3.31 \\
+ Emu Edit~\cite{sheynin2024emu}   & \underline{2.90}$^*$ & $\mathbf{-0.15}$ & \underline{3.32}$^*$ & \textbf{0.196} & \underline{0.264}$^*$ & 3.29 \\
+ VAE-LFA~\cite{wang2026vaelfa}    & 2.80$^*$ & $-0.89^*$ & 3.04$^*$ & 0.289$^*$ & 0.313$^*$ & 3.32 \\
+ FreqEdit~\cite{liao2026freqEdit} & 2.83$^*$ & $-0.82^*$ & 3.18$^*$ & 0.283$^*$ & 0.331$^*$ & \textbf{3.39} \\
Ours                               & \textbf{3.19} & $\underline{-0.17}$ & \textbf{3.85} & \underline{0.203} & \textbf{0.254} & \underline{3.38} \\
\bottomrule
\end{tabular*}
\end{table*}

\begin{figure*}[t]
  \centering
  \begingroup

  \def\abldir{images/ablations/playroom}

  \setlength{\tabcolsep}{1.5pt}
  \renewcommand{\arraystretch}{1.0}

  \newcommand{\ablimg}[1]{%
    \raisebox{-0.5\height}{%
      \includegraphics[width=0.18\textwidth]{#1}%
    }%
  }
  \newcommand{\ablinstruction}[1]{%
    \parbox[t]{0.18\textwidth}{%
      \centering\scriptsize #1%
    }%
  }
  \newcommand{\abllabel}[1]{%
  \raisebox{-0.5\height}{%
    \rotatebox{90}{\scriptsize\shortstack[c]{#1}}%
  }%
}
  \newcommand{\ablturn}[1]{{\scriptsize #1}}
  \newcommand{\ablrow}[2]{%
    \abllabel{#1}
    & \ablimg{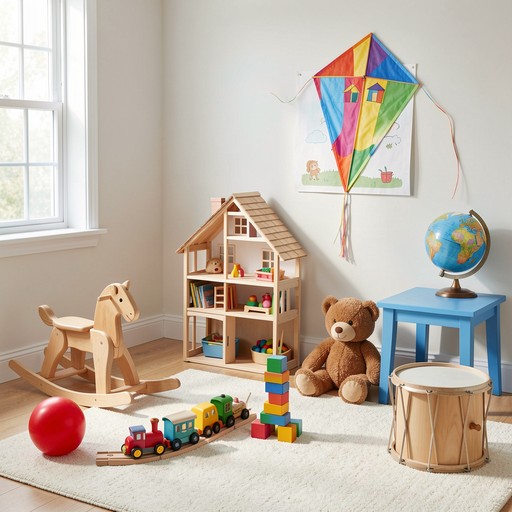}
    & \ablimg{\abldir/#2/edit_03.jpg}
    & \ablimg{\abldir/#2/edit_06.jpg}
    & \ablimg{\abldir/#2/edit_08.jpg}
    & \ablimg{\abldir/#2/edit_09.jpg}
  }

  \begin{adjustbox}{width=\textwidth}
    \begin{tabular}{@{}c *{5}{c}@{}}

      & \ablturn{Original}
      & \ablturn{Edit 3}
      & \ablturn{Edit 6}
      & \ablturn{Edit 8}
      & \ablturn{Edit 9}
      \\[3pt]

      \ablrow{Base}{base} \\ \noalign{\vskip 4pt}
      \ablrow{\textbf{Ours}}{ours} \\ \noalign{\vskip 4pt}
      \ablrow{Teacher sees\\previous image}{prevref} \\ \noalign{\vskip 4pt}
      \ablrow{Student sees\\source image}{anchor} \\ \noalign{\vskip 4pt}

      &
      & \ablinstruction{``Change the color of the blue stool to yellow''}
      & \ablinstruction{``Change the color of the red ball to green''}
      & \ablinstruction{``Replace the globe on the stool with a small potted cactus''}
      & \ablinstruction{``Make the white teddy bear look like it is made of knitted wool''}
      \\

    \end{tabular}
  \end{adjustbox}

  \endgroup
\vspace{-8pt}
  \caption{%
    \textbf{Conditioning ablations.}
    Turns 3, 6, 8 and 9 of a ten-turn editing sequence on a source scene generated with FLUX.2-klein-9B independently of the training data; the instruction below each column is the one applied at that turn.
    Our model changes only what each instruction names, and the image stays clean.
    The base model degrades the image: it turns yellow and washed out over the turns, and the knitted texture requested for the teddy bear at turn 9 also covers the rocking horse.
    Training with a teacher that also sees the previous image does not prevent this degradation: blotchy noise spreads over the walls and the rug from turn to turn.
    When the student also sees the source image, earlier edits are undone: the stool made yellow at turn 3 is blue again by turn 6, and the ball made green at turn 6 is red again by turn 8.
  }
  \vspace{-8pt}
  \label{fig:ablation_conditioning}
\end{figure*}

Figures~\ref{fig:teaser} and~\ref{fig:qualitative_results} compare our model with its backbone model on editing sequences interspersed with no-op instructions.
We further compare against the training-free methods and MT-OPSD on sequences of consecutive semantic edits in Figure~\ref{fig:garage_removal}.
The source scenes for these comparisons were generated independently of the training data using FLUX.2-klein-9B.
Figure~\ref{fig:garage_removal} demonstrates that degradation can accumulate even under simple editing instructions: after ten successive object removals, all competing methods visibly damage the surrounding scene, while our model successfully preserves it.

Each training-free method ties every output to part of the previous image, and this is also its main weakness: when an edit has to change exactly that part, content from earlier turns leaks into the result.
Emu Edit reverts pixels that change only slightly, so remnants of earlier images survive; VAE-LFA aligns low-frequency latent statistics with earlier outputs, so their colours persist; and FreqEdit steers high-frequency detail toward the previous image, so their textures persist.
Figure~\ref{fig:teddy_backgrounds} in Appendix~\ref{app:backgrounds} illustrates these effects over three successive background replacements. MT-OPSD leaves no such traces but blurs the image and washes out its colours, consistent with the degradation observed in Figure~\ref{fig:garage_removal}.

Further comparisons with the backbone model, including a full sequence of fifteen edits, are shown in Figs.~\ref{fig:qualitative_results_supp} and~\ref{fig:consecutive_edits} in the appendix.

\subsection{Quantitative Evaluation}
\label{sec:exp:comparison}

\paragraph{Evaluation benchmark.}
\label{sec:exp:benchmark}
Released multi-turn editing benchmarks are short: EdiVal~\cite{chen2026edival} chains three edits, MagicBrush~\cite{zhang2023magicbrush} at most three, and MSE-Bench~\cite{qu2026vincie} five, too few for drift to build up.
Omni IIE Bench~\cite{yang2026omniiie} advertises editing sequences of up to 16 turns, but they average 4.35 turns, and only 9 of its 260 reach ten.
The one benchmark we know of that is built around ten turns, LME-Bench~\cite{zhao2026mtopsd}, contains 100 sessions on synthetic images, and each session opens with a global edit (an art style, a black-and-white or sepia grade, or a change of weather or time of day) and includes three more later. After such a turn, nothing in the image is meant to stay as it was, so unintended changes can no longer be distinguished from those required by the edit through comparison with the source.
We therefore introduce EdiVal-10, in which we extend EdiVal to ten turns of local edits, and report results on LME-Bench in Appendix~\ref{app:lme}.
It contains 252 chains of 10 cumulative edits on real photographs from GEdit-Bench~\cite{liu2025step1xedit}, center-cropped and resized to $1024\times1024$.
Following EdiVal, a language model lists the objects in each photo using EdiVal's object-listing prompt, and GroundingDINO~\cite{liu2024groundingdino} grounds them.
Every edit targets a grounded object and belongs to one of eight local types: add, remove, replace, color, material, text, count, and position.
Each chain also reserves one or two \emph{anchor} objects that no instruction may edit or mention.
The language model then writes the instructions from a per-chain plan of objects and edit types.
They are checked by rule-based validators and by the language model acting as a critic, and rejected turns are sent back for rewriting.
Every language-model step of its construction uses Claude Sonnet 5.5~\cite{anthropic2026sonnet55}.

\paragraph{Metrics.}
We score the final image of each chain (turn 10) on three aspects and average over chains.
\emph{Quality}: a VLM judge (Gemma-4-31B~\cite{gemma2026gemma4}) rates the image with the image-quality prompt of KRIS-Bench~\cite{wu2025krisbench}, and $\Delta$HPSv3~\cite{ma2025hpsv3} measures how much a learned quality estimate changes from turn 1 to turn 10.
\emph{Preservation}: the same judge rates, with the KRIS-Bench consistency prompt, how well everything the instructions do not ask to change is kept; locally, we measure the DINOv2~\cite{oquab2024dinov2} distance to the source on the unedited anchor objects and the LPIPS~\cite{zhang2018lpips} distance on the background outside all edited objects.
\emph{Editing}: the judge's KRIS-Bench instruction-following score checks that the requested edits are carried out, so that preservation cannot come from skipping them.
Prompts, masks and the judge's controls are detailed in Appendix~\ref{app:metrics}.

Table~\ref{tab:comparison} reports quality, preservation and editing after ten cumulative edits.
To compare a method with ours, we take the difference between the two on each chain and report the mean difference $\pm$ its standard error over the chains; we call a difference significant when it is more than twice its standard error.

\paragraph{Quality.}
Our method produces the highest-quality final images: its KRIS quality score is significantly higher than that of every other method, the smallest margin being over Emu Edit ($0.30 \pm 0.08$). This result is aligned with the qualitative results in Figure~\ref{fig:garage_removal}.
Emu Edit's $\Delta$HPSv3 does not differ significantly from ours ($-0.15$ vs.\ $-0.17$), and its visual artifacts are demonstrated in Figure~\ref{fig:garage_removal}.

\paragraph{Preservation of unedited content.}
Judged on the whole image, our method makes the fewest unintended changes: its KRIS consistency score is 3.85, compared with 3.04 to 3.32 for the training-free methods, 3.03 for the base model and 1.29 for MT-OPSD.
On the background, our method attains the lowest LPIPS distance (0.254), a massive improvement over the base model at 0.351. Compared to our method, every other method is worse, Emu Edit by the smallest margin (0.264; $+0.010 \pm 0.005$, $t = 2.1$).
On the anchor objects, our method and Emu Edit attain the lowest DINOv2 distances (0.203 and 0.196; the difference, $-0.007 \pm 0.012$, is not significant), while all other methods are significantly worse than ours.
Emu Edit is comparable to our method only on two local metrics, not on whole-image consistency; our model matches or exceeds it on both without copying pixels.

\begin{table*}[t]
\centering
\caption{\textbf{Ablations on EdiVal-10.}
Each row changes one aspect of our model; all variants are trained for 100 student updates and evaluated after 10 cumulative edits on all 252 chains (the anchor distance on the 148 chains whose anchor object lies mostly outside the edited objects; Appendix~\ref{app:metrics}).
$^*$: significantly different from our model (more than two standard errors, paired over chains).}
\label{tab:ablations_full}
\newcommand{\tabhdr}[1]{\begin{tabular}[b]{@{}c@{}}#1\end{tabular}}%
\newcommand{\tablab}[1]{\begin{tabular}[c]{@{}l@{}}#1\end{tabular}}%
\small
\setlength{\tabcolsep}{4pt}
\begin{tabular*}{\textwidth}{@{\extracolsep{\fill}}lcccccc}
\toprule
 & \multicolumn{2}{c}{Quality} & \multicolumn{3}{c}{Preservation} & Editing \\
\cmidrule(lr){2-3} \cmidrule(lr){4-6} \cmidrule(lr){7-7}
Variant & \tabhdr{Quality\\(KRIS, 1--5) $\uparrow$} & \tabhdr{$\Delta$HPSv3\\(turn 1$\to$10) $\uparrow$} & \tabhdr{Consistency\\(KRIS, 1--5) $\uparrow$} & \tabhdr{Anchor dist.\\(DINOv2) $\downarrow$} & \tabhdr{Background\\dist. (LPIPS) $\downarrow$} & \tabhdr{Instruction\\following\\(KRIS, 1--5) $\uparrow$} \\
\midrule
\textbf{Ours} & 3.19 & $-0.17$ & 3.85 & 0.203 & 0.254 & 3.38 \\
\midrule
\tablab{Teacher sees\\previous image} & 2.75$^*$ & $-0.86^*$ & 2.81$^*$ & 0.307$^*$ & 0.366$^*$ & 3.32 \\
\tablab{Teacher sees only previous\\image and instruction} & 2.71$^*$ & $-1.12^*$ & 2.28$^*$ & 0.366$^*$ & 0.381$^*$ & 3.23$^*$ \\
\tablab{Student sees\\source image} & 3.46$^*$ & $-0.08$ & 4.08$^*$ & 0.188 & 0.230$^*$ & 2.80$^*$ \\
\tablab{5-edit training\\rollouts (vs.\ 10)} & 3.23 & $-0.28^*$ & 3.29$^*$ & 0.265$^*$ & 0.334$^*$ & 3.25$^*$ \\
\bottomrule
\end{tabular*}
\end{table*}

\paragraph{Editing.}
The gains in quality and preservation do not come at the expense of instruction following, with instruction-following scores remaining comparable.
On the KRIS-Bench instruction-following score, our method (3.38) does not differ significantly from the highest-scoring method, FreqEdit (3.39), and scores significantly higher than MT-OPSD (3.16) and its base model (3.10).

\paragraph{Single edits.}
Stability over many turns should not come at the cost of single edits.
We therefore also apply each method once to each of the 606 English instructions of GEdit-Bench~\cite{liu2025step1xedit} and measure the change in HPSv3 from the source to the edited image (Appendix~\ref{app:single}).
A single edit lowers HPSv3 for every method, and least for ours: by 0.18, against 0.41 for the base model ($0.23 \pm 0.02$), while every other method loses significantly more.

\subsection{Ablation Study}
\label{sec:exp:ablations}

We evaluate the role of asymmetric conditioning on EdiVal-10 by varying the inputs provided to the teacher, auxiliary editor, and student, and the length of the training rollouts.
The teacher and auxiliary editor receive identical conditioning in all variants.
All variants are trained for 100 student updates.
Table~\ref{tab:ablations_full} reports the quantitative results, and Figure~\ref{fig:ablation_conditioning} illustrates the conditioning variants qualitatively.

\paragraph{Teacher and auxiliary conditioning.}
Our teacher and auxiliary editor are conditioned on the original image and the compound instruction, anchoring score distillation to the clean source.
We compare this design with two alternatives: additionally providing the previous student output to both models, and replacing the original image and compound instruction with the previous output and current instruction.
Both alternatives substantially reduce preservation: consistency drops from 3.85 to 2.81 and 2.28, respectively, compared with 3.03 for the pretrained base model.
Figure~\ref{fig:ablation_conditioning} illustrates the former variant: as with the base model, the image degrades over the turns, and blotchy noise spreads over the walls and the rug.
These results suggest that conditioning the score models on intermediate student outputs can reinforce accumulated artifacts rather than correct them.

\paragraph{Student conditioning.}
Our student receives only the previous output and current instruction.
Providing the original image as an additional reference improves consistency from 3.85 to 4.08, but reduces instruction following from 3.38 to 2.80.
As shown in Figure~\ref{fig:ablation_conditioning}, earlier edits are undone as later ones are applied: the recoloured stool and ball return to their original colours.
Thus, stronger preservation of the source does not necessarily imply successful multi-turn editing.
Our asymmetric design maintains both scene consistency and the changes introduced across successive turns.

\paragraph{Rollout length.}
We train on rollouts of 10 edits, the same length as the evaluated sequences.
A student trained on 5-edit rollouts still improves over the base model after ten turns (consistency 3.29 vs.\ 3.03), but recovers only about a third of our gain (3.85) and is significantly worse than our model on every preservation measure.
Training on its own outputs over the full length of the sequences it edits at inference is therefore important.

\section{Conclusions, Limitations and Future Work}
\label{sec:conclusion}

\begin{figure}[h]
  \centering
  \begingroup

  \def\rolloutdir{images/limitations/hiker}

  \setlength{\tabcolsep}{1.5pt}
  \renewcommand{\arraystretch}{1.0}

  \newcommand{\limimg}[1]{%
    \raisebox{-0.5\height}{%
      \includegraphics[width=0.215\linewidth]{#1}%
    }%
  }
  \newcommand{\liminstruction}[1]{%
    \parbox[t]{0.215\linewidth}{%
      \centering\scriptsize #1%
    }%
  }
  \newcommand{\limlabel}[1]{%
  \raisebox{-0.5\height}{%
    \rotatebox{90}{\scriptsize\shortstack[c]{#1}}%
  }%
}
  \newcommand{\limturn}[1]{{\scriptsize #1}}

  \begin{adjustbox}{width=\linewidth}
    \begin{tabular}{@{}c *{4}{c}@{}}

      & \limturn{Original}
      & \limturn{Edit 5}
      & \limturn{Edit 8}
      & \limturn{Edit 9}
      \\[3pt]

      \limlabel{FLUX.2 Klein}
      & \limimg{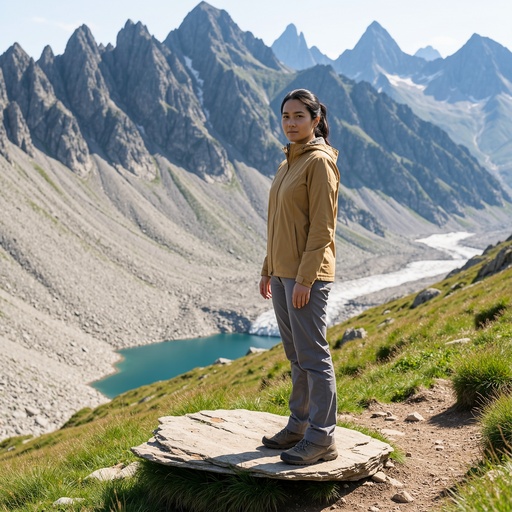}
      & \limimg{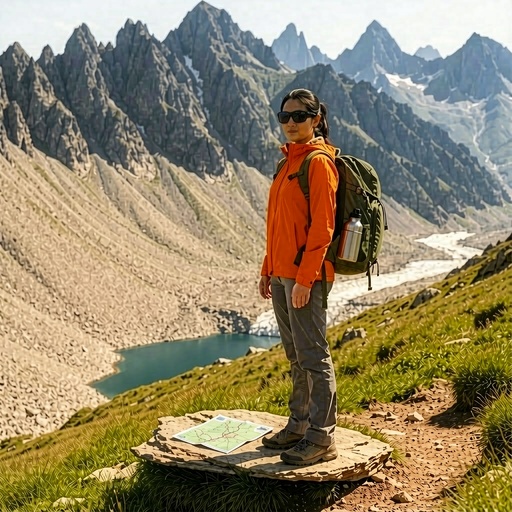}
      & \limimg{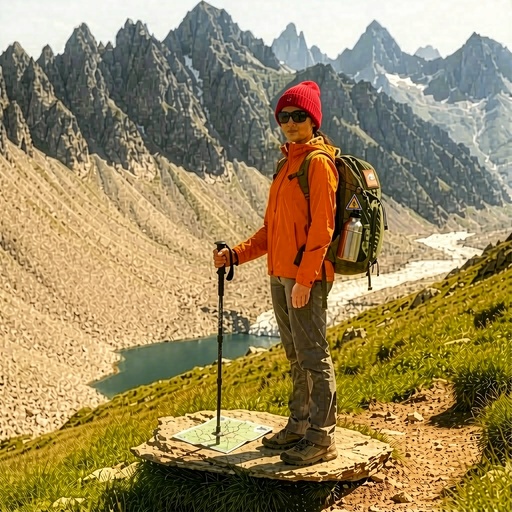}
      & \limimg{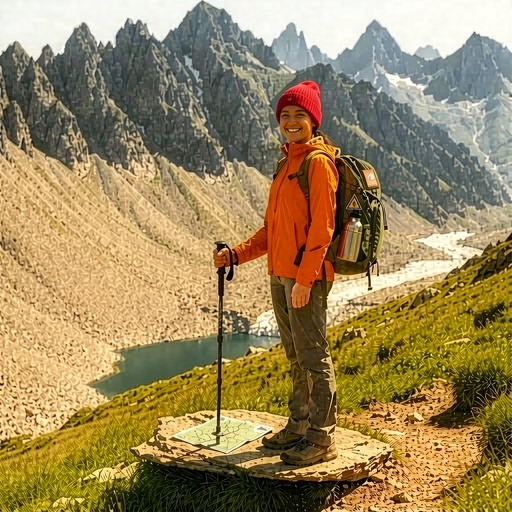}
      \\
      \noalign{\vskip 4pt}

      \limlabel{+Ours}
      & \limimg{\rolloutdir/original.jpg}
      & \limimg{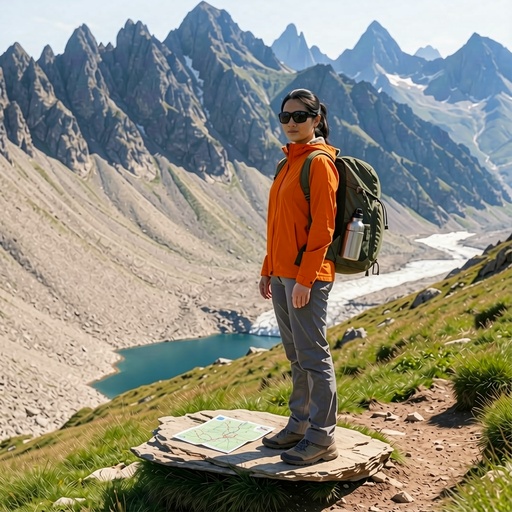}
      & \limimg{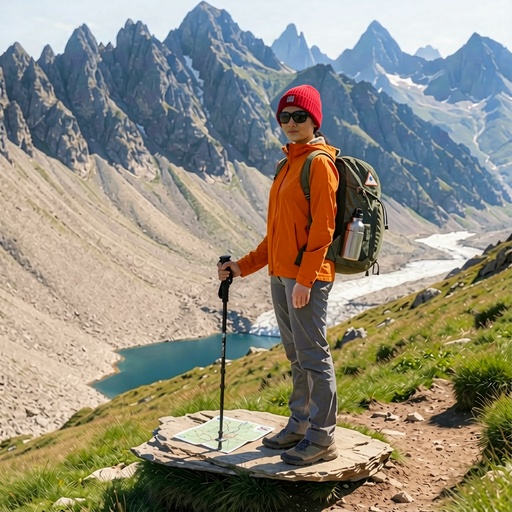}
      & \limimg{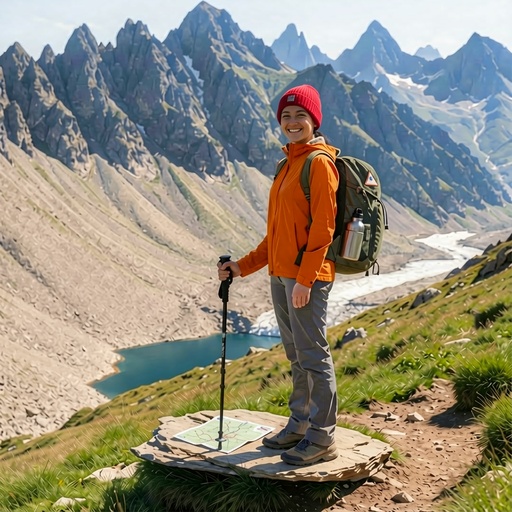}
      \\
      \noalign{\vskip 4pt}

      &
      & \liminstruction{``Add sunglasses''}
      & \liminstruction{``Add a red beanie''}
      & \liminstruction{``Make the hiker smile''}
      \\[3pt]

    \end{tabular}
  \end{adjustbox}

  \endgroup

  \caption{%
  \textbf{Limitations.}
  We show selected steps from a nine-edit sequence.
  Both the base model and ours follow the final instruction ``Make the hiker smile,''
  but also remove the sunglasses added in a previous edit.
  While our method reduces accumulated drift, it inherits the editing capabilities and limitations of the base model.
}
  \label{fig:editability_limitation}
\end{figure}

We introduced a training approach for stable multi-turn image editing.
Our central idea is that stability should be learned over the entire
recursive process, rather than at individual editing steps. We achieve
this through \emph{asymmetric conditioning}: the student learns from
its own generated editing trajectories, while the teacher remains
anchored to the original clean image and the accumulated editing
instructions. This allows us to supervise a long, imperfect sequence
of edits through a shorter, clean alternative path, without exposing
the teacher to the errors accumulated by the student.

Interestingly, multi-turn editing does not always lead to degradation.
When edits repeatedly replace the subject and background, existing
models can maintain visual quality even over long sequences
(Fig.~\ref{fig:replacement_cycle}). We attribute this to the fact that
such edits generate new content rather than preserving existing
details. In contrast, edits that require content to persist across
many turns are more susceptible to accumulated errors. Our work
therefore focuses on sequences in which unaffected content must
remain consistent.

Our approach substantially reduces accumulated drift, but
some challenges remain. Particularly difficult editing instructions may
still affect previously edited content
(Fig.~\ref{fig:editability_limitation}), suggesting that stability
and instruction following are related but distinct objectives.
Furthermore, although our method improves stability beyond
the sequence lengths seen during training, extending this
robustness to much longer editing sequences remains an
interesting direction for future work.

Beyond image editing, our approach suggests a more general training
principle for recursive generative processes. Self-Forcing exposes
the model to the distribution it will encounter during inference,
while asymmetric teacher supervision provides reliable guidance
from a clean reference. Score distillation connects these two
paths without requiring paired target outputs, which is particularly
useful when many valid results are possible. We believe this
combination may be useful in other generative tasks where models
repeatedly operate on their own outputs.

\section*{Acknowledgments}
We thank Ron Mokady for his valuable feedback. This research was supported in part by the Israel Science Foundation (grants no. 3441/21 and 1473/24), Len Blavatnik and the Blavatnik Family Foundation, and the Bronfman Family Foundation. It was made possible through a GPU compute resource grant funded by the Association of University Heads, the Council for Higher Education and the AI Research Compute Center.

{
    \small
    \bibliographystyle{ieeenat_fullname}
    \bibliography{main}
}

\clearpage
\appendix
\raggedbottom

\section*{\LARGE Appendix}

\section{Training Pseudocode}
\label{sec:training_pseudocode}

Algorithms~\ref{alg:self_forced_rollout} and~\ref{alg:training} summarize
the training procedure in Sec.~\ref{sec:method}. Hyperparameters are the rollout
length $E$, sampling-step count $K$, warm-up duration $W$, student-update
interval $R$, guidance scale $\gamma$, and training-iteration count $T$. Here,
$\operatorname{sg}$ denotes stop-gradient, and $\|\cdot\|_2^2$ the mean squared error over latent elements. Clean estimates are computed as
$\hat{x}_0=x_\sigma-\sigma v$. All images passed between edits are represented as
latents. $\operatorname{Sample}_{<s}$ runs the sampling steps before the exit
step, and $\operatorname{Sample}_{>s}$ completes sampling from the exit-step
state and velocity. $\operatorname{Update}$ denotes an optimizer step,
$\mathcal{D}$ the training data, and $p_\tau$ the noise-level distribution.
The teacher $v_\phi$ is evaluated with classifier-free guidance,
$v^\gamma_\phi(x;I_0,h)=v_\phi(x;I_0,\varnothing)+\gamma\,(v_\phi(x;I_0,h)-v_\phi(x;I_0,\varnothing))$,
where $\varnothing$ is the empty instruction; the auxiliary editor $v_\psi$ is not guided.
The compound instruction $h_e$ expresses the accumulated edit intent through
turn $e$.

\begin{algorithm}[h]
\caption{Self-generated editing rollout (Sec.~\ref{sec:rollouts})}
\label{alg:self_forced_rollout}
\begin{algorithmic}[1]
\Require Student $G_\theta$, source $I_0$, edits $c_{1:E}$,
    sampling-step count $K$, gradient flag $b$
\Ensure Clean exit estimates $\hat{x}^s_{1:E}$
\State $s \sim \operatorname{Uniform}\{1,\ldots,K\}$
\For{$e=1,\ldots,E$}
    \State $z \sim \mathcal{N}(0,\mathbf{I})$
    \State $x_s \gets \operatorname{sg}(\operatorname{Sample}_{<s}
        (G_\theta,z;I_{e-1},c_e))$
    \State $v_s \gets v_\theta(x_s;I_{e-1},c_e)$
    \State $\hat{x}^s_e \gets x_s-\sigma_s v_s$
    \If{$\neg b$}
        \State $\hat{x}^s_e \gets \operatorname{sg}(\hat{x}^s_e)$
    \EndIf
    \If{$e<E$}
        \State $I_e \gets \operatorname{sg}(\operatorname{Sample}_{>s}
            (G_\theta,x_s,v_s;I_{e-1},c_e))$
    \EndIf
\EndFor
\State \Return $\hat{x}^s_{1:E}$
\end{algorithmic}
\end{algorithm}

\begin{algorithm}[h]
\caption{Training (Sec.~\ref{sec:training})}
\label{alg:training}
\begin{algorithmic}[1]
\Require Student $G_\theta$, teacher $v_\phi$, data $\mathcal{D}$
\Require Hyperparameters $E,K,W,R,\gamma,T$
\State $v_\psi \gets v_\phi$; freeze $v_\phi$
\For{$t=1,\ldots,T$}
    \If{$t>W$ \textbf{and} $(t-W-1)\bmod R=0$}
        \State $(I_0,c_{1:E},h_{1:E}) \sim \mathcal{D}$
        \State $\hat{x}^s_{1:E} \gets
            \Call{Rollout}{G_\theta,I_0,c_{1:E},K,\mathrm{true}}$
        \For{$e=1,\ldots,E$}
            \State $\tau \sim p_\tau$, $\epsilon \sim \mathcal{N}(0,\mathbf{I})$
            \State $x_\tau \gets (1-\sigma_\tau)\hat{x}^s_e
                +\sigma_\tau\epsilon$
            \State $\hat{x}^{\,\mathrm{teacher}}_0 \gets \operatorname{sg}(
                x_\tau-\sigma_\tau v^\gamma_\phi(x_\tau;I_0,h_e))$
            \State $\hat{x}^{\,\mathrm{aux}}_0 \gets \operatorname{sg}(
                x_\tau-\sigma_\tau v_\psi(x_\tau;I_0,h_e))$
            \State $g_e \gets
                \dfrac{\hat{x}^{\,\mathrm{aux}}_0-\hat{x}^{\,\mathrm{teacher}}_0}
                {\operatorname{mean}|\hat{x}^s_e-\hat{x}^{\,\mathrm{teacher}}_0|}$
            \State $\ell^G_e \gets \frac{1}{2}
                \|\hat{x}^s_e-\operatorname{sg}(\hat{x}^s_e-g_e)\|_2^2$
        \EndFor
        \State $\theta \gets \operatorname{Update}(\theta,
            \nabla_\theta\frac{1}{E}\sum_e\ell^G_e)$
    \EndIf
    \State $(I_0,c_{1:E},h_{1:E}) \sim \mathcal{D}$
    \State $\hat{x}^s_{1:E} \gets \Call{Rollout}{G_\theta,I_0,c_{1:E},K,\mathrm{false}}$
    \For{$e=1,\ldots,E$}
        \State $\tau' \sim p_\tau$, $\epsilon' \sim \mathcal{N}(0,\mathbf{I})$
        \State $x_{\tau'} \gets (1-\sigma_{\tau'})\hat{x}^s_e
            +\sigma_{\tau'}\epsilon'$
        \State $\ell^{\mathrm{aux}}_e \gets
            \|v_\psi(x_{\tau'};I_0,h_e)
            -(\epsilon'-\operatorname{sg}(\hat{x}^s_e))\|_2^2$
    \EndFor
    \State $\psi \gets \operatorname{Update}(\psi,
        \nabla_\psi\frac{1}{E}\sum_e\ell^{\mathrm{aux}}_e)$
\EndFor
\end{algorithmic}
\end{algorithm}
\vspace{-8pt}

\section{Evaluation Metrics}
\label{app:metrics}

All metrics are computed at the final turn (turn 10) of each EdiVal-10 chain and averaged over chains; comparisons between methods are paired over chains and reported with standard errors.
The three judged scores use the prompts of KRIS-Bench~\cite{wu2025krisbench} unchanged, with Gemma-4-31B~\cite{gemma2026gemma4} as the judge in place of GPT-4o.

\vspace{-8pt}
\paragraph{Source images.}
EdiVal-10 draws its chains only from GEdit-Bench photographs whose short side is at least 768 pixels, so the center crop and resize to $1024\times1024$ upsamples a source by at most $1.33\times$.

\vspace{-14pt}
\paragraph{Quality.}
The judge rates the final image with the image-quality prompt, checking structural coherence, natural appearance, absence of artifacts and texture fidelity, and yielding a score from 1 to 5.
The prompt describes every image as AI-generated, and on our real photographs the judge also reports defects in untouched sources (their mean score is 3.57); since all methods are scored on the same photographs and compared chain by chain, this affects the absolute scores but not the comparisons.
$\Delta$HPSv3 is the change in HPSv3~\cite{ma2025hpsv3}, a learned estimate of perceived image quality, from the first edited image (turn 1) to the final image.
Measuring from turn 1 rather than from the source separates the quality lost over the chain from the effect of a single edit.

\vspace{-12pt}
\paragraph{Preservation.}
On the whole image, the judge rates consistency with the source using the visual-consistency prompt.
It sees the source, the final image and the chain's ten instructions as one numbered list, and scores from 1 to 5 how well everything the instructions do not ask to change is preserved (5: all such elements unchanged; 1: most of the scene altered).
As controls, on the first 20 EdiVal-10 chains it scores each source compared with itself 5, and each source compared with the final image of a different chain 1.
Locally, we measure each chain's anchor objects, which no instruction edits or mentions (Sec.~\ref{sec:exp:benchmark}).
We crop the box of each anchor from the source and from the final image, grey out in both crops the parts that fall inside the box of any edited object, and compute the DINOv2~\cite{oquab2024dinov2} distance (one minus the cosine similarity of the CLS features) between the two crops.
Because the boxes are coarse, an anchor can lie largely inside the box of an edited object (e.g., a shirt worn by a person who is edited), leaving too little of it to compare.
We therefore use an anchor only if at least 30\% of its box remains after the masking, and take a chain's anchor distance as the mean over its usable anchors; 148 of the 252 chains have at least one, and the anchor distance is averaged over these chains.
We also compute LPIPS~\cite{zhang2018lpips} between the source and the final image over the background, i.e., the whole image outside the boxes of all edited objects (a median of 57\% of the image).
Added objects are placed in this region, so the background distance should be read together with the instruction-following score.

\vspace{-8pt}
\paragraph{Editing.}
A model can preserve content trivially by not editing, so we check that the requested edits are carried out and survive the chain.
The judge applies the instruction-following prompt: it sees the source, the final image and the chain's ten instructions as one numbered list, and scores from 1 to 5 whether the requested changes are present and correct (5: all present and accurate; 3: partially; 1: not at all), leaving unrequested changes to the consistency score.
As a negative control, showing the source image instead of the final image yields a score of 1 for all 20 chains.

\vspace{-3pt}
\section{Long Sequences with Repeated Content Replacement}
\vspace{-3pt}
\label{sec:appendix_replacement}

\begin{figure*}[t]
  \centering
  \begingroup
  \setlength{\tabcolsep}{2pt}
  \newcommand{\cycleturn}[1]{\scriptsize #1}
  \begin{tabular}{@{}ccccc@{}}
    \cycleturn{Dog / living room} & \cycleturn{Cat / living room}
    & \cycleturn{Cat / beach} & \cycleturn{Dog / beach}
    & \cycleturn{Dog / living room} \\[3pt]
    \includegraphics[width=0.19\textwidth]{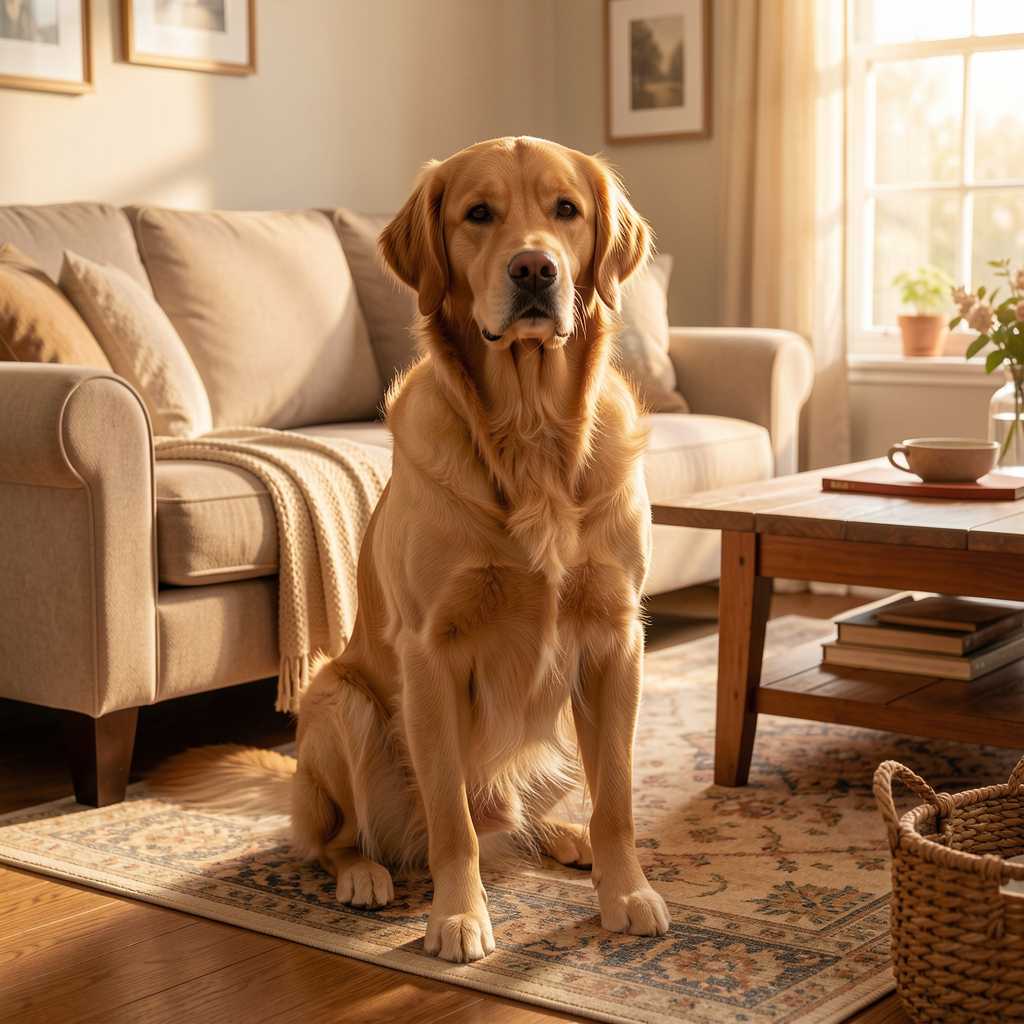} & \includegraphics[width=0.19\textwidth]{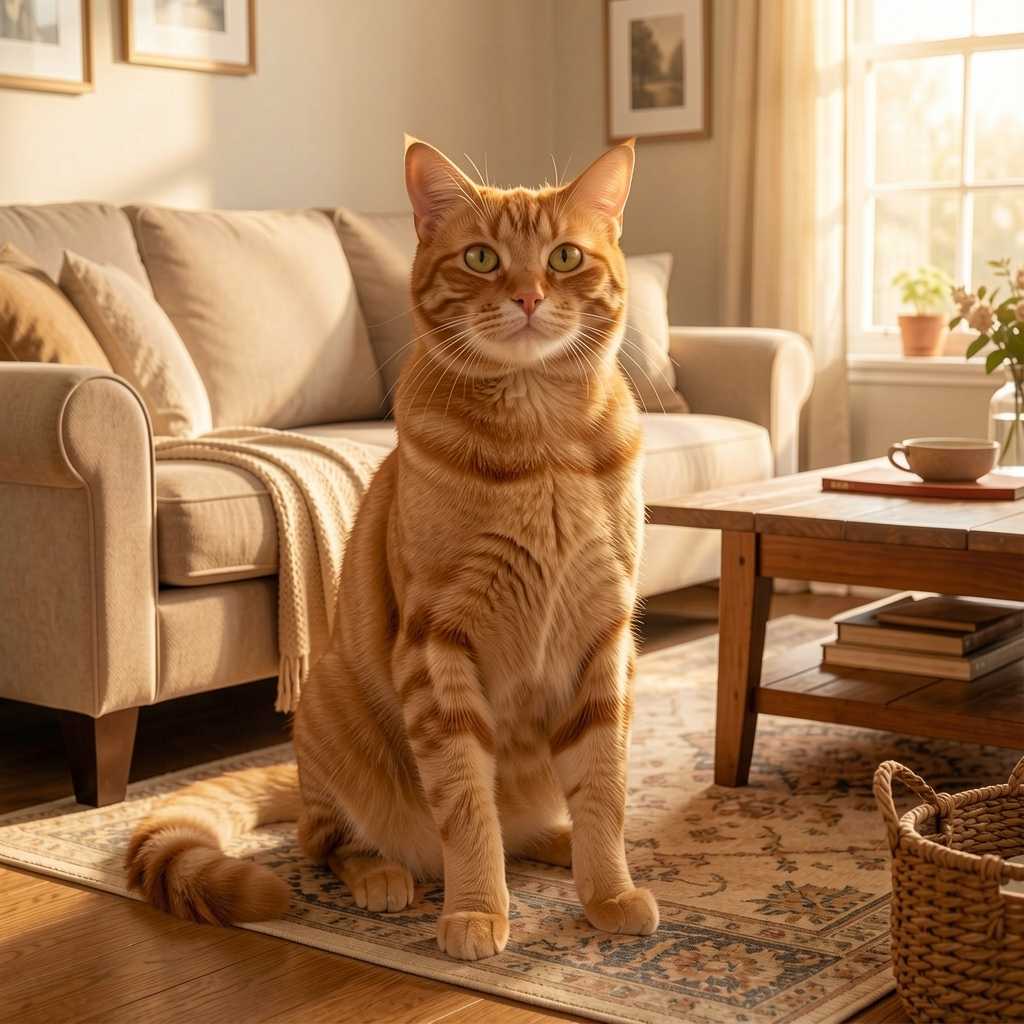}
    & \includegraphics[width=0.19\textwidth]{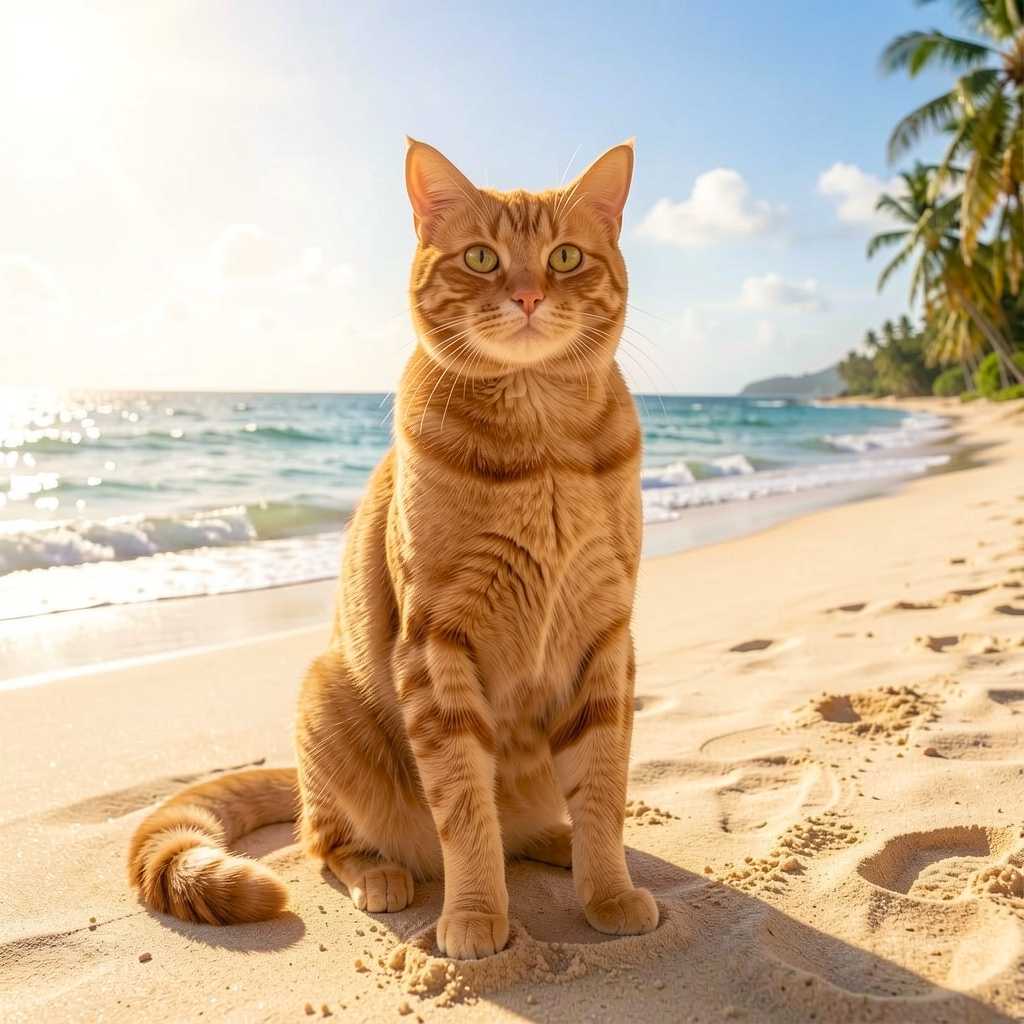} & \includegraphics[width=0.19\textwidth]{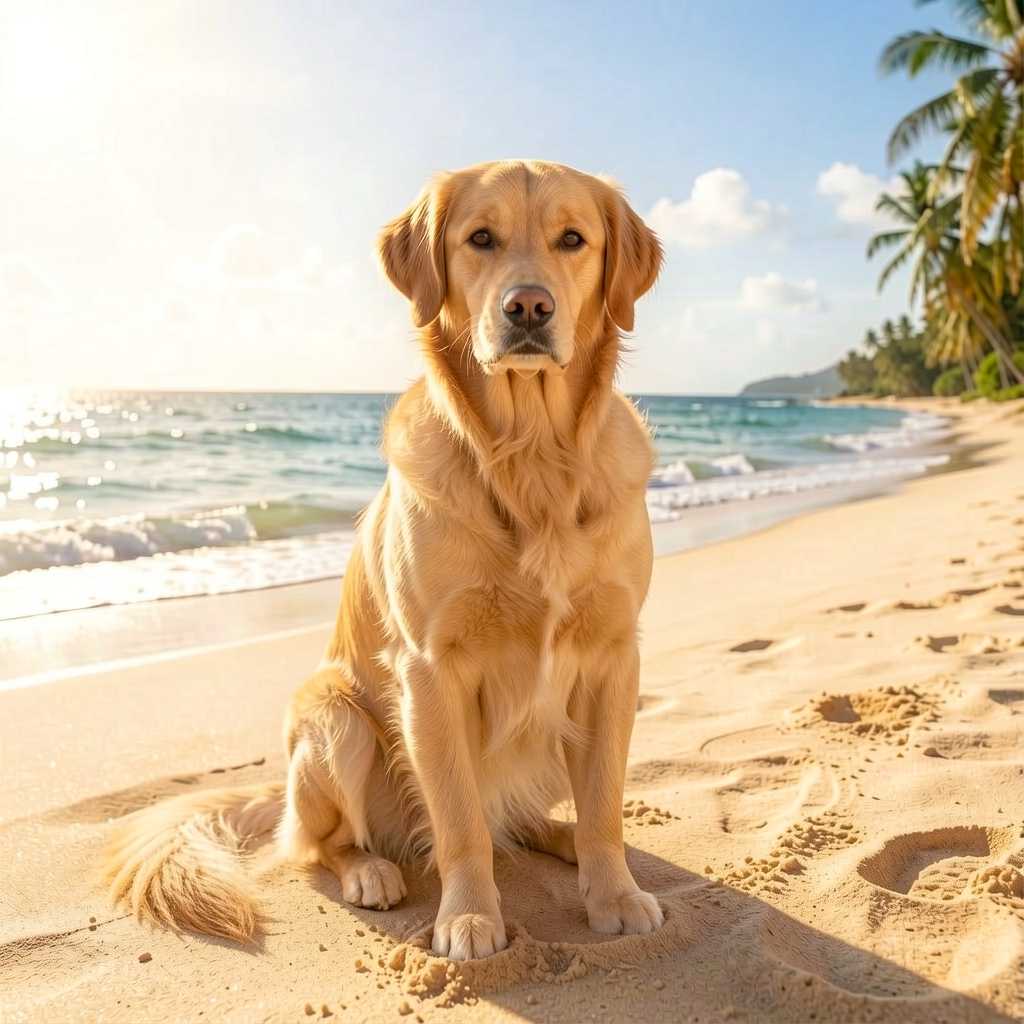}
    & \includegraphics[width=0.19\textwidth]{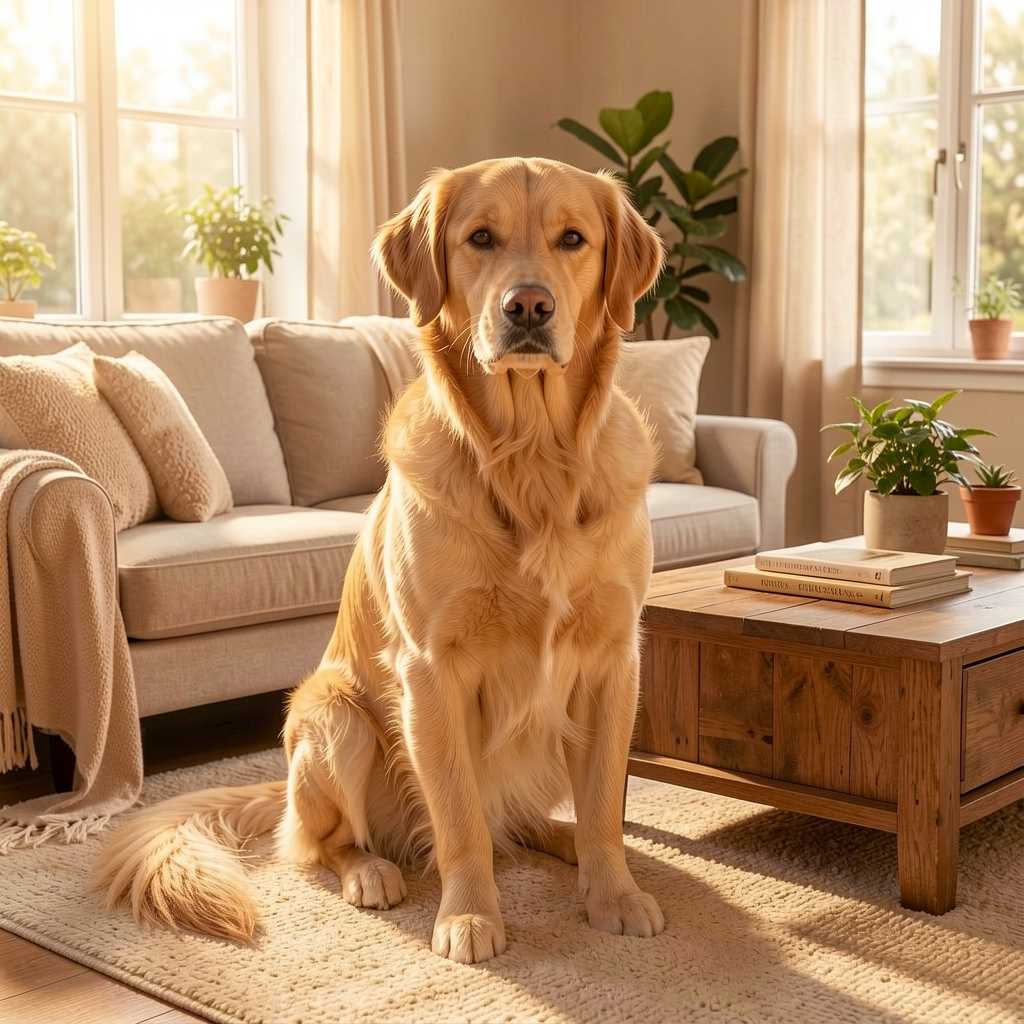} \\
    \cycleturn{Turn 0 (input)} & \cycleturn{Turn 1}
    & \cycleturn{Turn 2} & \cycleturn{Turn 3}
    & \cycleturn{Turn 4} \\[6pt]
    \includegraphics[width=0.19\textwidth]{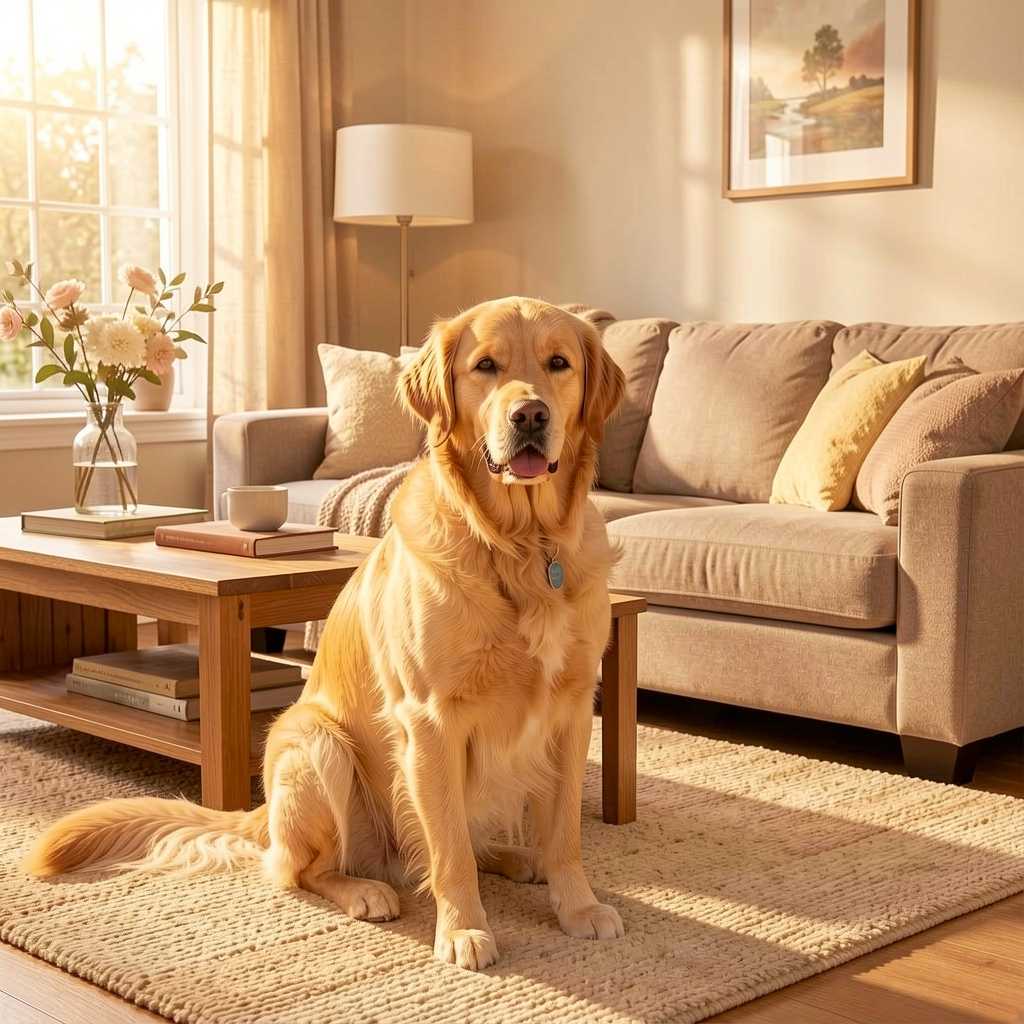} & \includegraphics[width=0.19\textwidth]{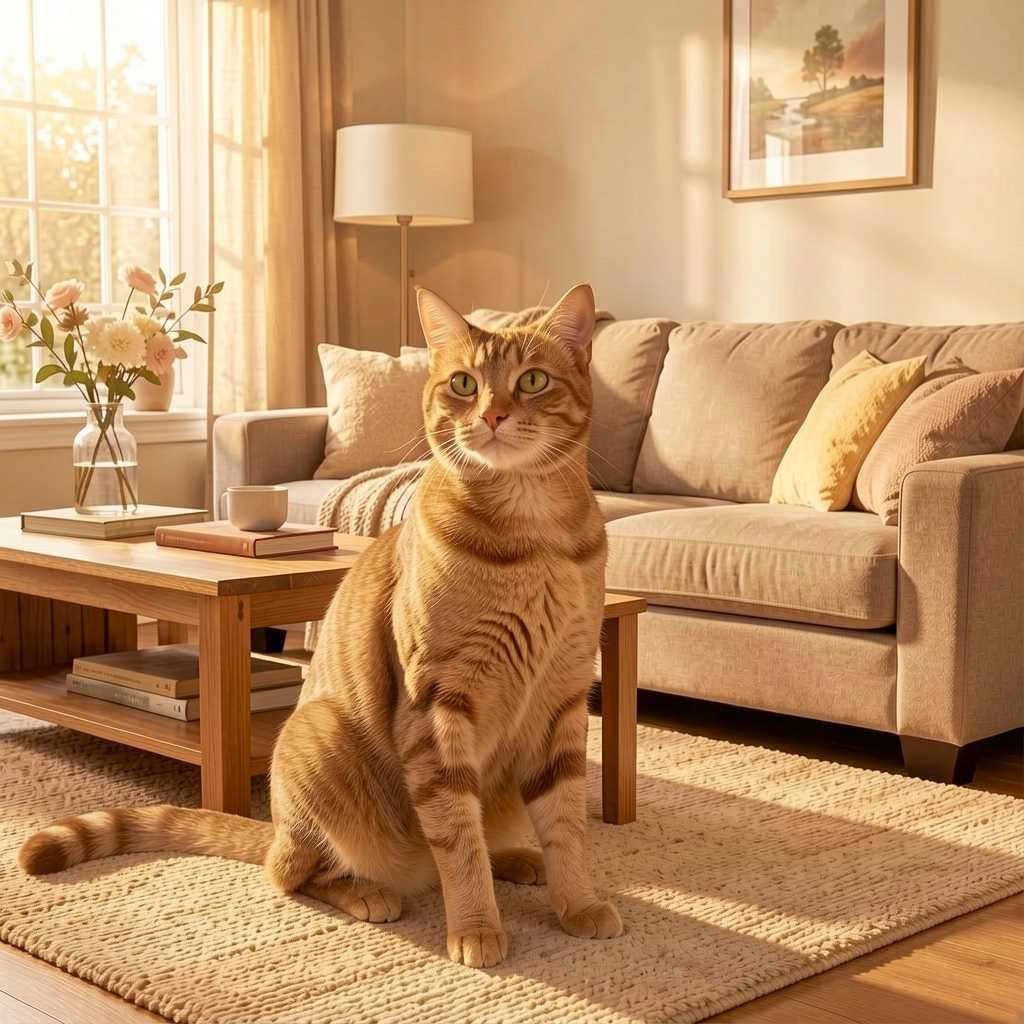}
    & \includegraphics[width=0.19\textwidth]{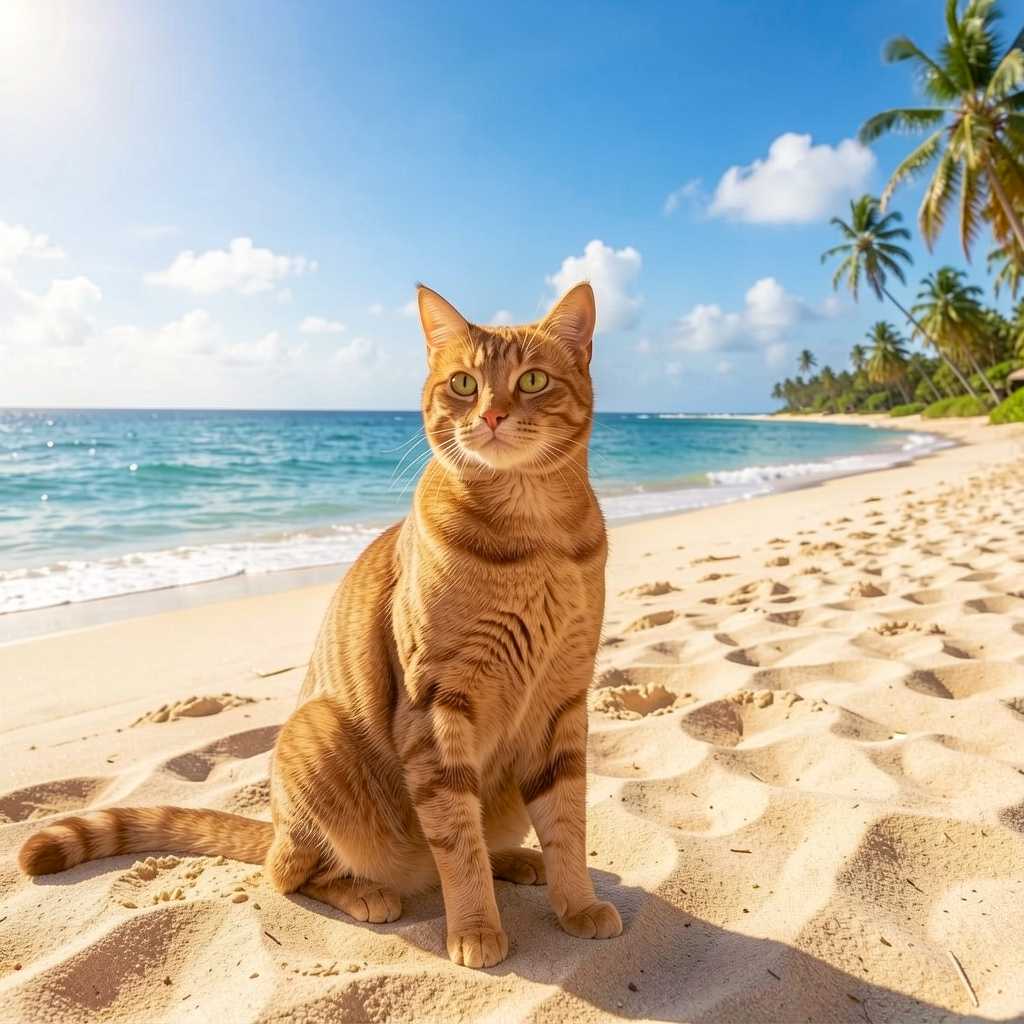} & \includegraphics[width=0.19\textwidth]{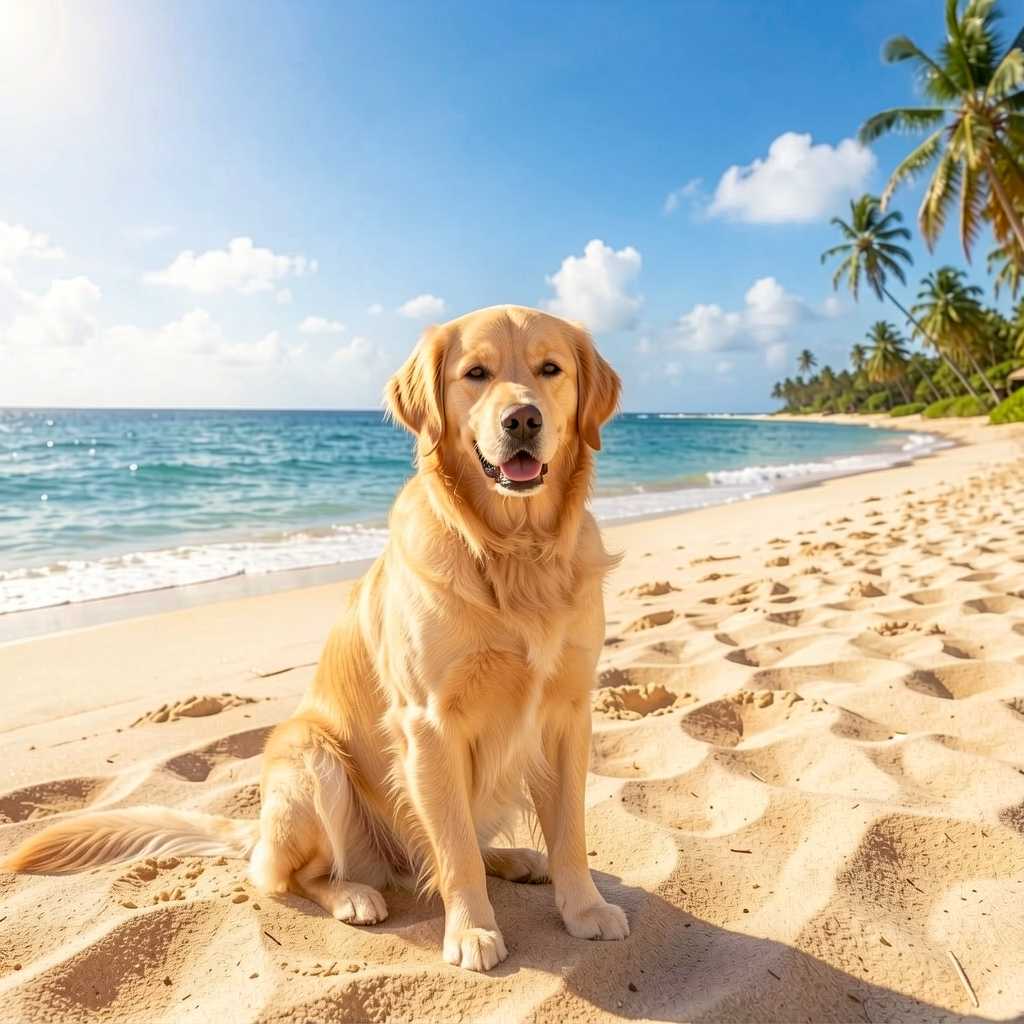}
    & \includegraphics[width=0.19\textwidth]{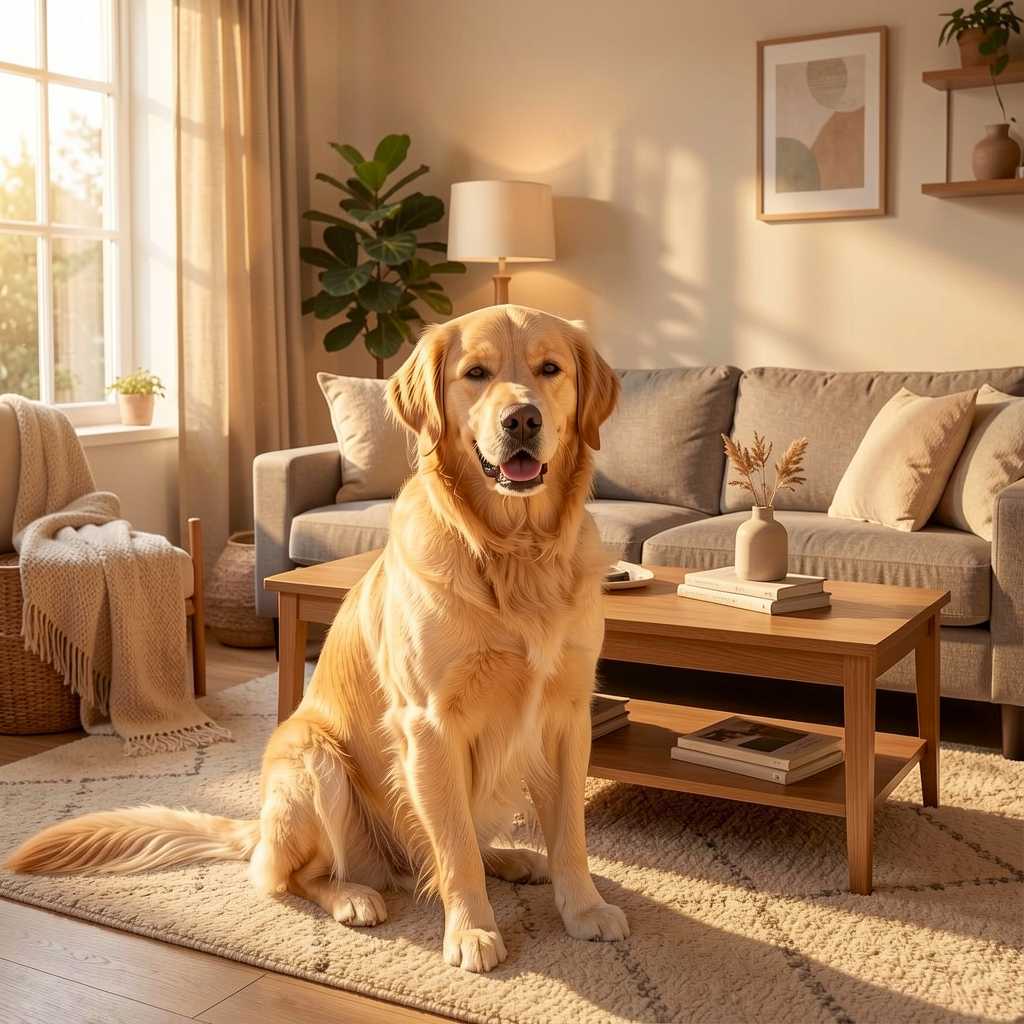} \\
    \cycleturn{Turn 16} & \cycleturn{Turn 17}
    & \cycleturn{Turn 18} & \cycleturn{Turn 19}
    & \cycleturn{Turn 20} \\[6pt]
    \includegraphics[width=0.19\textwidth]{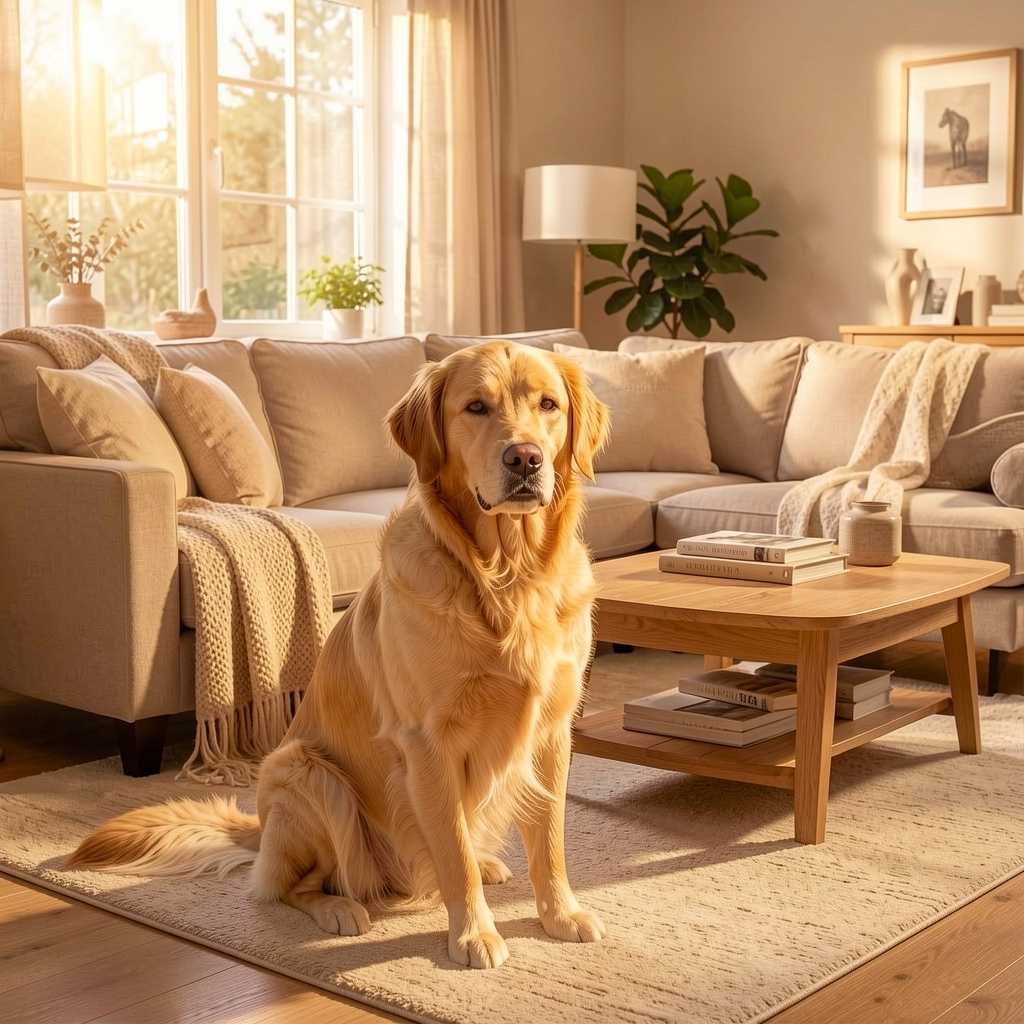} & \includegraphics[width=0.19\textwidth]{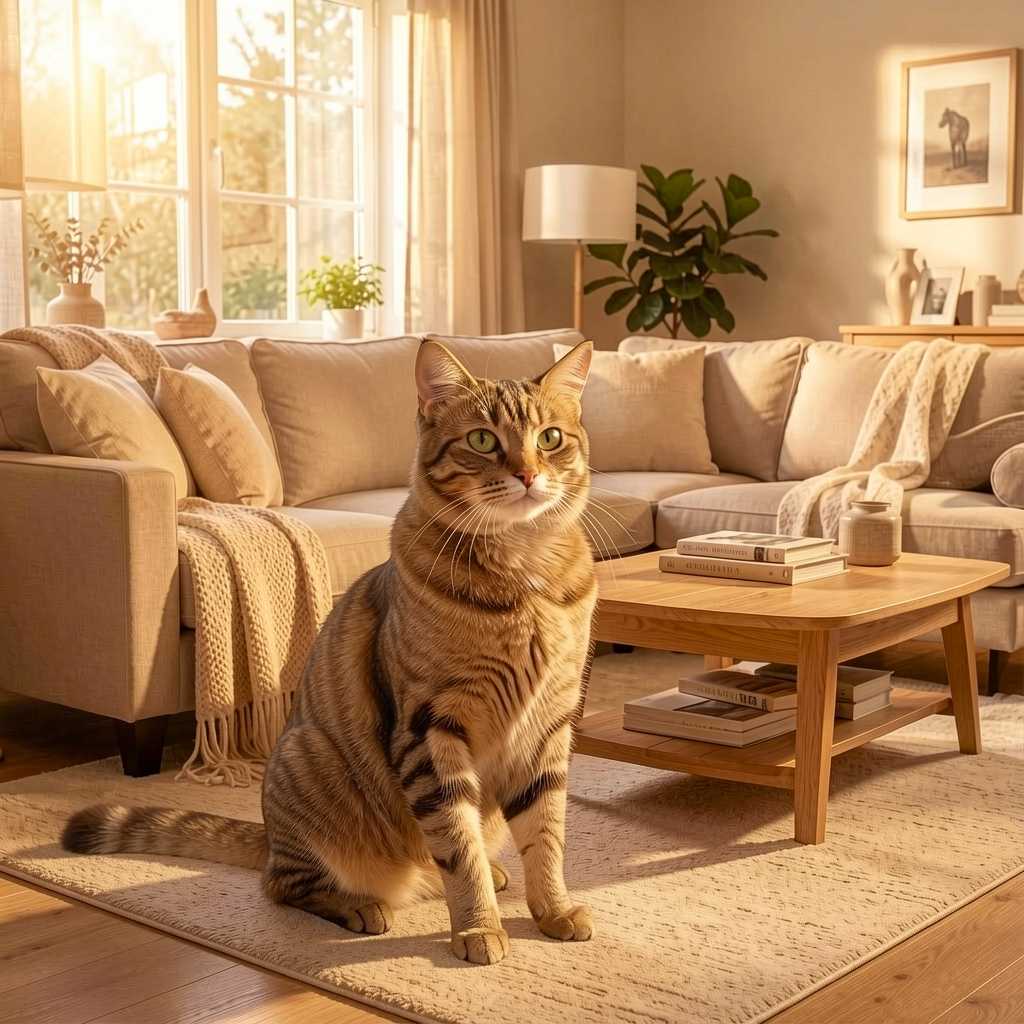}
    & \includegraphics[width=0.19\textwidth]{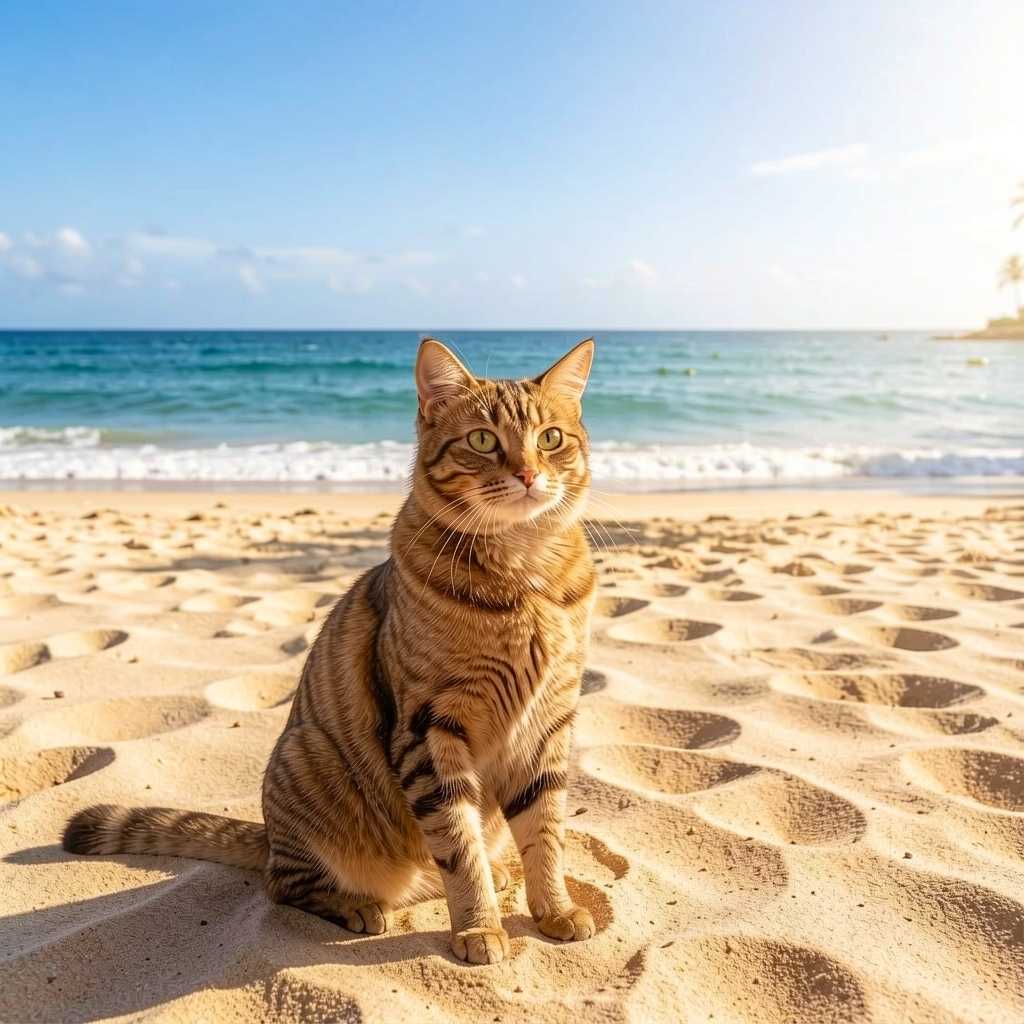} & \includegraphics[width=0.19\textwidth]{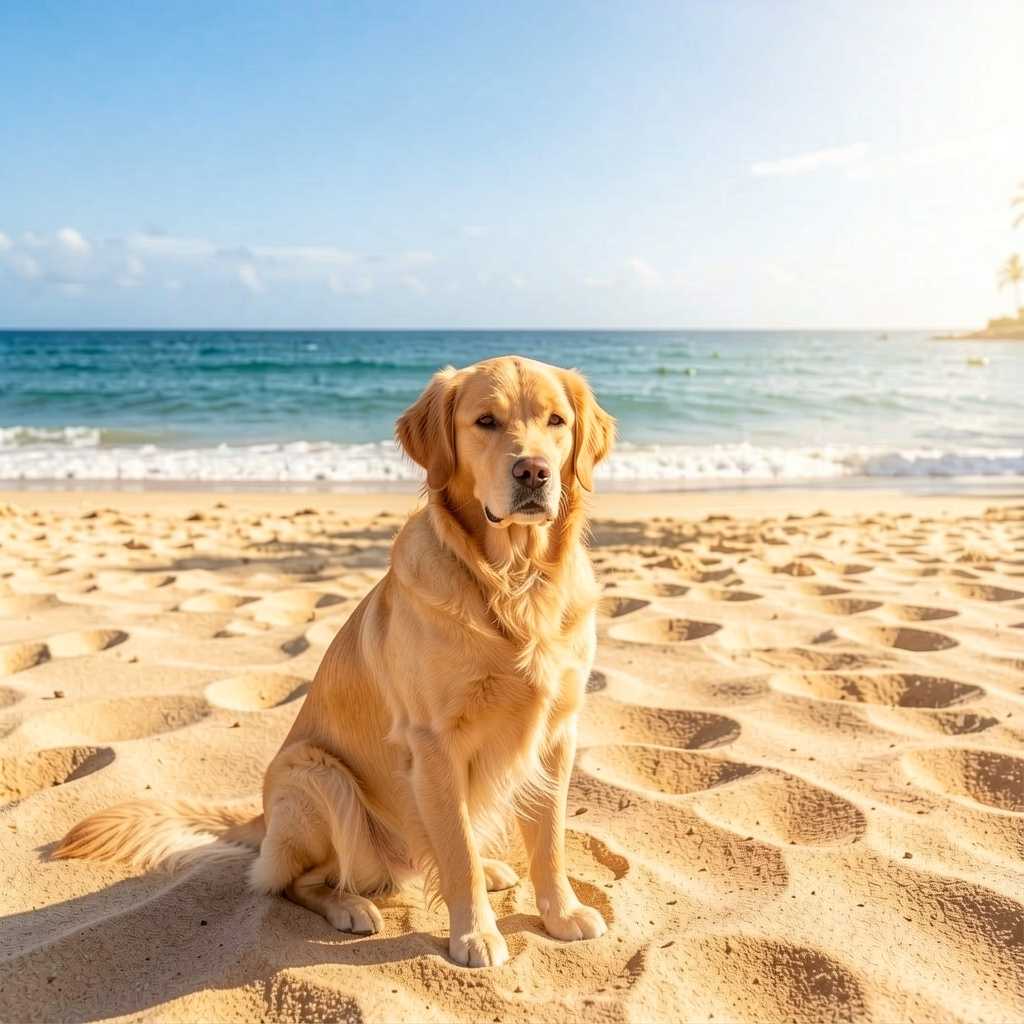}
    & \includegraphics[width=0.19\textwidth]{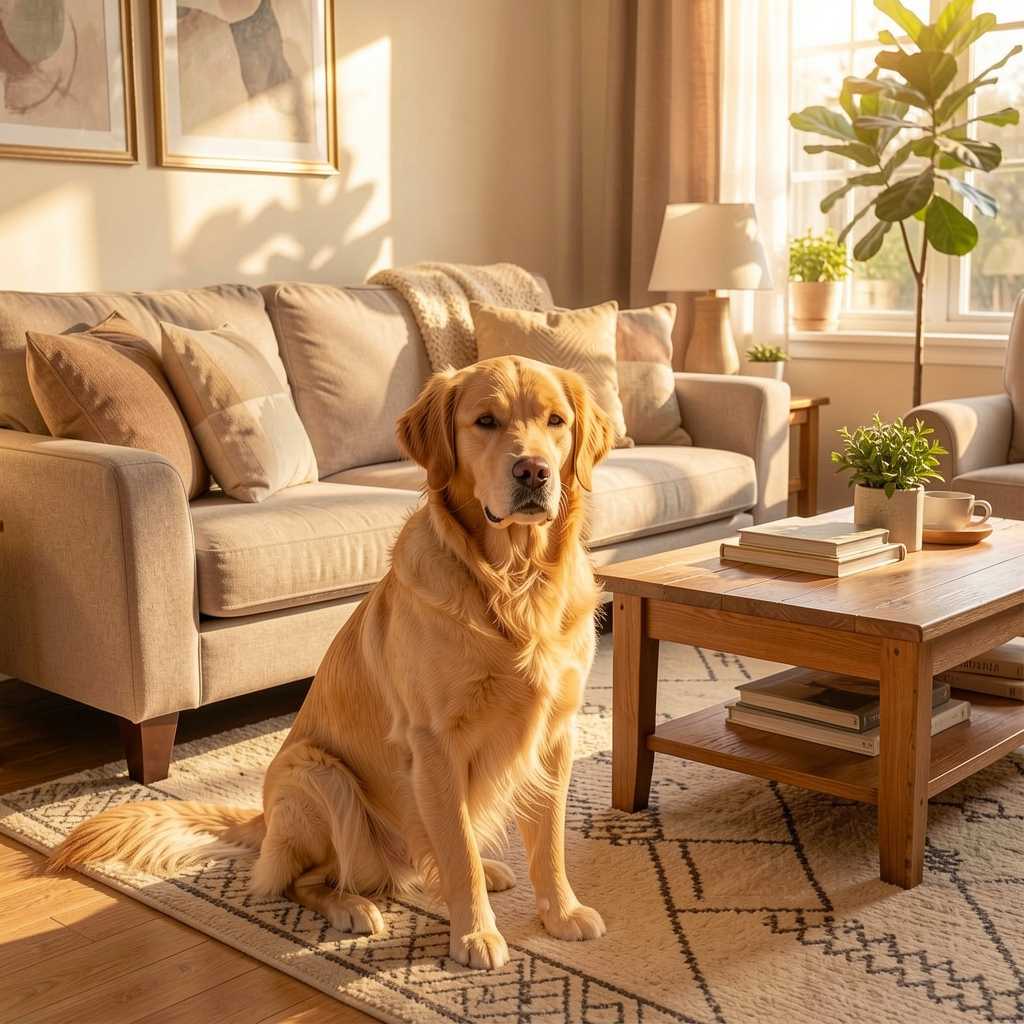} \\
    \cycleturn{Turn 36} & \cycleturn{Turn 37}
    & \cycleturn{Turn 38} & \cycleturn{Turn 39}
    & \cycleturn{Turn 40}
  \end{tabular}
  \endgroup
  \vspace{-4pt}
  \caption{\textbf{Repeated content replacement over 40 edits.}
    The unmodified distilled FLUX.2 Klein 9B model cycles from a dog in a
    living room to a cat in a living room, a cat on a beach, a dog on a beach,
    and a dog in a living room. We show the first, fifth, and tenth cycles
    of this sequential rollout. Visual quality remains coherent at late
    turns, while subject appearance and scene details change across cycles.
    Replacing both the subject and background every two turns limits
    long-term preservation demands, illustrating that sequence length alone
    does not necessarily lead to progressive visual degradation.}
    \vspace{-4pt}
  \label{fig:replacement_cycle}
\end{figure*}

To distinguish sequence length from long-term preservation demands, we
consider a 40-edit sequence produced by the unmodified distilled
FLUX.2 Klein 9B model (\cref{fig:replacement_cycle}). Starting from a dog in a
living room, we repeat four edits ten times: replace the dog with a cat,
move the scene to a beach, replace the cat with a dog, and move the scene
back to a living room. The subject and background are each replaced every
two turns, so neither must retain its original details throughout the sequence.
The outputs remain visually coherent through turn 40, without an obvious
progressive loss of visual quality. However, the animal's appearance and the
room layout vary across cycles; returning to the same semantic categories
does not imply recovery of the original image. This qualitative example
shows that a long editing sequence need not exhibit progressive degradation
when content is repeatedly replaced, and motivates evaluating stability in
settings that require content to persist across many edits.

\vspace{-3pt}
\section{Results on LME-Bench}
\vspace{-3pt}
\label{app:lme}

Table~\ref{tab:lme} reports results on LME-Bench~\cite{zhao2026mtopsd} (Sec.~\ref{sec:exp:benchmark}) in MT-OPSD's format.
We score it with MT-OPSD's evaluation code, using Gemma-4-31B in place of GPT-4o as for our other judged metrics, and show every image to the judge at 512 pixels.
A turn succeeds if the judge rates both its instruction following and its consistency above 6 out of 10; the success rate (SR) at turn $k$ is the share of the 100 sessions whose turns 1 to $k$ all succeed.
The collapse rate (CR) at turn $k$ is the share of sessions in which, by turn $k$, two consecutive turns have been judged destroyed (severe content loss or surface corruption relative to the source).
MT-OPSD and its base model are sampled as in Sec.~\ref{sec:exp:setup}, without upsampling; we run our model and its base model with their own 4-step sampler, also generating at 512 pixels and passing each turn's output image to the next turn.
Absolute numbers depend on the judge: MT-OPSD reports an SR at turn 10 of 0.38 for its FLUX model under GPT-4o, against 0.10 under our judge, so Table~\ref{tab:lme} is not comparable with the numbers reported in~\cite{zhao2026mtopsd}.

Our model completes all ten turns in 53 of the 100 sessions, against 10 for MT-OPSD, and collapses in 2, against 16.
Most of this gap comes from the backbone and its sampler, as our base model already completes 49 sessions.
Compared with it, our model succeeds significantly more often at turns 5 and 8, but not at turn 10 (0.53 vs.\ 0.49), and the base model rarely collapses on this benchmark (0.04 at turn 10), leaving little room for improvement.
At 1024 pixels, the resolution of our other experiments, the two are tied in every column (SR at turn 10: 0.53 vs.\ 0.46).

\begin{table}[h]
\centering
\vspace{-4pt}
\caption{\textbf{LME-Bench} (100 sessions of 10 turns; Gemma-4-31B judge, 512-pixel protocol).
SR: share of sessions whose turns 1 to $k$ all succeed; CR: share of sessions collapsed by turn $k$.
$^*$: significantly different from ours (exact sign test over sessions, $p<0.05$).}
\vspace{-4pt}
\label{tab:lme}
\footnotesize
\setlength{\tabcolsep}{1.5pt}
\begin{tabular*}{\columnwidth}{@{\extracolsep{\fill}}lcccccccc}
\toprule
 & \multicolumn{4}{c}{SR at turn $k$ $\uparrow$} & \multicolumn{4}{c}{CR at turn $k$ $\downarrow$} \\
\cmidrule(lr){2-5} \cmidrule(lr){6-9}
Method & 3 & 5 & 8 & 10 & 3 & 5 & 8 & 10 \\
\midrule
\multicolumn{9}{l}{\textit{FLUX.2-klein-base-9B (undistilled)}} \\
Base model & 0.65$^*$ & 0.41$^*$ & 0.19$^*$ & 0.09$^*$ & 0.01 & 0.07 & 0.35$^*$ & 0.47$^*$ \\
MT-OPSD~\cite{zhao2026mtopsd} & 0.64$^*$ & 0.50$^*$ & 0.27$^*$ & 0.10$^*$ & 0.03 & 0.03 & 0.06 & 0.16$^*$ \\
\midrule
\multicolumn{9}{l}{\textit{FLUX.2-klein-9B (4-step distilled)}} \\
Base model & 0.89 & 0.79$^*$ & 0.63$^*$ & 0.49 & 0.02 & 0.02 & 0.04 & 0.04 \\
Ours & 0.93 & 0.88 & 0.73 & 0.53 & 0.01 & 0.01 & 0.02 & 0.02 \\
\bottomrule
\end{tabular*}
\vspace{-8pt}
\end{table}

\begin{table}[t]
\centering
\caption{\textbf{Single edits on GEdit-Bench} (606 English instructions, one edit each).
Change in HPSv3 from the source to the edited image, averaged over instructions.
$^*$: significantly different from our model (more than two standard errors, paired over instructions).
$^\dagger$: VAE-LFA does not act on the first edit.}
\label{tab:single}
\footnotesize
\begin{tabular*}{\columnwidth}{@{\extracolsep{\fill}}lc}
\toprule
Method & $\Delta$HPSv3 (source$\to$edit) $\uparrow$ \\
\midrule
\multicolumn{2}{l}{\textit{FLUX.2-klein-base-9B (undistilled)}} \\
Base model & $-0.68^*$ \\
MT-OPSD~\cite{zhao2026mtopsd} & $-0.75^*$ \\
\midrule
\multicolumn{2}{l}{\textit{FLUX.2-klein-9B (4-step distilled)}} \\
Base model & $-0.41^*$ \\
+ Emu Edit~\cite{sheynin2024emu} & $-0.56^*$ \\
+ VAE-LFA~\cite{wang2026vaelfa} & same as base model$^\dagger$ \\
+ FreqEdit~\cite{liao2026freqEdit} & $-0.38^*$ \\
Ours & $\mathbf{-0.18}$ \\
\bottomrule
\end{tabular*}
\end{table}

\vspace{-4pt}
\section{Single Edits}
\label{app:single}
Every training rollout starts with an ordinary single edit (Sec.~\ref{sec:method}), but the training targets long sequences, so we check that stability does not come at the cost of single-edit quality.
We apply each method once to every English instruction of GEdit-Bench~\cite{liu2025step1xedit}: 606 instructions over 604 photographs, in eleven edit types that include global edits such as style and tone changes.
The photographs keep their aspect ratio.
Our model, its base model and the training-free methods generate at the photograph's resolution, capped at about one megapixel; MT-OPSD and its base model generate at about $512\times512$ pixels with the authors' script, and their change is measured from the source at that resolution.
We measure the change in HPSv3 from the source to the edited image, with the same HPSv3 settings as in Table~\ref{tab:comparison}, and average it over the instructions (Table~\ref{tab:single}).

A single edit lowers HPSv3 for every method, and least for our model: by 0.18, against 0.41 for its base model (difference $0.23 \pm 0.02$).
Every other method also loses significantly more than our model, MT-OPSD the most (0.75).
VAE-LFA aligns each output with earlier outputs and does not act on the first edit, so on single edits it is identical to the base model.

\section{Successive Background Changes}
\label{app:backgrounds}

\begin{figure*}[t]
  \centering
  \begingroup

  \def\teddir{images/teddy}

  \setlength{\tabcolsep}{1pt}
  \renewcommand{\arraystretch}{1.0}

  \newcommand{\tedimg}[1]{%
    \raisebox{-0.5\height}{%
      \includegraphics[width=0.185\textwidth]{#1}%
    }%
  }
  \newcommand{\tedlabel}[1]{%
  \raisebox{-0.5\height}{%
    \rotatebox{90}{\scriptsize #1}%
  }%
}
  \newcommand{\tedhead}[1]{\makebox[0.185\textwidth]{\scriptsize #1}}
  \newcommand{\tedturnrow}[2]{%
    \tedlabel{#1}
    & \tedimg{\teddir/ours/#2}
    & \tedimg{\teddir/emu/#2}
    & \tedimg{\teddir/lfa/#2}
    & \tedimg{\teddir/freqedit/#2}
    & \tedimg{\teddir/mtopsd/#2}
  }

  \begin{adjustbox}{width=\textwidth}
    \begin{tabular}{@{}c *{5}{c}@{}}

      & \tedhead{\textbf{Ours}}
      & \tedhead{Emu Edit}
      & \tedhead{VAE-LFA}
      & \tedhead{FreqEdit}
      & \tedhead{MT-OPSD}
      \\[2pt]

      \tedlabel{Source}
      & \tedimg{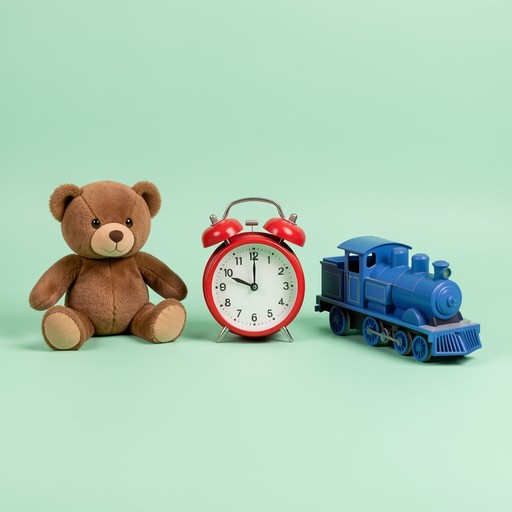}
      & \tedimg{\teddir/source.jpg}
      & \tedimg{\teddir/source.jpg}
      & \tedimg{\teddir/source.jpg}
      & \tedimg{\teddir/source.jpg}
      \\ \noalign{\vskip 2pt}
      \tedturnrow{Edit 1}{edit_01.jpg} \\ \noalign{\vskip 2pt}
      \tedturnrow{Edit 2}{edit_02.jpg} \\ \noalign{\vskip 2pt}
      \tedturnrow{Edit 3}{edit_03.jpg} \\

    \end{tabular}
  \end{adjustbox}

  \endgroup

  \caption{%
    \textbf{Successive background changes.}
    Each training-free method carries part of an earlier background into the next turn.
    The teddy bear, alarm clock and toy train of a synthetic scene are placed in a lavender field at sunset (edit 1), in front of a red brick wall (edit 2), and on a smooth gradient backdrop from warm cream to light grey (edit 3).
    Our model keeps the three objects and gives a clean gradient.
    Emu Edit keeps the low, warm light of the sunset in front of the brick wall (edit 2), and at edit 3 leaves a clump of lavender stalks on the floor of the gradient backdrop and turns the blue train grey.
    VAE-LFA keeps the colour of the sunset: at edit 3 the whole image has a purple-magenta cast, the bear turns reddish and the clock face pink.
    FreqEdit prints the pattern of the brick wall into the smooth gradient.
    MT-OPSD leaves no trace of the earlier backgrounds but blurs the image and washes out its colours.
  }
  \label{fig:teddy_backgrounds}
\end{figure*}

Figure~\ref{fig:teddy_backgrounds} shows the example discussed in Sec.~\ref{sec:exp:qualitative}: three successive background changes, in which each training-free method carries part of the previous image into the next one.

\begin{figure*}[t]
  \centering
  \begingroup

  \setlength{\qualimgsize}{0.14\textwidth}
  \setlength{\tabcolsep}{2pt}
  \renewcommand{\arraystretch}{1.0}

  \edef\qualSepOne{\the\dimexpr-4\qualimgsize-16pt\relax}
  \edef\qualSepTwo{\the\dimexpr-2\qualimgsize-8pt\relax}

  \begin{adjustbox}{width=\textwidth}
    \begin{tikzpicture}
      \node[inner sep=0pt, outer sep=0pt] (qualTable) {%
        \begin{tabular}{@{}c cc cc cc@{}}

          {\scriptsize Original}
          & \multicolumn{2}{c}{%
              \qualimg{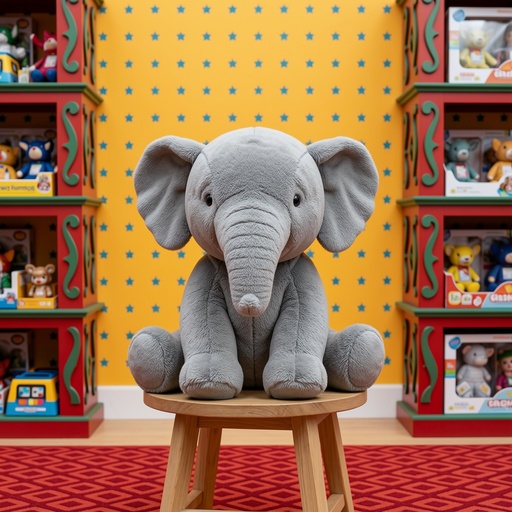}}
          & \multicolumn{2}{c}{%
              \qualimg{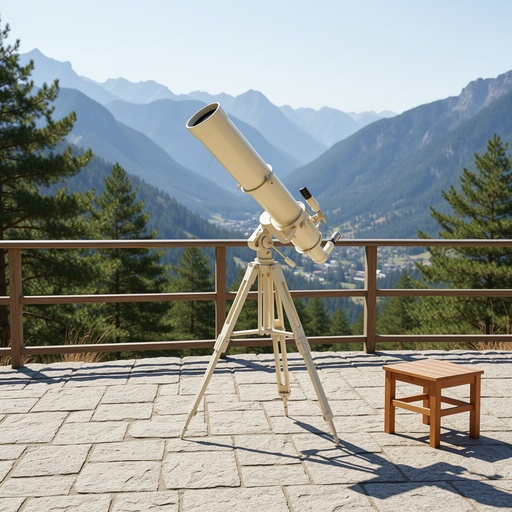}}
          & \multicolumn{2}{c}{%
              \qualimg{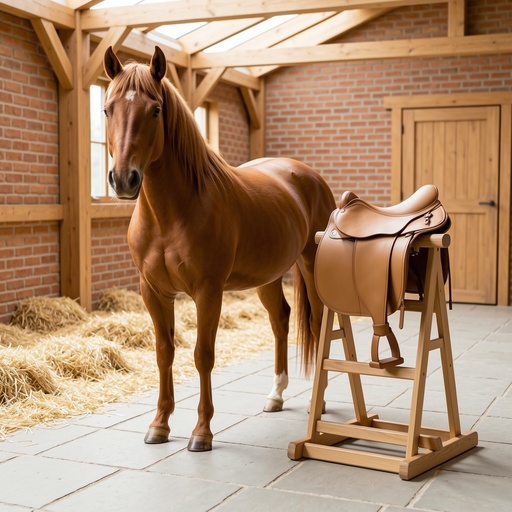}}
          \\[7pt]

          \\[3pt]

          & \qualheader{Flux.2 Klein}
          & \qualheader{+Ours}
          & \qualheader{Flux.2 Klein}
          & \qualheader{+Ours}
          & \qualheader{Flux.2 Klein}
          & \qualheader{+Ours}
          \\[5pt]

          \rotatebox{90}{\scriptsize Edit 5}
          & \qualimg{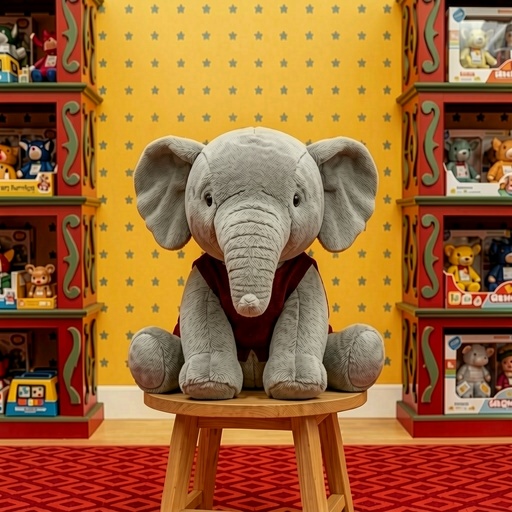}
          & \qualimg{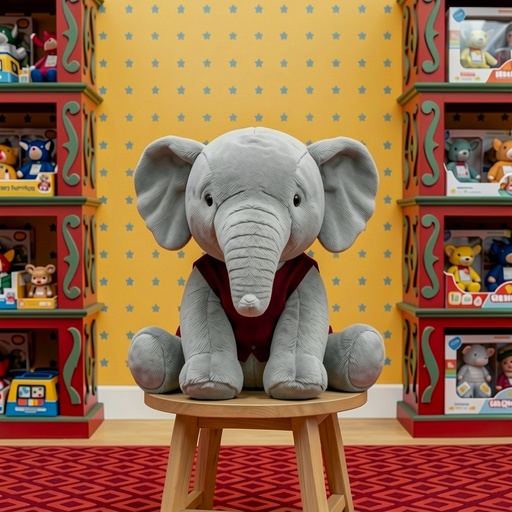}
          & \qualimg{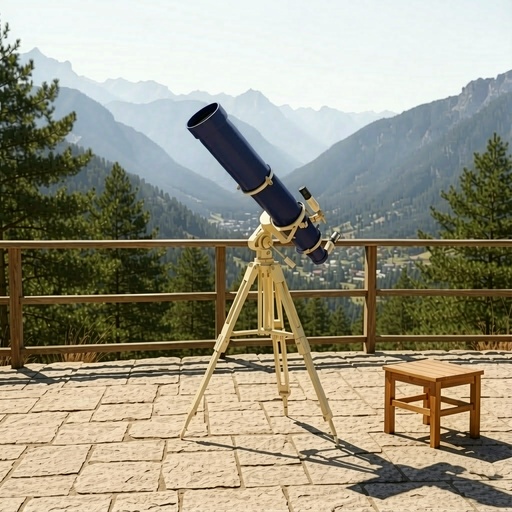}
          & \qualimg{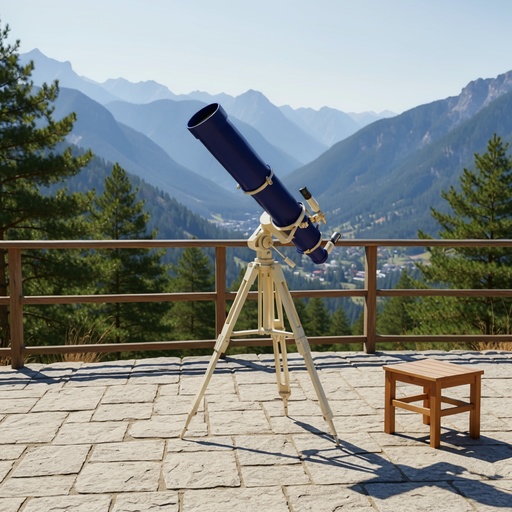}
          & \qualimg{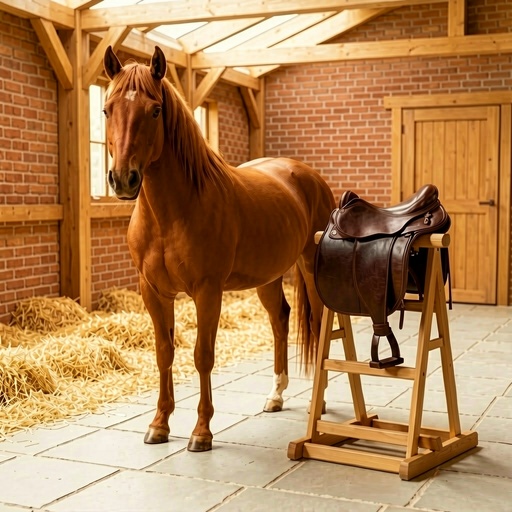}
          & \qualimg{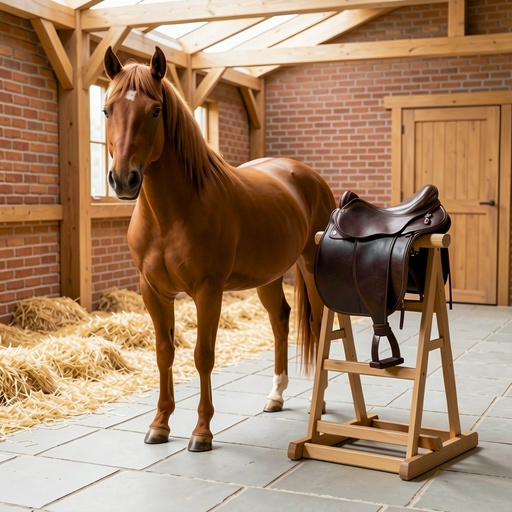}
          \\[3pt]

          & \multicolumn{2}{c}{%
              \qualinstruction{``Add a red vest''}}
          & \multicolumn{2}{c}{%
              \qualinstruction{``Change the tube color to navy''}}
          & \multicolumn{2}{c}{%
              \qualinstruction{``Change the saddle leather to dark brown''}}
          \\[8pt]

          \rotatebox{90}{\scriptsize Edit 10}
          & \qualimg{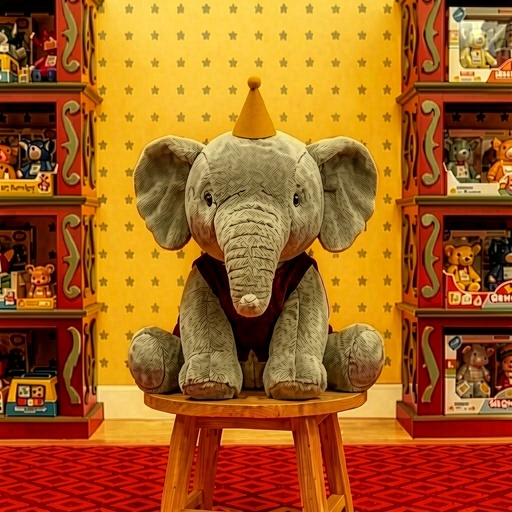}
          & \qualimg{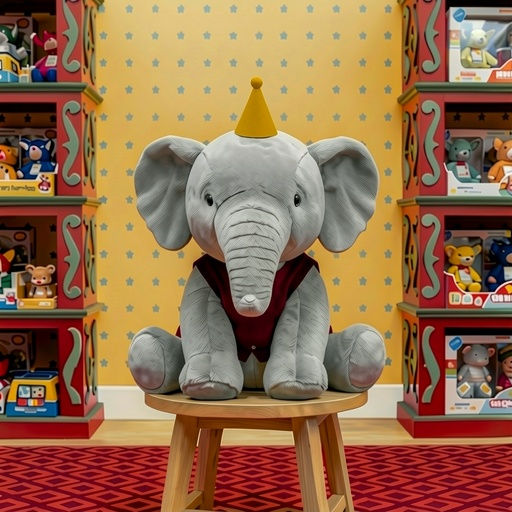}
          & \qualimg{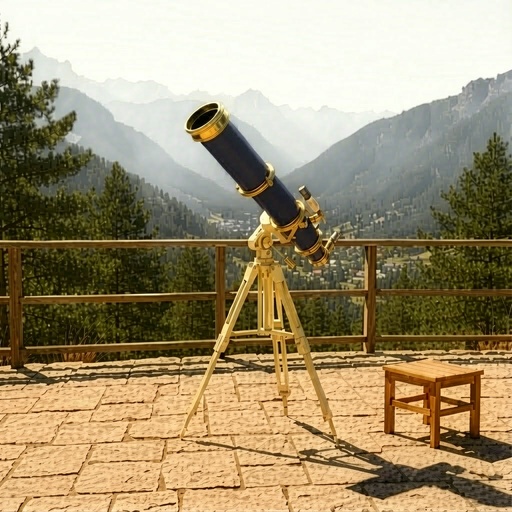}
          & \qualimg{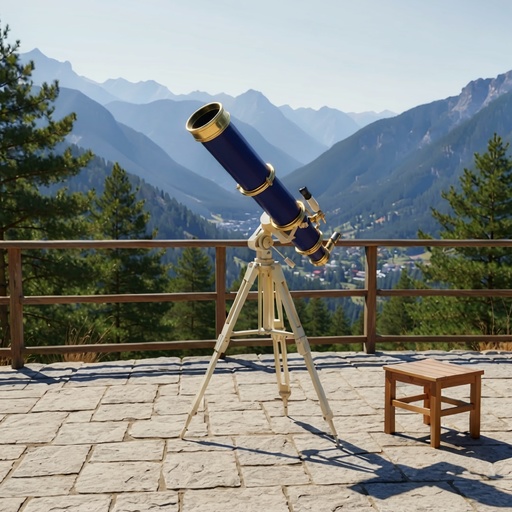}
          & \qualimg{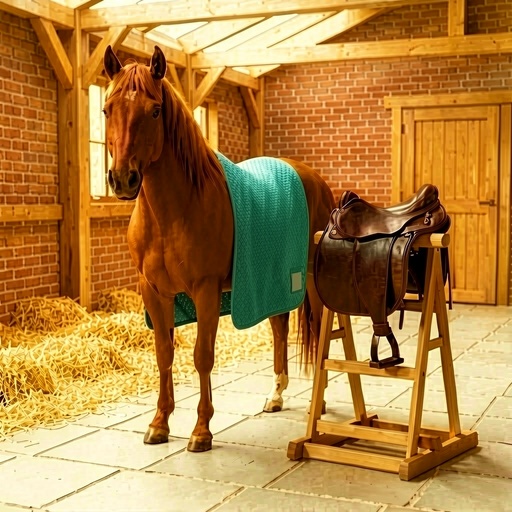}
          & \qualimg{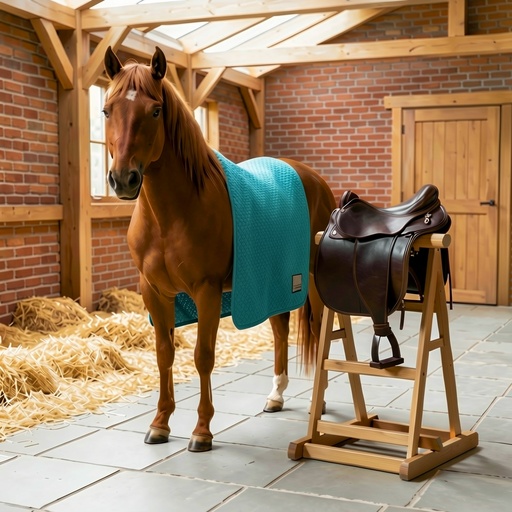}
          \\[3pt]

          & \multicolumn{2}{c}{%
              \qualinstruction{``Add a mustard party hat''}}
          & \multicolumn{2}{c}{%
              \qualinstruction{``Add a brass trim''}}
          & \multicolumn{2}{c}{%
              \qualinstruction{``Add a turquoise saddle blanket''}}
          \\[8pt]

          \rotatebox{90}{\scriptsize Edit 15}
          & \qualimg{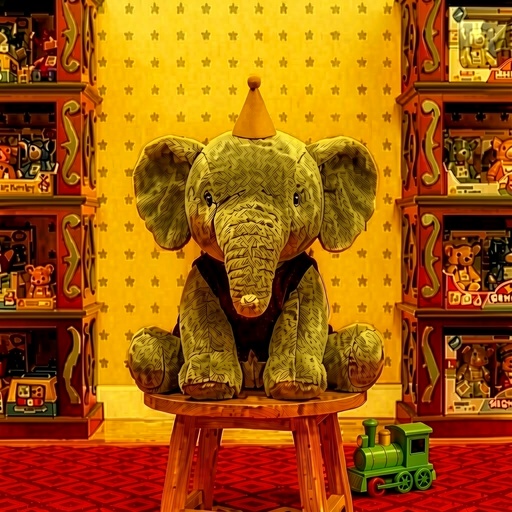}
          & \qualimg{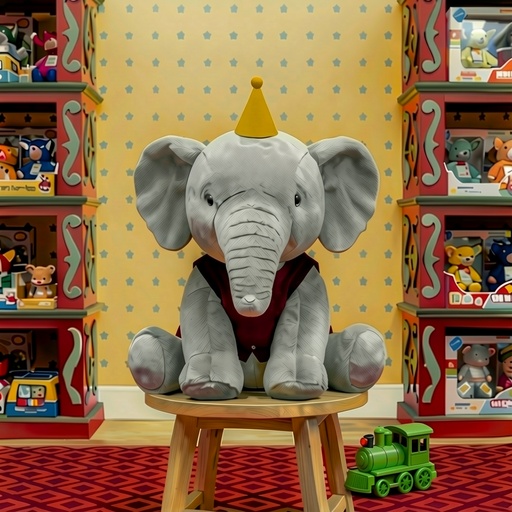}
          & \qualimg{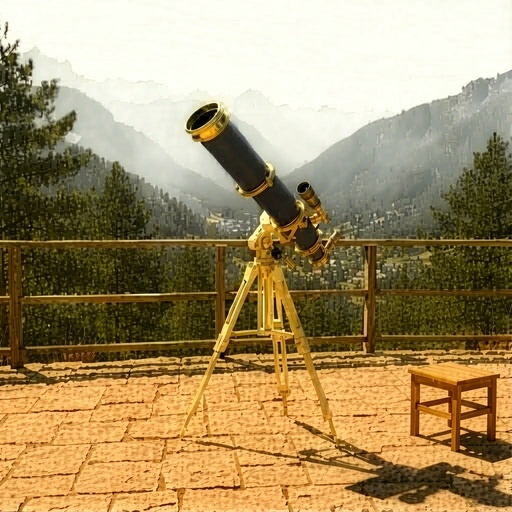}
          & \qualimg{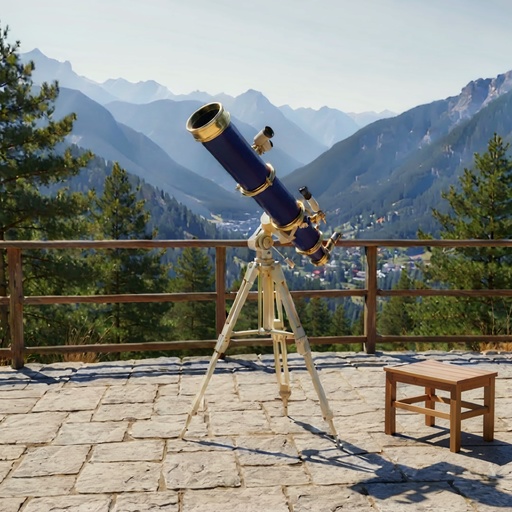}
          & \qualimg{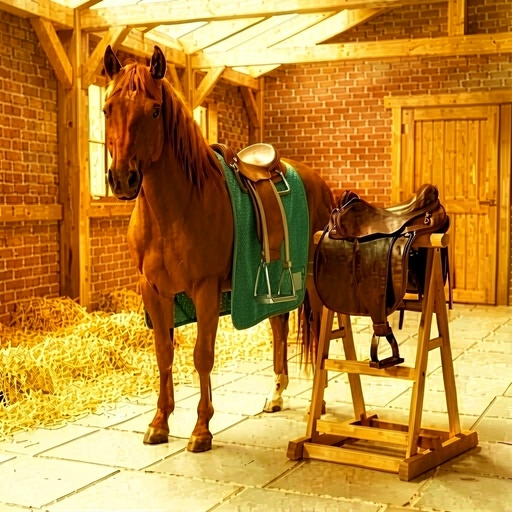}
          & \qualimg{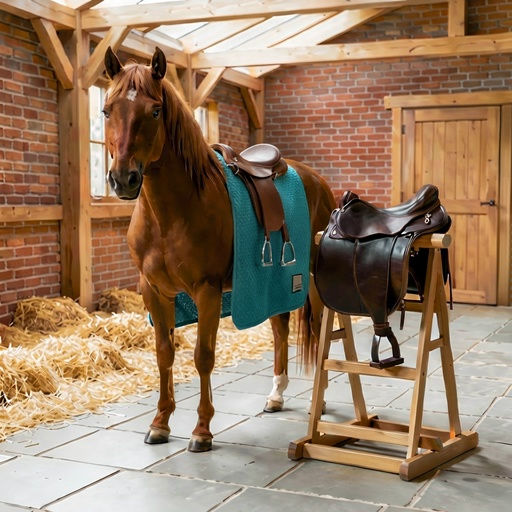}
          \\[3pt]

          & \multicolumn{2}{c}{%
              \qualinstruction{``Place a green toy train''}}
          & \multicolumn{2}{c}{%
              \qualinstruction{``Attach a small finder scope''}}
          & \multicolumn{2}{c}{%
              \qualinstruction{``Add stirrups''}}
          \\[8pt]

          \rotatebox{90}{\scriptsize Edit 20}
          & \qualimg{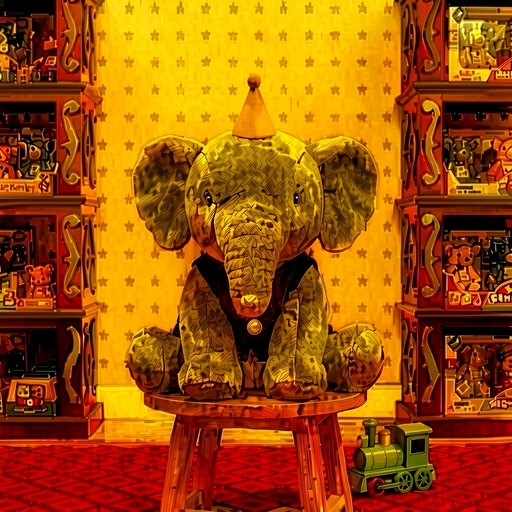}
          & \qualimg{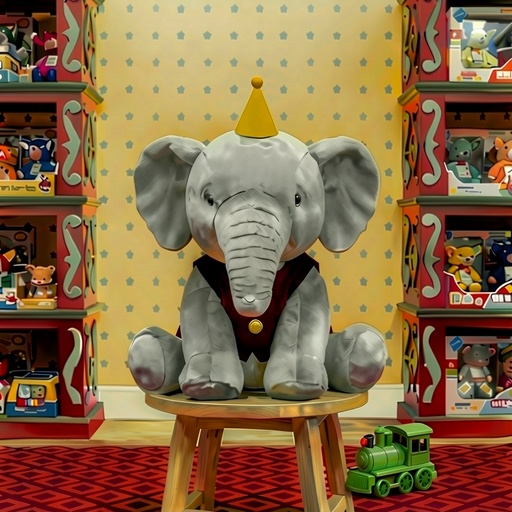}
          & \qualimg{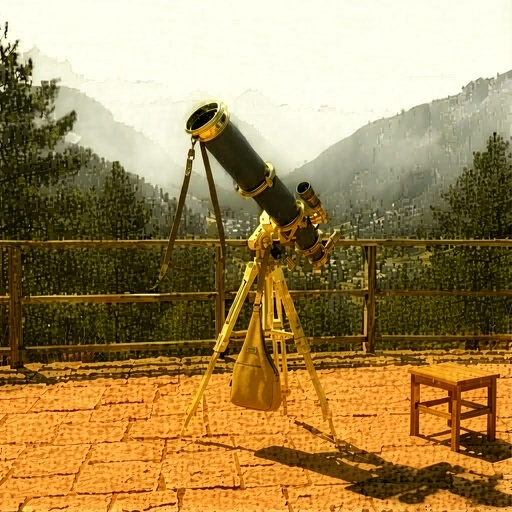}
          & \qualimg{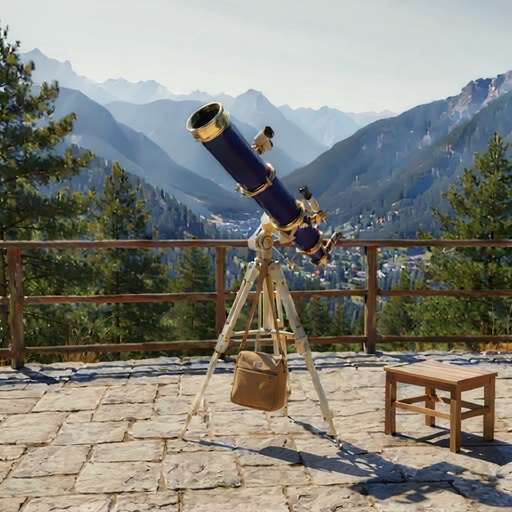}
          & \qualimg{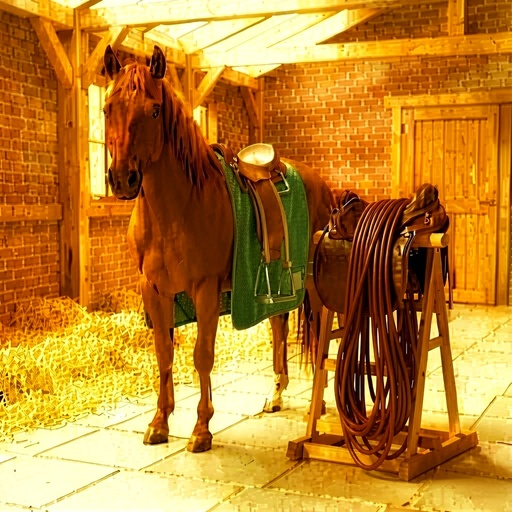}
          & \qualimg{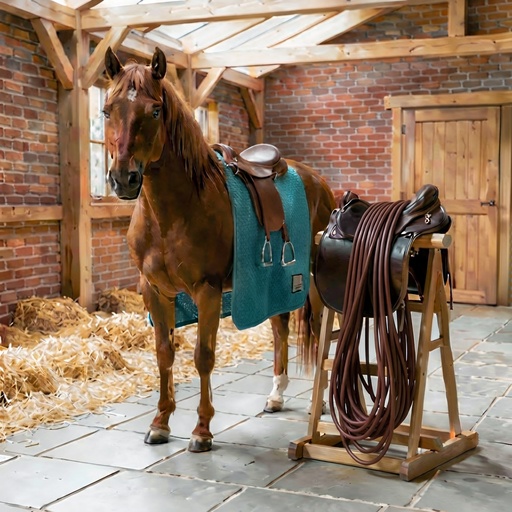}
          \\[3pt]

          & \multicolumn{2}{c}{%
              \qualinstruction{``Add a gold round button to the vest''}}
          & \multicolumn{2}{c}{%
              \qualinstruction{``Hang a tan carrying pouch''}}
          & \multicolumn{2}{c}{%
              \qualinstruction{``Hang a coiled lead rope''}}
          \\[3pt]

        \end{tabular}%
      };

      \draw[
        gray,
        line width=0.4pt,
        dash pattern=on 2pt off 2pt
      ]
        ([xshift=\qualSepOne]qualTable.north east)
        --
        ([xshift=\qualSepOne]qualTable.south east);

      \draw[
        gray,
        line width=0.4pt,
        dash pattern=on 2pt off 2pt
      ]
        ([xshift=\qualSepTwo]qualTable.north east)
        --
        ([xshift=\qualSepTwo]qualTable.south east);

    \end{tikzpicture}
  \end{adjustbox}

  \endgroup

  \caption{%
    \textbf{Additional multi-turn image editing results.}
    We show selected turns from 20-step editing sequences.
    Each displayed instruction is followed by four no-op instructions
    (``keep the image as is''), and the images show the results after
    those no-ops.
    While the original FLUX.2 Klein accumulates color shifts and loses
    detail, fine-tuning it with our training paradigm (+Ours) yields
    a more stable editing process, preserving the scene and prior edits
    while following the requested changes.
  }
  \label{fig:qualitative_results_supp}
\end{figure*}

\begin{figure*}[t]
  \centering
  \begingroup

  \def\rolloutdir{images/results/otter_rollout}

  \setlength{\tabcolsep}{1.5pt}
  \renewcommand{\arraystretch}{1.0}

  \newcommand{\rollimg}[1]{%
    \raisebox{-0.5\height}{%
      \includegraphics[width=0.11\textwidth]{#1}%
    }%
  }
  \newcommand{\rollinstruction}[1]{%
    \parbox[t]{0.11\textwidth}{%
      \centering\scriptsize #1%
    }%
  }
  \newcommand{\rolllabel}[1]{%
  \raisebox{-0.5\height}{%
    \rotatebox{90}{\scriptsize\shortstack[c]{#1}}%
  }%
}
  \newcommand{\rollturn}[1]{{\scriptsize #1}}

  \begin{adjustbox}{width=\textwidth}
    \begin{tabular}{@{}c *{8}{c}@{}}

      & \rollturn{Original}
      & \rollturn{Edit 1}
      & \rollturn{Edit 2}
      & \rollturn{Edit 3}
      & \rollturn{Edit 4}
      & \rollturn{Edit 5}
      & \rollturn{Edit 6}
      & \rollturn{Edit 7}
      \\[3pt]

      \rolllabel{FLUX.2 Klein}
      & \rollimg{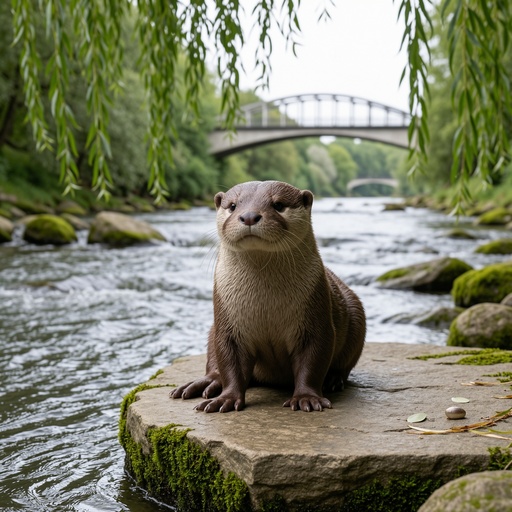}
      & \rollimg{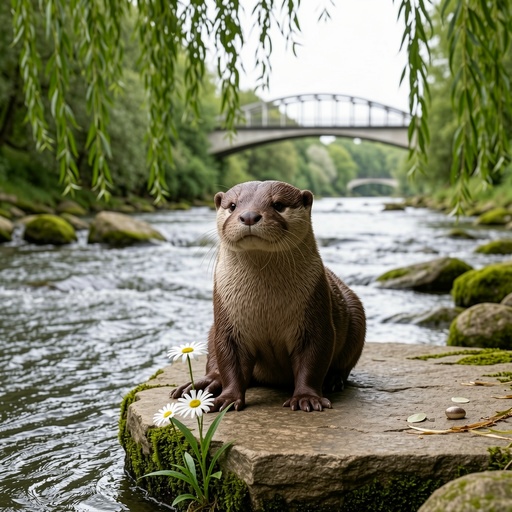}
      & \rollimg{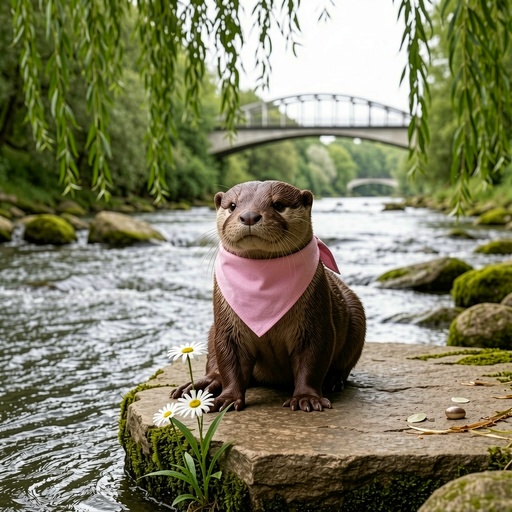}
      & \rollimg{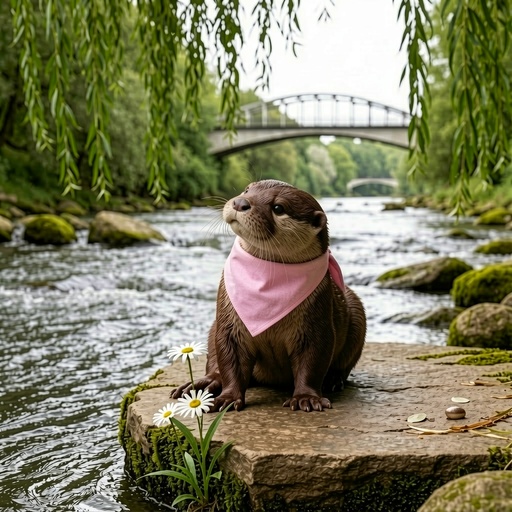}
      & \rollimg{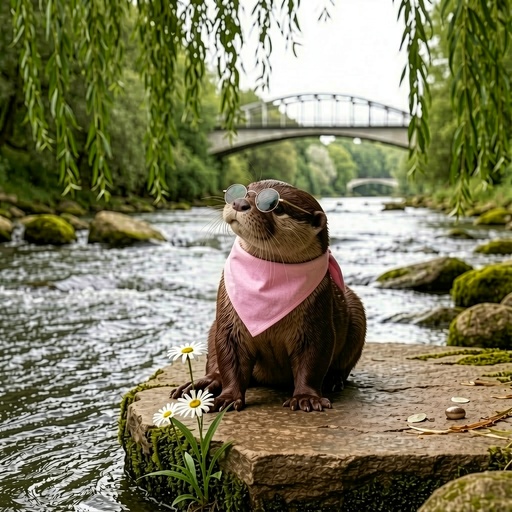}
      & \rollimg{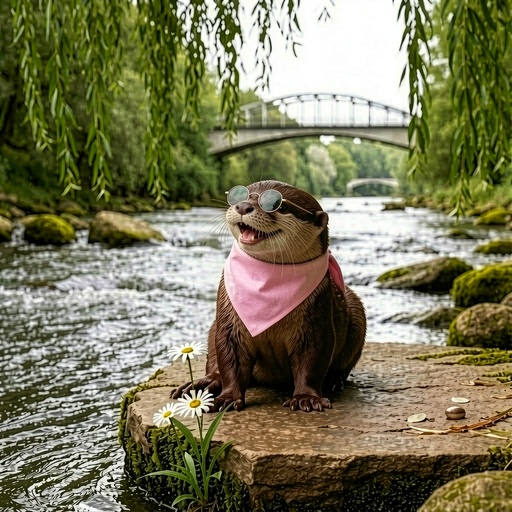}
      & \rollimg{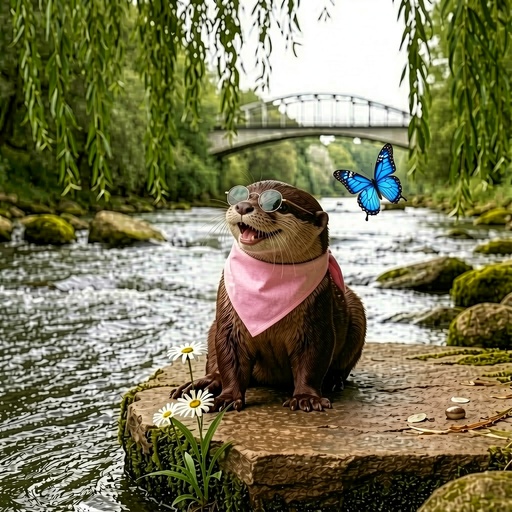}
      & \rollimg{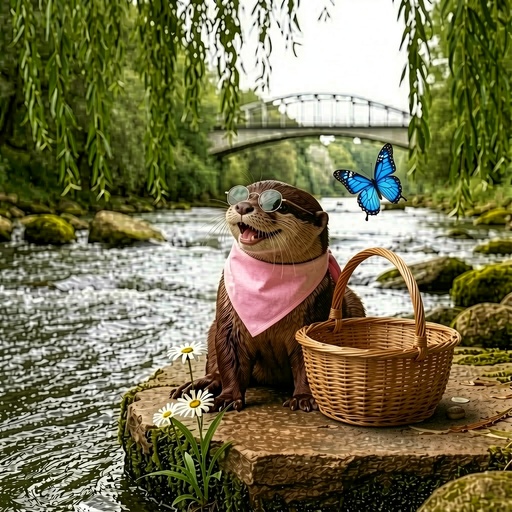}
      \\
      \noalign{\vskip 4pt}

      \rolllabel{+Ours}
      & \rollimg{\rolloutdir/original.jpg}
      & \rollimg{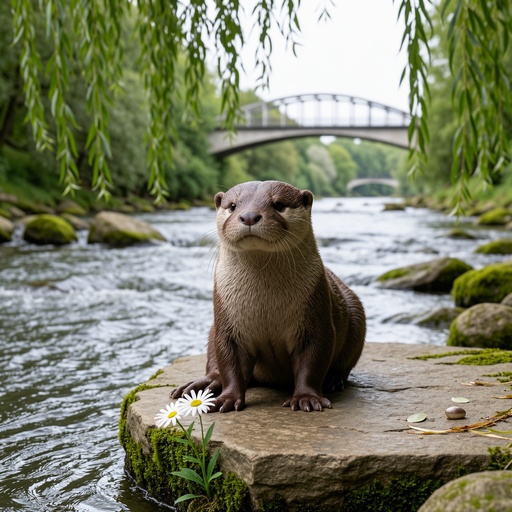}
      & \rollimg{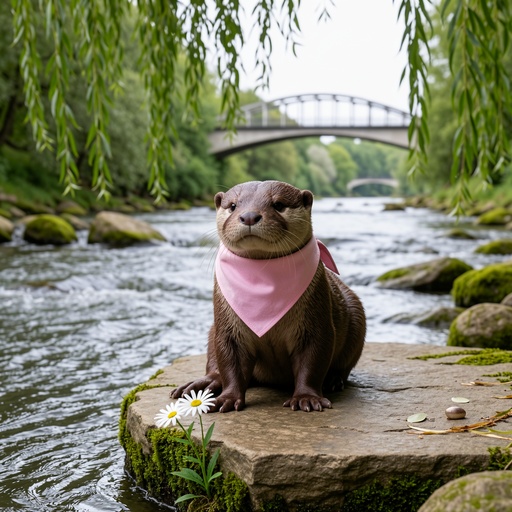}
      & \rollimg{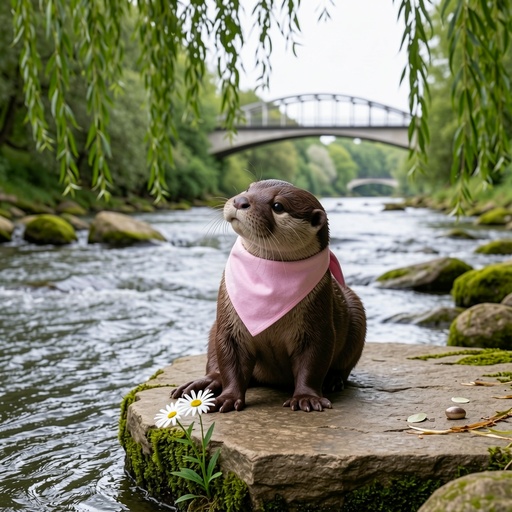}
      & \rollimg{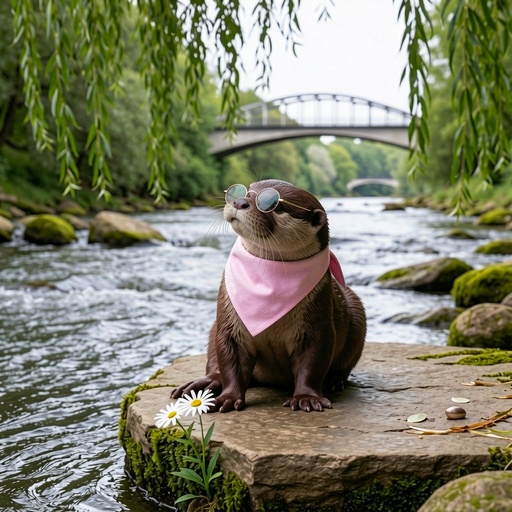}
      & \rollimg{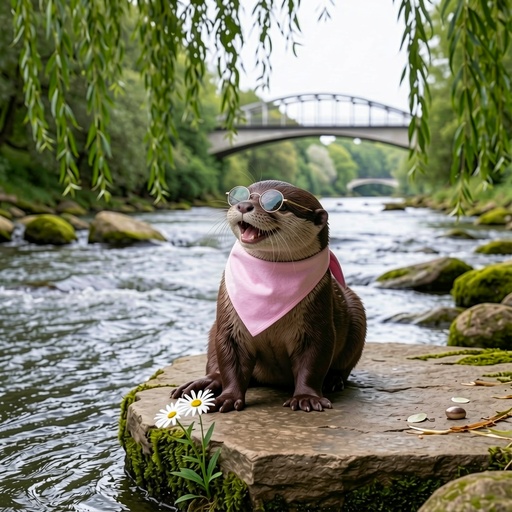}
      & \rollimg{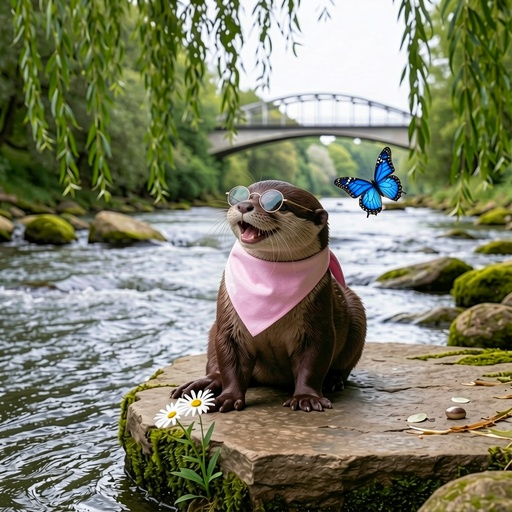}
      & \rollimg{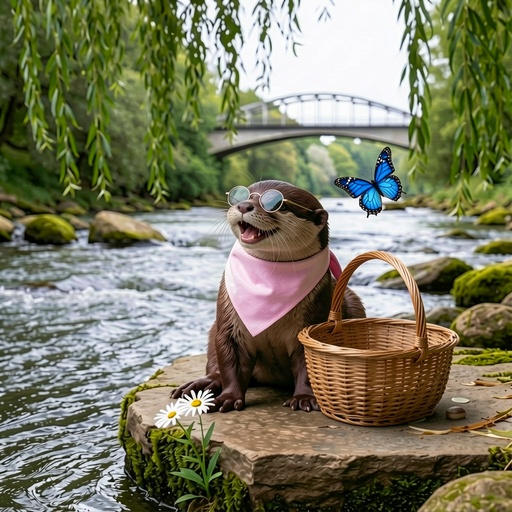}
      \\
      \noalign{\vskip 4pt}

      &
      & \rollinstruction{``Add a white daisy''}
      & \rollinstruction{``Add a pink neckerchief''}
      & \rollinstruction{``Tilt the otter's head to the side''}
      & \rollinstruction{``Add small round glasses''}
      & \rollinstruction{``Make the otter smile''}
      & \rollinstruction{``Add a blue butterfly''}
      & \rollinstruction{``Place a woven basket''}
      \\[12pt]

      & \rollturn{Edit 8}
      & \rollturn{Edit 9}
      & \rollturn{Edit 10}
      & \rollturn{Edit 11}
      & \rollturn{Edit 12}
      & \rollturn{Edit 13}
      & \rollturn{Edit 14}
      & \rollturn{Edit 15}
      \\[3pt]

      \rolllabel{FLUX.2 Klein}
      & \rollimg{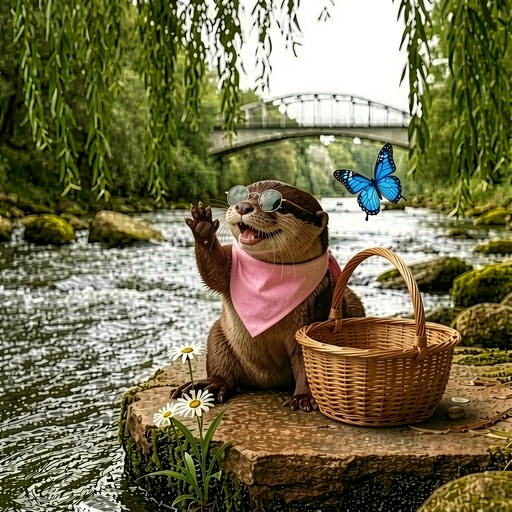}
      & \rollimg{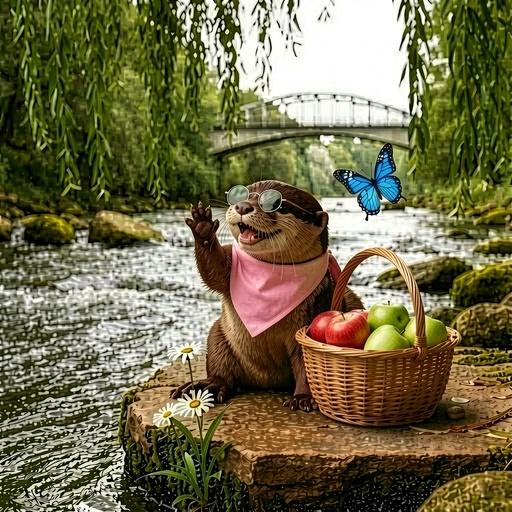}
      & \rollimg{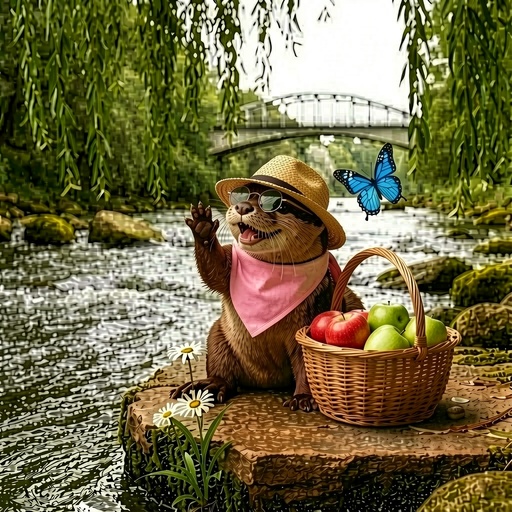}
      & \rollimg{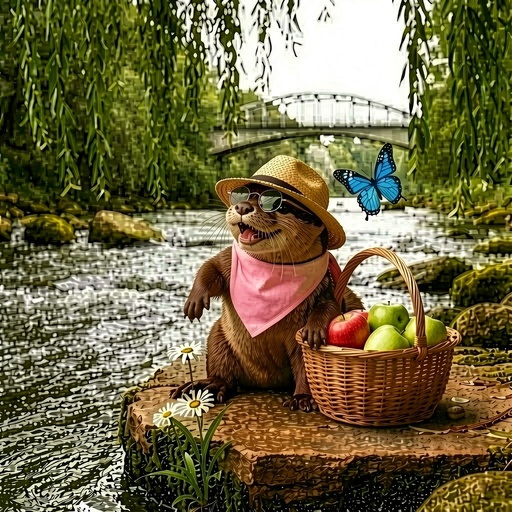}
      & \rollimg{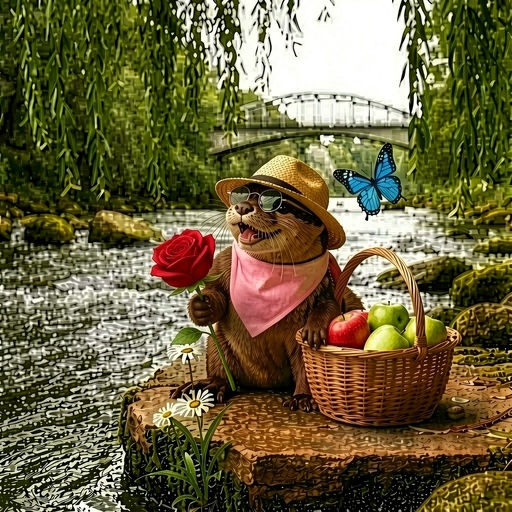}
      & \rollimg{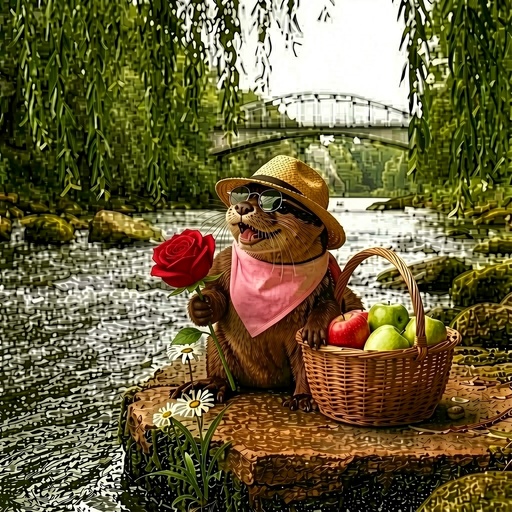}
      & \rollimg{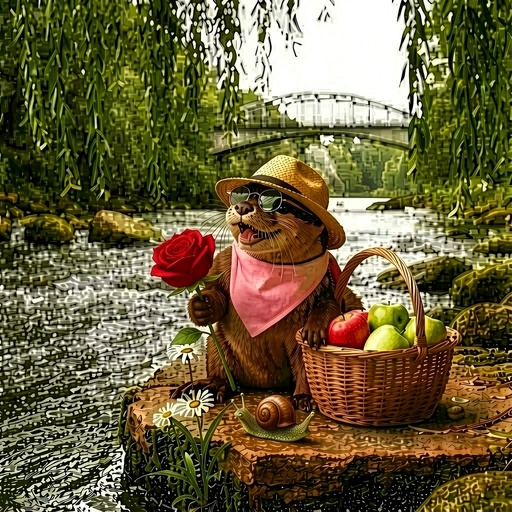}
      & \rollimg{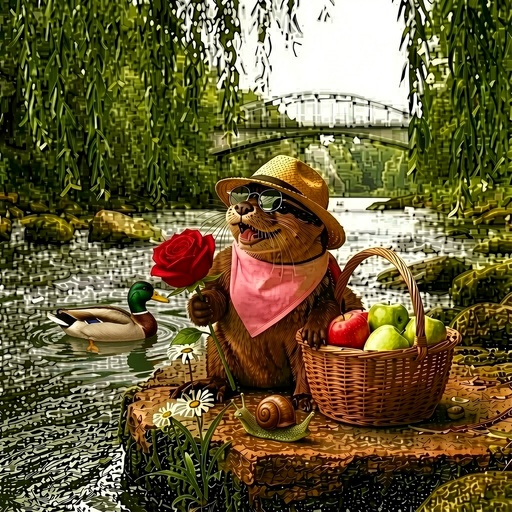}
      \\
      \noalign{\vskip 4pt}

      \rolllabel{+Ours}
      & \rollimg{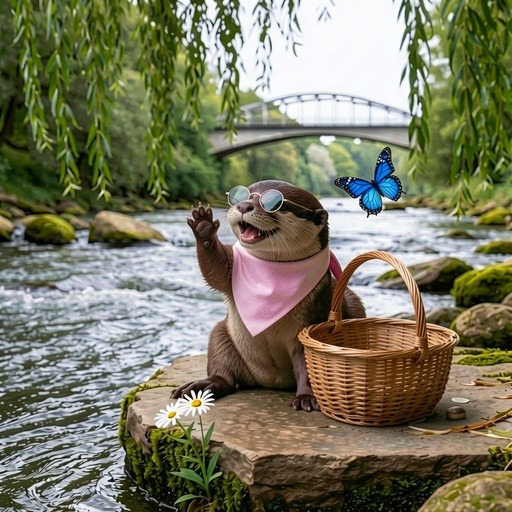}
      & \rollimg{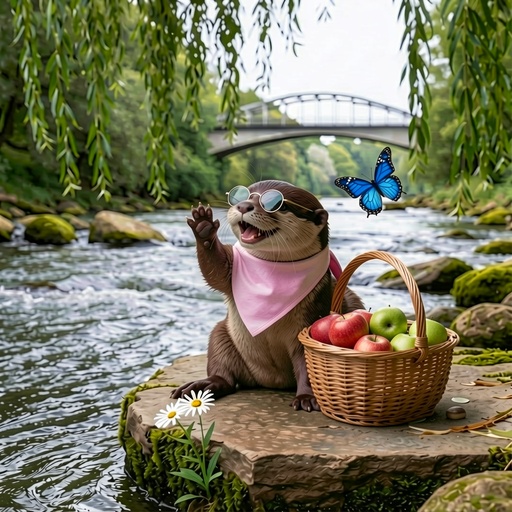}
      & \rollimg{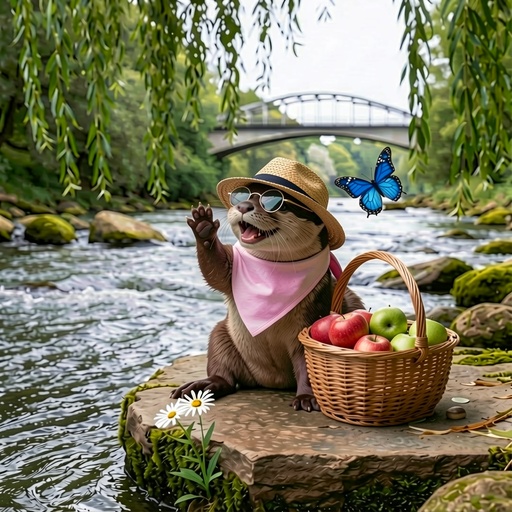}
      & \rollimg{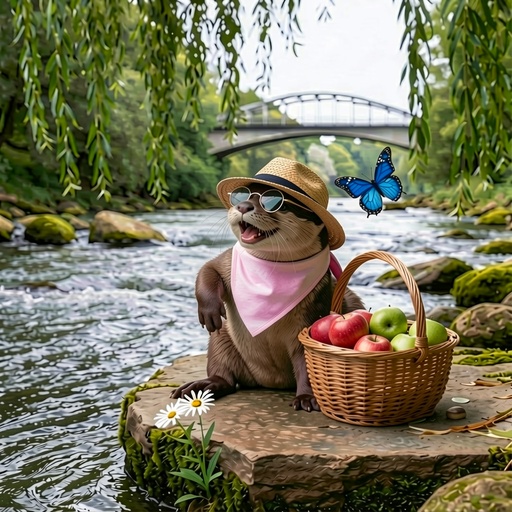}
      & \rollimg{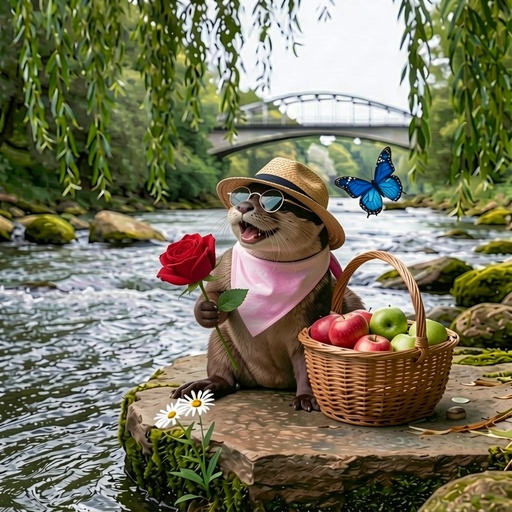}
      & \rollimg{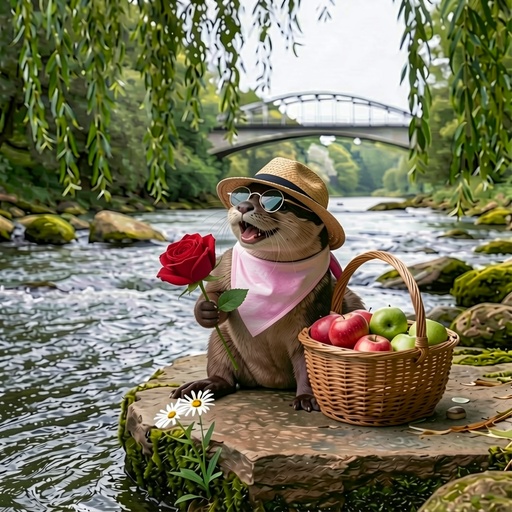}
      & \rollimg{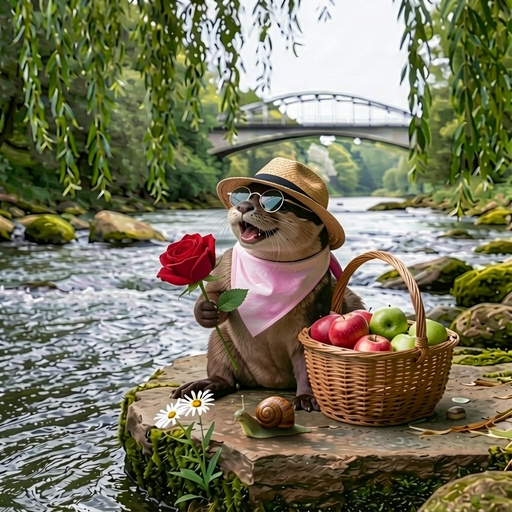}
      & \rollimg{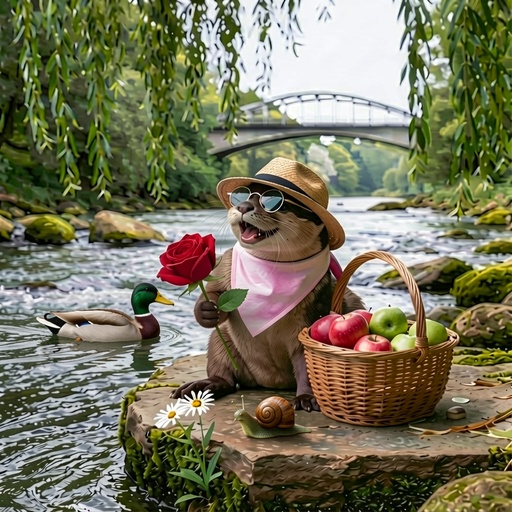}
      \\
      \noalign{\vskip 4pt}

      & \rollinstruction{``Make the otter raise its paw''}
      & \rollinstruction{``Put apples in the basket''}
      & \rollinstruction{``Add a tiny straw hat''}
      & \rollinstruction{``Make the otter lower its paw''}
      & \rollinstruction{``Place a rose in the otter's hand''}
      & \rollinstruction{``Remove the butterfly''}
      & \rollinstruction{``Add a small brown snail''}
      & \rollinstruction{``Add a duck in the river''}
      \\[3pt]

    \end{tabular}
  \end{adjustbox}

  \endgroup

  \caption{%
    \textbf{Fifteen consecutive semantic edits.}
    We compare the original FLUX.2 Klein with our fine-tuned model
    (+Ours) over a sequence of 15 edits.
    Each result becomes the input to the next edit.
    Our method reduces accumulated drift in color and scene details
    while retaining the original model's ability to follow diverse
    editing instructions, including object additions, removals,
    and pose and expression changes.
  }
  \label{fig:consecutive_edits}
\end{figure*}

\end{document}